\documentclass[10pt,twocolumn,letterpaper]{article}
\newcommand{\ARXIV}[2]{#1} %arxiv version
\usepackage{templates/CVPR/cvpr}      % To produce the REVIEW version

\usepackage[dvipsnames]{xcolor}
\usepackage{makecell}
\usepackage{graphicx}
\usepackage{amsmath}
\usepackage{amssymb}
\usepackage{times}
\usepackage{mathtools}
\usepackage{amssymb}
\usepackage{booktabs}
\usepackage{multirow}
\usepackage{tabularx}
\usepackage{balance}
\usepackage{placeins}
\usepackage{subcaption}
\usepackage{enumitem}
\usepackage{tikz}
\usepackage{pgfplots}
\usepackage{styles}
\usepackage{microtype}
\usepackage{pifont}
\usepackage{xcolor}
\usepackage{colortbl}
\usepackage{pgfplotstable}
\usepackage{worldflags}
\usepackage{environ}
\usepackage{rotating}
\usepackage[export]{adjustbox}
\usepackage[accsupp]{axessibility}

\usetikzlibrary{tikzmark,patterns,shapes.misc,shapes.geometric,positioning}
\pgfplotsset{compat=1.17}
\setlist[itemize]{itemsep=0pt, topsep=0pt}

\def\cvprPaperID{3539} % *** Enter the CVPR Paper ID here
\def\confName{CVPR}
\def\confYear{2024}

\definecolor{cvprblue}{rgb}{0.21,0.49,0.74}
\usepackage[pagebackref,breaklinks,colorlinks,citecolor=cvprblue]{hyperref}

\usetikzlibrary{shapes}
\usepackage[accsupp]{axessibility}  % Improves PDF readability for those with disabilities.
\def\paperID{3539} % *** Enter the Paper ID here
\def\confName{CVPR}
\def\confYear{2024}

\begin{document}
\newlength{\oldtextfloatsep}
\setlength{\oldtextfloatsep}{\textfloatsep}
%\addtolength{\parskip}{-20mm}

%%%%%%%%% TITLE
\title{OpenStreetView-5M: The Many Roads to Global Visual Geolocation}
% \title{Efficient 3D Semantic Segmentation with Hierarchical \METHOD}

\author{
    Guillaume Astruc\thanks{Denotes equal contributions.}\;\,\textsuperscript{1,2,5}
    \and
    Nicolas Dufour\footnotemark[1]\;\,\textsuperscript{1,6}
    \and
    Ioannis Siglidis\footnotemark[1]\;\,\textsuperscript{1}
    \and
    Constantin Aronssohn \textsuperscript{1}
    \and 
    Nacim Bouia \textsuperscript{1}
    \and
    Stephanie Fu  \textsuperscript{1,4}
    \and
    Romain Loiseau \textsuperscript{1,2}
    \and
    Van Nguyen Nguyen \textsuperscript{1}
    \and 
    Charles Raude \textsuperscript{1}
    \and
    Elliot Vincent \textsuperscript{1,3}
    \and
    Lintao XU \textsuperscript{1}
    \and 
    Hongyu Zhou \textsuperscript{1}
    \and
    Loic Landrieu\textsuperscript{1}
    \and 
    {\textsuperscript{1} \small LIGM, Ecole des Ponts, CNRS, UGE}
    \and
    {\textsuperscript{2} \small UGE, IGN, ENSG, LASTIG}
    \and 
    {\textsuperscript{3} \small Inria Paris}
    \and 
    {\textsuperscript{4} \small UC Berkeley}
    \and 
    {\textsuperscript{5} \small CESBIO, Univ de Toulouse, CNES/CNRS/IRD/INRAE/UPS}
    \and 
    {\textsuperscript{6} \small LIX, CNRS, Ecole Polytechnique, IP Paris}
}

\maketitle

%%%%%%%%% ABSTRACT
\begin{abstract}
 Determining the location of an image anywhere on Earth is a complex visual task, which makes it particularly relevant for evaluating computer vision algorithms. Yet, the absence of standard, large-scale, open-access datasets with reliably \emph{localizable} images has limited its potential. To address this issue, we introduce OpenStreetView-5M, a large-scale, open-access dataset comprising over $5.1$ million geo-referenced street view images, covering $225$ countries and territories. In contrast to existing benchmarks, we enforce a strict train/test separation, allowing us to evaluate the relevance of learned geographical features beyond mere memorization.
 To demonstrate the utility of our dataset, we conduct an extensive benchmark of various state-of-the-art image encoders, spatial representations, and training strategies.
 All associated codes and models can be found at \GITHUB. 
\end{abstract}
\setlength{\parskip}{-0.14em}
%%%%%%%%% BODY TEXT
%===================================================
\section{Introduction}
\begin{figure}
\definecolor{myred}{RGB}{255,0,0}
\definecolor{myMagenta}{RGB}{181,86,255}
\definecolor{mygreen}{RGB}{21,226,17}
\definecolor{myblue}{RGB}{10,5,255}
    \centering
    \begin{tikzpicture}
        \node (img) at (0,0) {\includegraphics[width=1\linewidth]{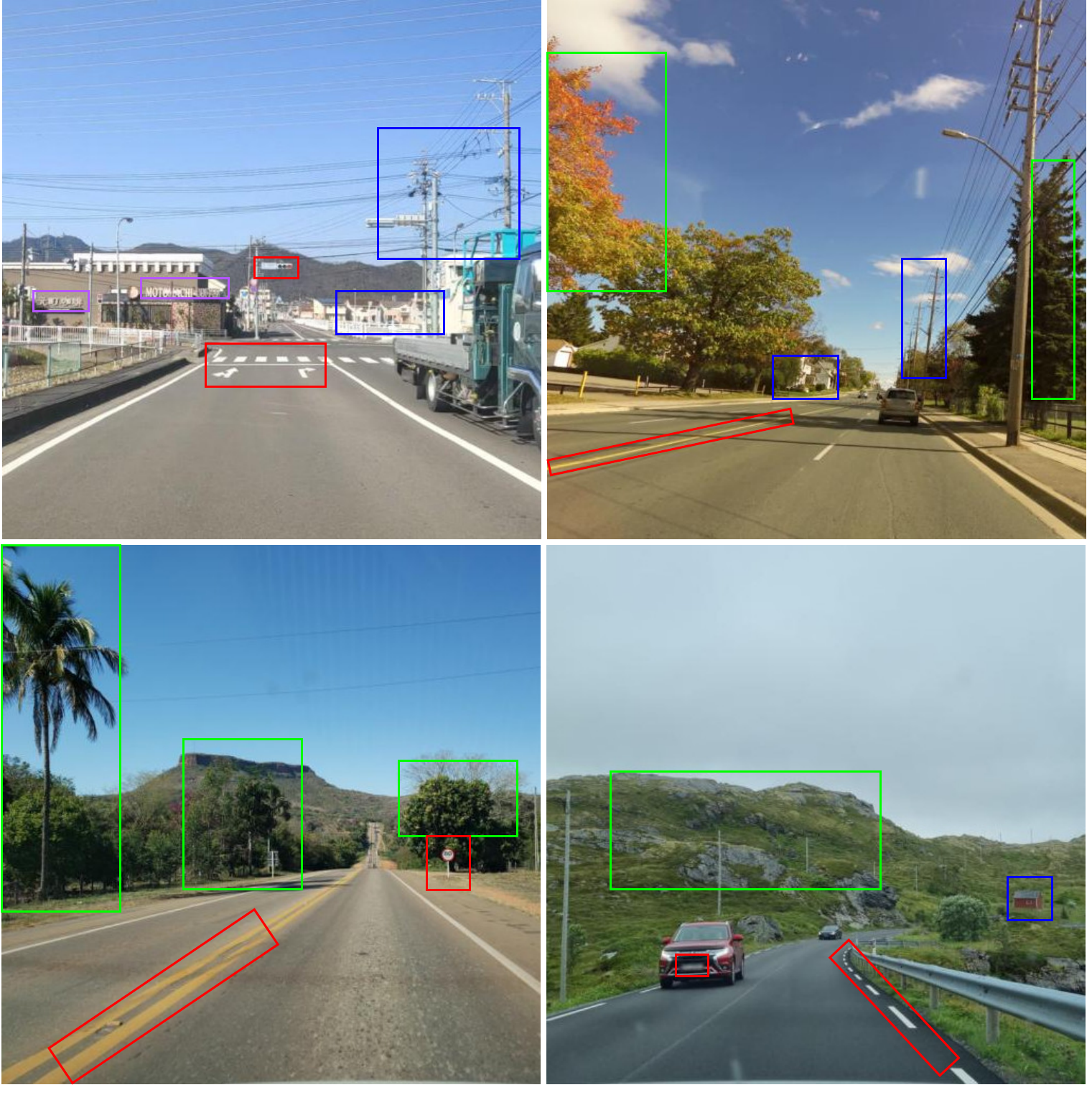}};
        \node[below right = -2mm and -61mm of img.south east, align=right] {\tiny drivephotograph, and\_eng, gclem, bootprint, Mapillary, licensed under CC-BY-SA.};
    \node[below=0mm of img.south] {
    \begin{tabular}{c}
    \begin{tabularx}{.9\linewidth}{cccc}
    \footnotesize               
    \hspace{-5mm}\textcolor{mygreen}{climate/vegetation} &
    \footnotesize  
    \textcolor{myred}{traffic markers} & 
    \footnotesize  
    \textcolor{myblue}{architecture}& 
    \footnotesize  
    \textcolor{myMagenta}{culture/script}
    \end{tabularx}
    \end{tabular}
    };
    \end{tikzpicture}  
    \vspace{-7mm}
    \caption{{\bf Global Visual Geolocation.} 
    Predicting the location of an image taken anywhere in the world from just pixels requires detecting a combination of clues of various abstraction levels \cite{mehta2016exploratory}. Can you guess where these images were taken?\protect\footnotemark  }
    \label{fig:teaser}
\end{figure}

%==================================================
While natural image classification is the standard for evaluating computer vision methods  \cite{imagenet-1k,ridnik2021imagenet,van2018inaturalist}, global geolocation offers a compelling alternative task.
In contrast to classification, where the focus is often a single object, geolocation involves detecting and combining various visual clues, like road signage, architectural patterns, climate, and vegetation.
Predicting a single GPS coordinate or location label from these observations necessitates a rich representation of both the Earth's culture and geography; see \figref{fig:teaser} for some examples. 
Furthermore, the abundance of geo-tagged street-view images depicting complex scenes with a clear and consistent point of view makes this task appropriate for training and evaluating modern vision models.

%While natural images are the standard for the training and evaluation of computer vision methods, in tasks such as classification \cite{imagenet-1k,ridnik2021imagenet,van2018inaturalist}, street view images, present a compelling alternative, thanks to the abundance of available geotagged images, and the complexity of their scenes. In contrast to natural image classification, where the focus is often a single object, geolocation involves identifying and combining various visual clues, like road signage, architectural patterns, or climate and vegetation, into a single GPS coordinate or location label. As such, robust global geolocation, implies rich representations of both the earth's culture and geography; see \figref{fig:teaser} for some examples.

Despite this potential, few supervised approaches are trained and evaluated for the task of geolocation.
%Despite the challenge, it is highly unlikely that a popular supervised approach will be trained and evaluated in the task of geolocation. 
We attribute this to the limitations of existing geolocation datasets:
(i) Large and open geolocation datasets contain a significant portion of noisy and non-localizable images \cite{Im2gps,Im2GPS++YFCC4k+Im2GPS3k, izbicki2020exploiting};
\footnotetext{
\rotatebox[origin=c]{180}
{
\begin{minipage}[t]{.96\linewidth}
{\bf From top left to bottom right:} Nagoya, Japan; Ontario, Canada; Mato Grosso, Brazil; Lofoten, Norway.
 \end{minipage}
}
}
(ii) Street view datasets are better suited for the task but are both proprietary and expensive to download \cite{luo2022geolocation, haas2023learning, Google-World-Streets-15K-geolocation,chen2011city,hausler2021patch,seymour2018semantically}. 
To address these issues, we introduce OpenStreetView-5M (OSV-5M), an open-access dataset of 5.1 million high-quality and crowd-sourced street view images. Our ambition is to make both street view images and global geolocation new standards for measuring progress in deep learning.

Automating visual geolocation has significant potential benefits, with direct applications in fields such as journalism, forensics, as well as historical and cultural studies. Learning robust geographical representations may also be valuable for various deep learning challenges, including self-supervised learning and generative modeling, or the development of more interpretable AI systems. Thanks to its size and scope, and its strict train/test split, OSV-5M serves as a robust and reliable benchmark for computer vision models.
To demonstrate this, we design an extensive evaluation experiment to measure the impacts of various factors such as pretraining strategies, model scale, spatial representations, fine-tuning approaches, contrastive losses, and auxiliary tasks.
%===================================================
\section{Related Work}
%===================================================
\definecolor{darkgreen}{rgb}{0, 0.5, 0}
\begin{figure}
\input{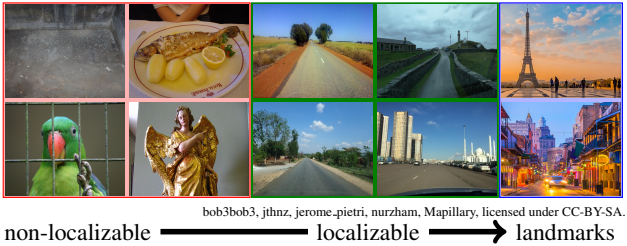}
\vspace{-0.5em}
\caption{{\bf Localizable vs Non-Localizable.}
Images from our dataset (\textcolor{darkgreen}{green}) occupy the space between weakly localizable images (\textcolor{red}{red}) like the ones from the test set of Im2GPS3k \cite{Im2GPS++YFCC4k+Im2GPS3k} and landmark images used to advertise CV conferences (\textcolor{blue}{blue}).
}
% Strongly localizable images offer clear visual cues for inferring or narrowing down location; (\ref{fig:data:geoloc:loc}) Weakly localizable images contain only indirect hints; (\ref{fig:data:geoloc:nonloc}) Non-localizable images lack apparent cues related to their location.}    
\label{fig:data:geoloc}
\end{figure}

In this section, we detail the notion of image localizability (\secref{sec:data:localizability}), the main existing geolocation datasets (\secref{sec:related:data}), and geolocation methods (\secref{sec:related:method}).

%introduce OpenStreetView-5M in \secref{sec:data:motivation} %We conclude this section by outlining the detailed dataset creation process in \secref{sec:data:constitution}.

%===============================================
\subsection{Localizability}
%==============================================
\label{sec:data:localizability}

As noted by Izbicki et al. \cite{izbicki2020exploiting}, images exhibit a range of localizability, an inherently perceptual concept, see \figref{fig:data:geoloc}. Non-localizable images lack information that connects them to a specific location or are of too low quality to properly analyze. Weakly localizable images only contain vague or indirect hints, such as people, animals, and objects in indoor scenes. Localizable images should contain enough information to allow for an informed guess relative to their location. For example, street view images are generally localizable as they typically contain salient features indicative of the local environment such as climate, nature, architecture, or utility and regulatory infrastructure. At the far end of the spectrum, landmark images showcase emblematic monuments or iconic landscapes, making their location instantly identifiable to most viewers.

According to this criteria, a visual inspection suggests that 35\% of the images in Im2GPS3k, a dataset commonly used to benchmark geolocation methods \cite{Im2GPS++YFCC4k+Im2GPS3k}, are non-localizable. When used for evaluation, this may lead to unreliable errors or promote methods that have memorized biases of the training distribution. When used for training, non-localizable images can lead to sub-optimal representations or encourage spurious correlations. OSV-5M predominantly comprises localizable street view images whose accurate geolocation requires robust geographical representations.

%Thus during evaluation, such a dataset can either provide unreliable errors or promote methods that have memorized biases of the training distribution. During training, the presence of non-localizable images within a benchmark dataset introduces noise that can either lead to sub-optimal representations or encourage spurious correlations. Our dataset predominantly comprises of street view images, which are typically localizable but whose accurate localization implies the learning robust geographical image representations.

%======================================================
\subsection{Geolocation Datasets}
%=======================================================

\label{sec:related:data} 
We motivate the need for OSV-5M by reviewing existing geolocation datasets from the two main sources of geotagged images: web-scraped and street view images, see  \tabref{tab:dataset}.

%the need to introduce our dataset by reviewing the available datasets for geolocation from the two current main sources of geotagged images: web-scraped and street view images. We present in \tabref{tab:dataset} a comparison between existing geolocation datasets and OpenStreetView-5M. 

\paragraph{Web-Scraped.}
Image hosting platforms like Flickr provide a near-endless source of geotagged images, which has been used to create large open datasets, like YFC100M~\cite{yfcc100m}.
Most images correspond to personal or amateur photographs representing food, art, and images of pets and friends, and are either weakly or non-localizable.
Even strongly localizable images are typically taken in tourist spots, injecting an often Western cultural bias towards recognizable landmarks \cite{yfcc-analysis}. The provided location metadata can be occasionally missing or inaccurate, and the online nature of these images implies they can be easily removed, hindering reproducibility\footnote{\label{fn:missing}60\% of the 2014 YFCC-split \cite{flickr-mouselly} was deleted by 2020 \cite{izbicki2020exploiting}!}. For evaluation purposes, cleaner subsets have been proposed that improve both the image distribution coverage and annotation quality \cite{YFCC-Val26k-Interpretable-semantic-photo-geolocation-DL,Im2GPS++YFCC4k+Im2GPS3k}, but remain still heavily biased and predominantly non-localizable. 
 %, yet not surpassing the fundamental limitations that come from such image distributions. 
Despite their small scope and size, these datasets are currently the primary means of evaluating geolocation models.
%Most images correspond often to personal or amateur photography involving food, art, and images of pets and friends, being either weakly or non-localizable.
%Even strongly localizable images, typically are taken in tourist spots, injecting a substantial cultural bias towards recognizable, often Western, landmarks \cite{yfcc-analysis}. The provided location metadata can be occasionally missing or inaccurate, and the online nature of these images implies they can be easily removed, hindering reproducibility\footnote{\label{fn:missing}60\% of the 2014 YFCC-split \cite{flickr-mouselly} was deleted by 2020 \cite{izbicki2020exploiting}!}. For purposes of evaluation, cleaner subsets have been proposed to improve both the image distribution coverage and annotation quality \cite{YFCC-Val26k-Interpretable-semantic-photo-geolocation-DL,Im2GPS++YFCC4k+Im2GPS3k}, yet not surpassing the fundamental limitations that come from such image distributions. 
%Despite their small scope and size, these datasets are currently the primary means of evaluating geolocation models.

\paragraph{Street View.} 
Conversely, street view images tend to be strongly localizable. Captured through panoramic cameras or dash-cams, they depict in high quality a vehicle's surroundings, which corresponds mostly to outdoor scenes with rich geographical cues. Google famously provides a global street view coverage, which is, however, expensive to acquire for academic purposes (\$1000 for 150k images) and cannot be shared. Existing open datasets from this source either only consist of dense samples from 3 US cities meant for navigation \cite{mirowski2019streetlearn-navigation,nn-UCF-GSV-Dataset-2014}, or are inaccessible \cite{haas2023learning,luo2022geolocation, Google-World-Streets-15K-geolocation}.

Luckily, crowd-sourced platforms such as Mapillary \cite{mapillary} offer a global and diverse source of open-access street view images for various environments, from dense cities and suburbs to remote and inhabited landscapes. These images have been used to construct several benchmarks for multiple tasks other than geolocation, including depth estimation \cite{map-depth}, semantic segmentation \cite{map-vistas}, traffic sign detection and classification \cite{map-traffic-signs}, place recognition \cite{map-lifelong-pr} and visual localization \cite{map-vloc}. 
With $5.1M$ Mappilary images taken across the globe, OSV-5M is the largest open-access street-view image dataset and the only one designed for global geolocation.
%Our dataset OSV-5M is the largest Mappilary-based, open street-view image dataset and the only one that has been created for the task of large-scale street-view geolocation benchmark, providing 5.1M of open-access images with strong global coverage.
OSV-5M has a similar order of magnitude to popular YFCC-based geolocation train sets \cite{flickr-mouselly,split-yfcc}, and comes with a clean test set that is 33 times bigger than the current largest street-view image test benchmark~\cite{Google-World-Streets-15K-geolocation} (which is not openly accessible).

\begin{table}[t]
     \centering
         \caption{{\bf Geolocation Datasets.} OpenStreetView-5M contains strongly localizable street views with access, scope, and size comparable to web-scraped databases.}
     \label{tab:dataset}
     \resizebox{.75\linewidth}{!}{
     \begin{tabular}{@{}llccc@{}}
          \toprule
         \multirow{2}{*}{Image Source} &\multirow{2}{*}{size} & open- & \multirow{2}{*}{scope}\\
         &&access&      \\\midrule
          Web-scraped \\\greyrule
        Im2GPS \cite{Im2gps}  & 237 &\true & biased\\
         Im2GPS3k \cite{Im2GPS++YFCC4k+Im2GPS3k}  & 2997 &\true & biased\\
        YFCC4k \cite{Im2GPS++YFCC4k+Im2GPS3k}  & 4536 & \true & biased\\
        YFCC26k \cite{YFCC-Val26k-Interpretable-semantic-photo-geolocation-DL}  & 26k &\true & biased\\
        MP-16 \cite{split-yfcc} & 4.7M & \true & biased\\
        Moussely \etal \cite{flickr-mouselly} & 14M/6M\footref{fn:missing}
        & \true & global \\
        YFCC100M \cite{yfcc100m}  & 100M & \true & biased \\
        PlaNet \cite{PlaNet}  & 125M & \false & biased \\
          \midrule
         Street view \\\greyrule
        Google-WS-15k \cite{Google-World-Streets-15K-geolocation}  & 15k & \false & global \\
        GMCP \cite{nn-UCF-GSV-Dataset-2014} & 105k & \false & 3 cities \\
        StreetCLIP  \cite{haas2023learning} & 1M & \false & unknown \\
          \bf OpenStreetView-5M & 5.1M &\true & global
         \\ \bottomrule
     \end{tabular}
    }

 \end{table}

%====================================================
\subsection{Geolocation Methods}
%===================================================
\label{sec:related:method}
Place recognition \cite{zhang2021visual} and visual localization \cite{sattler2018benchmarking,pion2020benchmarking,piasco2021improving,ding2019camnet,kendall2015posenet} are popular tasks that consist in finding the pose of images in a known scene. In contrast, visual geolocation predicts 2D coordinates or discrete locations (\eg, countries), and aims for lower accuracy and the ability to generalize to unseen areas \cite{hays2015large}.
Existing geolocation approaches can be categorized by whether they treat geolocation as an image retrieval problem, a classification problem, or both.

\paragraph{Image Retrieval-Based Approaches.}
A straightforward method for image localization is to find the most similar image in a large image database and predict its location~\cite{Im2gps}.
%its \emph{nearest neighbor} from a dense large dataset and output  its location~\cite{Im2gps}.
The first successful approaches involved retrieving the nearest image in a space of handcrafted features such as color histograms~\cite{Im2gps}, gist features~\cite{ oliva2006building}, or textons~\cite{martin2001database}. It was later improved with SIFT features and support vector machines \cite{Im2gps-pp}. Deep features further boosted the performance of these approaches \cite{Im2GPS++YFCC4k+Im2GPS3k}.
While such models typically exhibit high performance given a large and dense enough image database, they do not involve representation learning. Consequently, unless provided with robust features, they may perform poorly in sparsely represented or dynamically changing environments. %, and are limited in their interpretability.

\paragraph{Classification-Based Approaches.}
Geolocation can also be approached as a classification problem by discretizing latitude and longitude coordinates. The choice of partition is critical, ranging from regular \cite{PlaNet}, adaptive \cite{Google-World-Streets-15K-geolocation}, semantic-driven \cite{ISPN}, combinatorial \cite{CPlaNet}, administrative \cite{translocator,haas2023pigeon}, and hierarchical~\cite{Google-World-Streets-15K-geolocation,Im2GPS++YFCC4k+Im2GPS3k} partitions. 
Classification-based methods must strike a delicate balance between the quantity and size of cells; if the discretization is too coarse, the performance will be limited, while too many small cells may not have enough samples for learning-based methods. Furthermore, a typical classification loss such as cross-entropy does not incorporate the distance between regions:
confusing two adjacent cells is equivalent to mistaking the continent.

\paragraph{Hybrid Approaches.}
Retrieval and classification approaches can be combined to overcome the limitations of discretization. This can be achieved using ranking losses \cite{Im2GPS++YFCC4k+Im2GPS3k} or contrastive objectives \cite{efficientnet}. Haas \etal \cite{haas2023pigeon} follow a classification-then-regression approach based on prototype networks. Finally, Izbicki \etal \cite{izbicki2020exploiting} go beyond single-location prediction by estimating probability distributions based on spherical Gaussians.%, that properly combine benefits of both classification and regression.

\section{OpenStreetView-5M}
%===================================================

\label{sec:data:motivation} 
 OpenStreetView-5M establishes a new open benchmark for geolocation by providing a large, open, and clean dataset. 
The Appendix details the construction of the dataset. As detailed below, OpenStreetView-5M improves upon several limitations of current geolocation datasets.

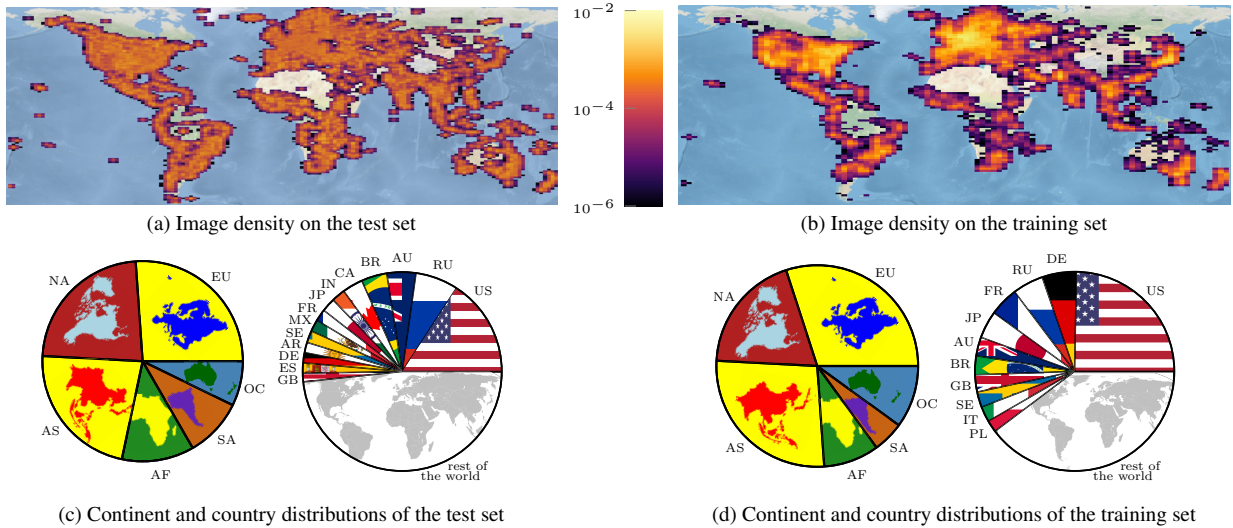
\begin{figure*}[t]
\pgfplotsset{%magma colormap
    /pgfplots/colormap={magma}{
        rgb255=(0,0,4)
        rgb255=(31,12,72)
        rgb255=(85,15,109)
        rgb255=(136,34,106)
        rgb255=(186,54,85)
        rgb255=(227,89,64)
        rgb255=(249,140,10)
        rgb255=(252,191,73)
        rgb255=(251,231,136)
        rgb255=(252,253,191)
    }
}
\centering
\begin{tabular}{c@{\;}c@{\;}c}

\begin{subfigure}{0.42\linewidth}
    \includegraphics[width=\linewidth]{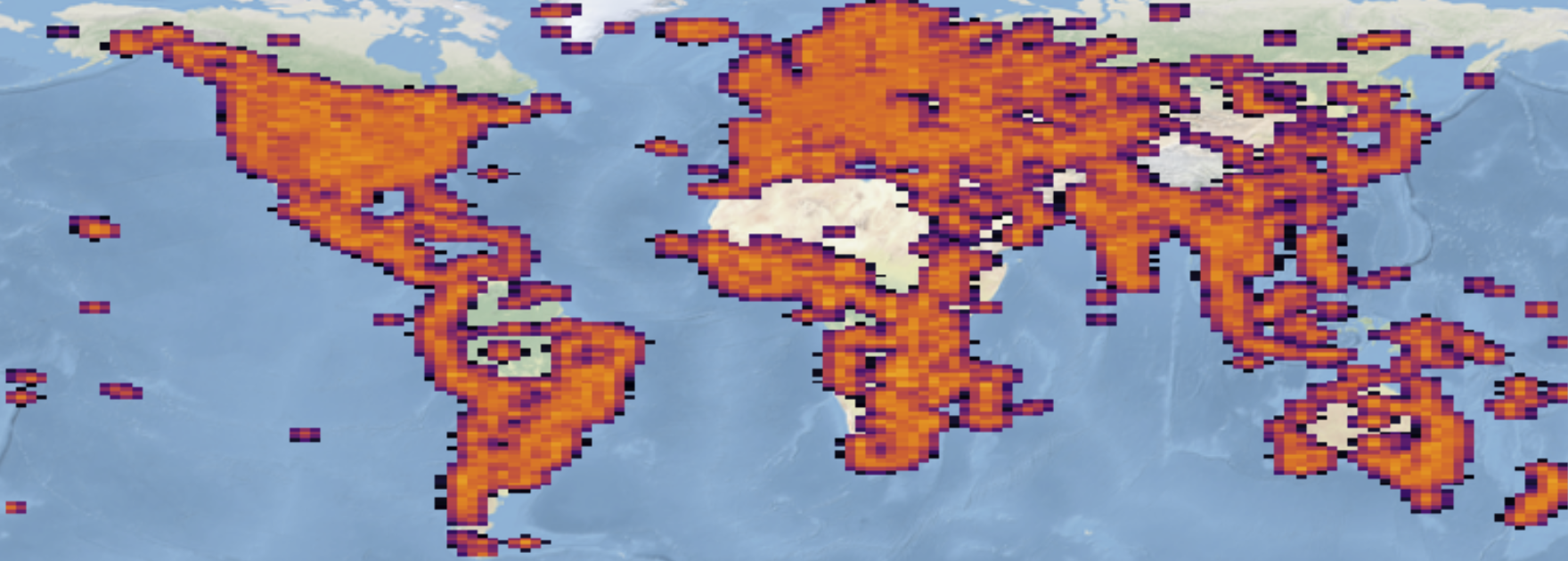}
     \caption{Image density on the test set}
\end{subfigure}
&  %shared colormap with 2axis, nightmarish
\hspace{-1.8cm}
\begin{tikzpicture}
\node [draw=none] at (0,-1.6) {};
\node [draw=none] at (0,0) {
\begin{axis}[
    hide axis,
    scale only axis,
    height=2.6cm,
    width=0.5cm,
    colormap name=magma,
    colorbar,
    colorbar style={
        ymode=log,
        ymin=0.000001, ymax=0.01,
        ytick={0.000001,0.0001,0.01},
        ytick pos=left,
        yticklabel pos=left,
        font=\fontsize{5pt}{5pt}\selectfont, % Extremely small font size
        y axis line style={draw=none} % <-- Hides the y-axis line
    },
    point meta min=0.000001, point meta max=0.01,
    at={(2cm,0cm)} % Adjust position to place it next to the left y-axis
]
% This is the dummy plot:
\addplot [draw=none, forget plot] coordinates {(0,1) (1,1000)};
\end{axis}
};
\end{tikzpicture}
&
\begin{subfigure}{0.42\linewidth}
    \includegraphics[width=\linewidth]{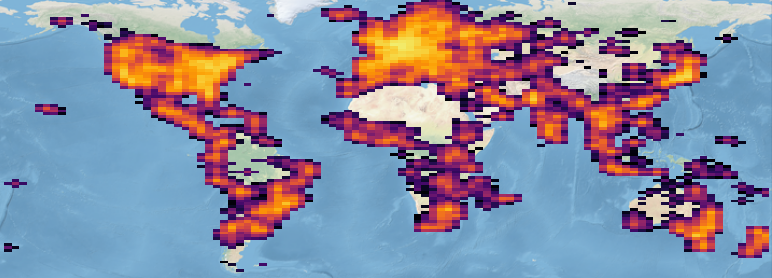}
    \caption{Image density on the training set}
\end{subfigure}
\\

%\multicolumn{3}{c}{
%\begin{tabular}{c@{}c}
  \begin{subfigure}{0.40\linewidth}
   \begin{tabular}{c@{}c}
 \includegraphics[width=.45\linewidth]{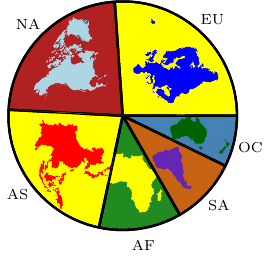}
 &
 \includegraphics[width=.44\linewidth]{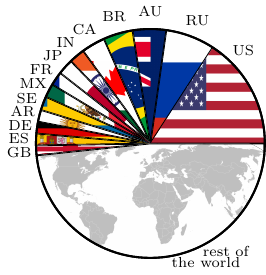}
  \end{tabular}
  \caption{Continent and country distributions of the test set}
 \end{subfigure}
 &
 &
\begin{subfigure}{0.40\linewidth}
 \begin{tabular}{c@{}c}
 \includegraphics[width=.45\linewidth]{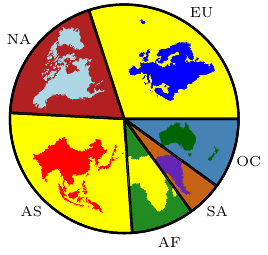}
 &
\includegraphics[width=.44\linewidth]{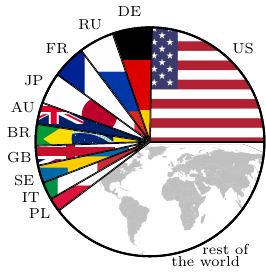}
 \end{tabular}
 \caption{Continent and country distributions of the training set}
 \end{subfigure}
%\end{tabular}
%}
\end{tabular}
\vspace{-2mm}
    \caption{{\bf OpenStreetView-5M.} Image density and proportions per country and continent for the train and test sets. To ensure an unbiased evaluation, we prioritize the uniformity of the test set's distribution across the globe over the training set distribution.}
\label{fig:data:dataset}
\end{figure*}

\hfill \break \noindent \textbf{Scale.} Deep neural networks have historically been selected over other machine learning methods because they benefit from larger amounts of data. OSV-5M consists of 4,894,685 training and 210,122 test images, with a height of $512$ pixels and an average width of $792\pm127$  pixels.

\hfill \break \noindent \textbf{Scope.} Many geolocation datasets are restricted to a few cities \cite{mirowski2019streetlearn-navigation,nn-UCF-GSV-Dataset-2014} or are significantly biased towards the Western world \cite{yfcc-analysis}. In contrast, OpenStreetView-5M images are uniformly sampled on the globe, covering 70k cities and 225 countries and territories, as shown in \figref{fig:data:dataset}. The distribution of test images across countries has a normalized entropy of $0.78$~\cite[Eq. 19]{wilcox1967indices}, suggesting high diversity. Our train set has a normalized entropy of $0.67$, which is comparable to the entropy of the distribution of the countries' area ($0.71$).

\hfill \break \noindent \textbf{Access.} OpenStreetView-5M is based on the crowd-sourced street view images of Mapillary \cite{mapillary} which follow the CC-BY-SA license: free of use with attribution~\cite{licence}.

\hfill \break \noindent \textbf{Quality Evaluation.}
We estimate through manual inspection of 4500 images that 96.1\% (±0.57\%) of the images in the OpenStreetView-5M dataset are localizable, with a 95\% confidence level \cite[Chap. 8]{illowsky2018introductory}. Among the weakly or non-localizable images, $70$\% ($2.7$\% total) are low-quality: under- or over-exposed, blurry, or rotated; $30$\% ($1.2$\% total) are poorly framed, indoor, or in tunnels.

\hfill \break \noindent \textbf{Spatial Separation.} Without carefully enforcing the spatial separation between train and test images, geolocation can reduce to place-recognition. As our goal is to assess the capacity of models to learn robust geographical representations, we ensure that no image in the OSV-5M training set lies within a $1$km radius of any image in the test set.

\hfill \break \noindent \textbf{Sequence Separation.} 
Street-view images are typically acquired by a limited number of camera sensors mounted on the top or front of a small fleet of vehicles assigned to a given region. This correlation between location, cars, and sensors can be exploited to simplify the geolocation task. Notoriously, players of the web-based geolocation game GeoGuessr \cite{geoguessr} can locate images from Ghana by spotting a piece of duct tape placed on the corner of the roof rack of the Google Street View car \cite{plonkit}. OpenStreetView-5M tries to avoid this pitfall by ensuring that no image sequence (a continuous series of images acquired by the same user) appears in both training and test sets. While this might not prevent images taken with the same vehicle on different days from being in both sets, it limits such occurrences.
%Due to the density of their acquisition, street-view images are often acquired by a limited number of camera sensors attached to face the windshield or are mounted on the top of a limited set of vehicles that are tasked to capture a certain region. This correlation between location, and cars and their sensors, which remains invariant across a certain area, can be exploited by an agent that tries to approximately localize a given image. Notoriously, players of the web-based geolocation game GeoGuessr \cite{geoguessr} can locate images from Ghana by spotting a piece of duct tape placed on the corner of the roof rack of the Google Street View car \cite{plonkit}. OpenStreetView-5M tries to avoid this pitfall by ensuring that no image sequence (a series of images acquired by the same user) appears in both training and test sets. While this might not eliminate sequences that come from the same vehicle, it significantly limits their occurrence.
%The average distance between images of the test set and their closest image in the train set is $8.9$km.

\hfill \break \noindent \textbf{Metadata.} Rich metadata beyond geographical coordinates can improve the robustness and versatility of geolocation models. Each image in our dataset is associated with four tiers of administrative data: country, region (\eg, state), area (\eg, county), and the nearest city [6]. Note that areas are not defined for one-third of the dataset.
%Applying a reverse geocoder~\cite{geocoder} to the latitude and longitude coordinates for each tag of our dataset with four tiers of administrative data: country, region (\eg, state), nearest city, and area (\eg, county), the latter being currently unavailable for one third of the dataset. % \footnote{We use \href{https://pypi.org/project/reverse_geocoder/}{\url{pypi.org/project/reverse_geocoder/}}}.
We also associate each image with a set of additional information: land cover, climate, soil type, the driving side, and distance to the sea where the image was taken. See the Appendix for more details on these attributes.

\begin{figure*}[!ht]
    \centering
    \definecolor{lighterpurple}{RGB}{191, 128, 191} \input{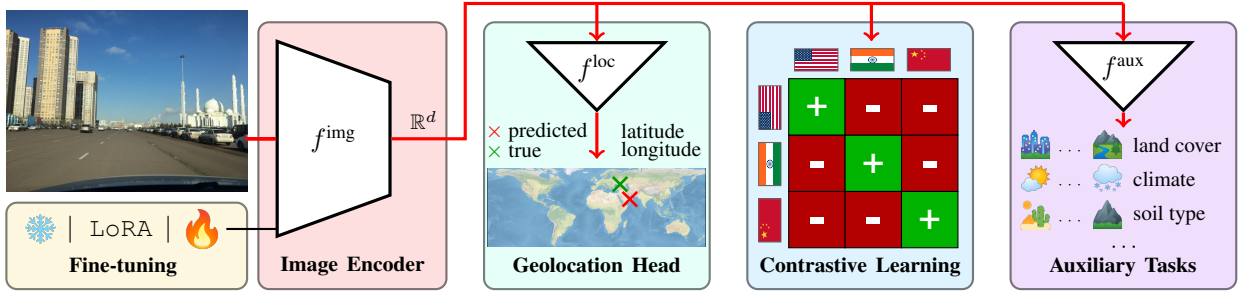}
    % \vspace*{-8pt}
    \caption{{\bf Visual Geolocation Model.} 
    We propose a simple and versatile framework for visual geolocation and explore the impact of various components of this approach in train-test performance on OpenStreetView-5M. Starting from the left, the input image is converted to a vector representation by an image encoder $f^{\text{img}}$ (\textcolor{red!70!black}{red}).
    Then a geolocation head $f^{\text{loc}}$ maps this vector to a set of geographical predictions (\textcolor{geocolorc}{mint}). 
    Then a contrastive objective is potentially added (\textcolor{blue!20!cyan}{cyan}), as well as auxiliary targets to learn better representations for geolocation (\textcolor{lighterpurple}{lila}). We also consider various parameter fine-tuning strategies for training our image encoder, by freezing all or part of $f^{\text{img}}$ (\textcolor{yellow!70!black}{yellow}).
}
\label{fig:method:pipeline}
\end{figure*}

\begin{table*}[!ht]
    \caption{{\bf Impact of Image Encoder.} Several pretrained backbones are evaluated in OpenStreetView-5M. We outline the influence of various architectures, pretraining strategies, and datasets. 
    Best scores are highlighted in \textbf{bold}. We denote closed datasets with $\dagger$.}
    \label{tab:backbone}
    \centering
    \resizebox{.85\linewidth}{!}
    {\small
    \begin{tabular}{cclllccccccc}
        \toprule
         & \multirow{2}{*}{Architecture}& \multirow{2}{*}{\makecell[b]{Size \\
         ($\times 10^6$)}}  & \multicolumn{2}{c}{Pretraining}& \multirow{2}{*}{\makecell[b]{Train. time \\ (in h)}} & \multirow{2}{*}{Geoscore $\uparrow$} & \multirow{2}{*}{Distance $\downarrow$}  & \multicolumn{4}{c}{Classification accuracy $\uparrow$} \\ 
         
        \cmidrule(lr){4-5}  \cmidrule(lr){9-12}
         % \arrayrulecolor{black!30} \cline{4-5} \arrayrulecolor{black}
        & &  & Objective & Dataset  & &  &  &  Country & Region & Area & City\\
        \midrule
       {\color{teal}\scriptsize 1} & \textcolor{BASELINECOLOR}{ViT-B-32}& \textcolor{white}{00}88 & \textcolor{BASELINECOLOR}{CLIP}& \textcolor{BASELINECOLOR}{LAION-2B} & \textcolor{white}{0}22 & 2052 & 2992 & 35.7 & \textcolor{white}{0}7.0 & 0.5 & 0.3 \\\greyrule
       \rowcolor{gray!10} {\color{teal}\scriptsize 2} & ResNet50 & \textcolor{gray!10}{00}23 & Classification & ImageNet-1k & \hphantom{0}45 & 1260 & 4171 & 20.8 & \hphantom{0}3.0 & 0.2 & 0.1\\
       {\color{teal}\scriptsize 3} & ViT-L-14& \hphantom{0}300 & DINOv2 &  DINOv2$^\dagger$ & 316 & 2530 & 2233 & 46.9 & 10.7 & 0.7 & 0.3\\
       \rowcolor{gray!10}  {\color{teal}\scriptsize 4} & ViT-L-14& \textcolor{white}{0}300 &  CLIP &  LAION-2B & 206 & 2474 & 2358 & 44.8 & 10.6 & 0.8 & 0.2\\
       {\color{teal}\scriptsize 5} & ViT-L-14& \hphantom{0}300 & CLIP &  DATA\_COMP & 206 & 2719 & 1964 & 50.6 & 12.8 & 1.0 & 0.4\\
       \rowcolor{gray!10}  {\color{teal}\scriptsize 6} & ViT-L-14& \textcolor{white}{0}300 &  CLIP &  Meta-CLIP  & 206 & 2724 & 1888 & 49.7 & 12.7 & 1.1 & 0.4\\
       {\color{teal}\scriptsize 7} & ViT-L-14& \hphantom{0}300 &  CLIP &  OpenAI$^\dagger$ & 206 & 2888 & 1688 & 53.3 & 14.6 & 1.2 & 0.5\\
       \rowcolor{gray!10} {\color{teal}\scriptsize 8} & ViT-L-14& \textcolor{white}{0}300 &  StreetCLIP  &  OpenAI$^\dagger$ + GSV$^\dagger$ & 206 & \bf 3028 & \bf 1481 & \bf 56.5 & \bf 16.3 & \bf 1.5 & \bf 0.7\\
       {\color{teal}\scriptsize 9} & ViT-bigG-14& 1800 & CLIP & LAION-2B  & 900 & 2878 & 1766 & 53.4 & 15.0 & 1.3 & 0.5\\
   
        \bottomrule
    \end{tabular}
    }
\end{table*}

%===================================================
\section{Benchmark}
%===================================================
%\setlength{\textfloatsep}{15pt plus 2pt minus 4pt}
%\setlength{\textfloatsep}{\oldtextfloatsep}
%
We use OSV-5M to benchmark supervised deep learning approaches in the context of visual geolocation. We first present our evaluation metrics (\secref{sec:evalMetric}) and framework (\secref{sec:framework}). We then explore several design choices, starting with the image encoder backbone (\secref{sec:backbone}), the prediction objective (\secref{sec:prediction}), the fine-tuning strategy (\secref{sec:training}), and the choices of contrastive losses (\secref{sec:contrast}).
In each experiment, we select the top-performing designs and integrate them into a \emph{combined model}, which we evaluate and analyze in \secref{sec:combined}.%., that are highlighted by \textcolor{FINETUNECOLOR}{orange} across the results of individual experiments.

% We use OSV-5M as a benchmark for supervised deep-learning approaches in the context of visual geolocation. In \secref{sec:baseline}, we outline our baseline model. In later sections, we explore several design choices, starting with the image encoder backbone in \secref{sec:backbone}, then the prediction objective in \secref{sec:prediction}, the optimization strategy in \secref{sec:training}, contrastive losses in \secref{sec:contrast}, and auxiliary supervision in \secref{sec:auxiliary}. We conclude in \secref{sec:combined} with an analysis of the model combining all best-performing designs. 

\subsection{Evaluation Metrics.}
\label{sec:evalMetric}

We denote the space of images by $\cI$ and the span of longitude and latitude coordinates by $\cC = [-180,180] \times [-90,90]$. Our objective is to design a model that maps an image from $\cI$ to its corresponding location in $\cC$. We measure the accuracy of predicted location across geolocation models with three complementary sets of metrics:
\begin{itemize}[itemsep=0.15em, wide, labelwidth=!, labelindent=2pt, topsep=0.15em, parsep=0em]
    \item[-] \textit{Haversine distance} \cite{van2012heavenly} $\delta$, between predicted and ground truth image locations;
    \item[-] \textit{Geoscore}, based on the famous GeoGuessr game \cite{geoguessr}, defined as $5000\exp(-\delta/1492.7)$~\cite{haas2023pigeon};
    \item[-] Accuracy of predicted locations across administrative boundaries: country, region, area, and city.
\end{itemize}

While the average distance between predictions and ground truth is sensitive to outliers (\ie, a few poor predictions can significantly undermine an otherwise high-performing algorithm), the accuracy metric based on administrative borders can avoid this issue. However, this metric can be too lenient for large divisions or arbitrarily punitive for small ones. The Geoscore offers a compromise by rewarding precise predictions without being overly sensitive to large but rare errors.

%==================================================
\subsection{Framework}
%=================================================
\label{sec:framework}

%We use a simple baseline model described below to evaluate the influence of many different machine learning techniques in the context of geolocation.
The models evaluated in this benchmark follow a consistent architecture, represented in \figref{fig:method:pipeline}. All considered networks contain the two following modules:
\begin{itemize}[itemsep=0.15em, wide, labelwidth=!, labelindent=2pt, topsep=0.15em, parsep=0em]
    \item[-] the image encoder $f^\text{img}:\cI \mapsto\bR^d$, which maps an image to a $d$-dimensional vector;
    \item[-] the geolocation head $f^\text{loc}:\bR^d \mapsto\cC$, which maps this vector to geographic coordinates.
\end{itemize}

\paragraph{Implementation details.} Unless stated otherwise, $f^\text{img}$ is always a pretrained and frozen CLIP ViT-B/32 model \cite{Radford2021LearningTV} with $d=768$ and $f^\text{loc}$ is a Multilayer Perceptron (MLP) with GroupNorms \cite{wu2018group}. This base model directly regresses geographical coordinates and uses the $L_1$ norm as loss function. The model is trained with a batch size of $512$ images for $30$ epochs ($260$k iterations) with a fixed learning rate of $2\times 10^{-4}$. 
{Throughout the paper we will denote in \textcolor{BASELINECOLOR}{blue} the frozen base model, in \textcolor{FINETUNECOLOR}{orange} its fine-tuned version, and in \textcolor{COMBINEDCOLOR}{green} the model combining all top-performing designs.}
%The resulting baseline model is denoted in \textcolor{BASELINECOLOR}{blue}.

%=============================================
\subsection{Image Encoder}
%=============================================
\label{sec:backbone}
We first benchmark various architectures for the image encoder module $f^\text{img}$, with varying  backnones, and pretraining strategies and datasets: 
%For the core component of our experimental framework, the image encoder $f^\text{img}$, we explore a wide variety of pretrained models, concentrating on the model size and datasets.
\begin{itemize}[itemsep=0.15em, wide, labelwidth=!, labelindent=2pt, topsep=0.15em, parsep=0em]
 \item[-] \textit{Architecture.} We test  a standard ResNet50~\cite{he2016deep}, and modern ViTs~\cite{dosovitskiy2020image} of multiple sizes (B-32, L-14, and bigG-14).
    \item[-] \textit{Pretraining.} We consider different types of pretraining objectives, including classification on ImageNet, self-supervized pretraining DINOv2 \cite{oquab2023dinov2}, text supervision CLIP \cite{Radford2021LearningTV}, as well as StreetCLIP ~\cite{haas2023learning}, which is finetuned specifically for geolocation. 
 \item[-] \textit{Dataset.} We consider several pretraining datasets, including LAION-2B \cite{schuhmann2022laion5b}, DATA\_COMP \cite{datacomp}, Meta-CLIP \cite{xu2023demystifying}, and the proprietary datasets of DINOv2, OpenAI, and StreetCLIP \cite{haas2023learning}.
\end{itemize}

%\vspace*{-10pt}
\paragraph{Analysis.} Our experimental results are presented in \tabref{tab:backbone}. Here, we summarize several key takeaways:
\begin{itemize}[itemsep=0.15em, wide, labelwidth=!, labelindent=2pt, topsep=0.15em, parsep=0em]
    \item[-] \textit{Model Size.} As shown in Rows 1, 2, 4, and 9 of Table~\ref{tab:backbone}, there is a direct correlation between the size of the image encoder and its geolocation performance. The large ViT, bigG-14 model with $1.8$ billion parameters (Row 9) improves significantly on the performance of its smaller versions. As the size of models correlates with their training time, we select ViT-L-14 as the best compromise.

    \item[-] \textit{Pretraining.} As seen in rows 3, 7, and 8,
    CLIP pretraining leads to better results than DINO or image classification. We thus focus on the latter for further comparisons.

    \item[-] \textit{Dataset.} Rows 4 to 8 show the significant impact of the choice of pretraining datasets. The geolocation-oriented StreetCLIP (row 8) leads to the best results, followed by OpenAI's CLIP (row 7). However, both datasets are not open access. We choose DATA\_COMP (row 5) as the best open-source dataset for its slightly better country classification rate compared to Meta-CLIP (row 6). %Choosing the correct pretraining dataset seems the biggest    
   % \item[-] \textit{Pre-training.} Comparing rows 2-4, shows the inferiority of ResNet and DinoV2 to CLIP for our task. We thus focus on the latter for further comparisons.
\end{itemize}

% Across all experiments, the choice of image encoder has the largest impact, illustrating the potential of OSV-5M to evaluate the expressivity of vision models. When combing the best designs in \secref{sec:combined}, we restrict ourselves to open models and opt for DATA\_COMP over its slightly better country classification rate compared to Meta-CLIP.

%=============================================
\subsection{Prediction Head}
%=============================================
\label{sec:prediction}
We examine three different possible supervision schemes for the geolocation head $f^\text{loc}$: regression, classification (including hierarchical classification), and a hybrid approach.

\begin{table}[t]
    \caption{{\bf Prediction Modules.} We report the performance of various prediction models and objectives. QuadTrees, hierarchical supervision, and hybrid models all significantly improve on direct regression or classification with administrative borders. We \underline{underline} the accuracy for divisions that the method is specifically trained to categorize.
    }
    \label{tab:method:prediction}
    \centering
    \resizebox{0.95\linewidth}{!}
    {
    \begin{tabular}{@{}llccccccc}
        \toprule
       & & \multirow{2}{*}{\makecell[b]{Number\\classes}} & \multirow{2}{*}{\makecell[b]{Geo $\uparrow$\\score} } & \multirow{2}{*}{\makecell[b]{Dis $\downarrow$\\tance} }  & \multicolumn{4}{c}{Classification accuracy $\uparrow$}\\
       \cmidrule(lr){6-9}
      & &  & &  & country & region & area & city \\\midrule
      \multirow{2}{*}{\rotatebox{90}{Reg.}} &
    \textcolor{BASELINECOLOR}{Coord.} & \textbf{-} &  2052 & 2992 & 35.7 & 7.0 & 0.5 & 0.3\\
       &\greycell Sin/cos& \greycell\textbf{-} & \greycell 1192 & \greycell 4797 & \greycell 13.6 & \greycell 2.1 & \greycell 0.1 & \greycell 0.0 
       \\\greyrule
      \multirow{7}{*}{\rotatebox{90}{Classification}}\!\!\!&
      Country & \hphantom{.}222\hphantom{k} & 2263 & 2981 & \underline{56.3}  & \textbf{-} & \textbf{-} & \textbf{-} \\
      & \greycell  Region & \greycell \hphantom{0}2.8k & \greycell 2683 & \greycell 2858 & \greycell 57.0 &\greycell \underline{30.2} & \greycell \textbf{-} & \greycell \textbf{-} \\
      & Area & \hphantom{0}9.3k & 1935 & 4454 & 36.3 & 19.7 &\underline{8.8} & \textbf{-} \\
      &  \greycell City & \greycell 69.8k & \greycell 2600 & \greycell 3217 & \greycell 52.2 & \greycell 28.5 & \greycell 7.3 & \greycell \underline{4.9} \\
      & + hierarchy & 69.8k & 2868 & 2768 & \underline{58.2}  &  \underline{34.3}  &  \textbf{\underline{9.6}}  &  \textbf{\underline{6.0}} \\
     & \greycell  QuadTree & \greycell 11.0k & \greycell 2772 & \greycell 2832 & \greycell 54.8 & \greycell 27.7 & \greycell 5.4 & \greycell 2.8\\
     & + hierarchy & 11.0k & 2890 & 2654 & 57.4 & 29.9 & 5.9 & 2.9\\
      \greyrule
      \rowcolor{gray!10} \multicolumn{2}{l}{Hybrid} & 11.0k & \textbf{3036} & \textbf{2518} & \textbf{60.8} & \textbf{36.3} & 9.5 & 5.7\\
      \bottomrule
    \end{tabular}}
\end{table}

\paragraph{Regression.} We start with the most straightforward approach: $f^\text{loc}$ directly regresses coordinates in $\cC$. We train an MLP supervised with the $L_1$ loss between true and predicted coordinates. To account for the periodicity of the latitude, we also test an approach where we regress instead the cosine and sine of the longitude and latitude and then recover the real coordinates with trigonometry \cite{mac2019presence}.

\paragraph{Classification.} %For classification supervision, 
We divide the train set into a set $\cK$ of $K$ \emph{divisions}, such as countries, regions, areas, and cities, which amount to $K=222$ , $2.8$k, $9.3$k, and $69.8$k, respectively. As some administrative borders can have vastly different sizes, we also consider an adaptive partition with a QuadTree of depth $10$ and maximum leaf size of $1000$, corresponding to $11$k cells.  We then train a classifier $f^\text{classif}: \bR^d \mapsto \bR^K$ which maps an image representation to the probability that the image was taken in each division. Then, to predict the final geographic location, we define $f^\text{lookup}$, which associates each division with the average location of its training images: $f^\text{lookup}:\cK \mapsto \cC$. The predicted geolocation can be summarized as: $f^\text{loc} = f^\text{lookup} \circ \argmax f^\text{classif}$. 

In our implementation, $f^\text{classif}$ is an MLP trained with cross-entropy, while $f^\text{lookup}$ is a look-up table obtained directly from the training set.   

\paragraph{Hierarchical Supervision.} 
We can exploit the nested nature of the administrative divisions and QuadTree cells to supervise all levels simultaneously \cite{Im2GPS++YFCC4k+Im2GPS3k,muller2018geolocation}. More precisely, we predict a probability vector \emph{at the finest resolution} (either city or maximum depth of the QuadTree), which we aggregate recursively to obtain predictions at all levels. We can now supervise with a cross-entropy term for each level. 

\paragraph{Hybrid Approach.} Inspired by approaches that combine both classification and retrieval \cite{haas2023pigeon,Im2GPS++YFCC4k+Im2GPS3k}, we perform regression and classification in a two-step approach. Given the output of our QuadTree classifier $f^\text{classif}: \bR^d \rightarrow \bR^K$, we define $f^\text{relative}: \bR^d \rightarrow [-1,1]^{2K}$ that outputs the relative coordinates of the predicted location inside each cell $k$. We scale these values such that $(0,0)$ points to the centroid of the training images in the cell and $[-1,1]^2$ spans the entire bounding box. 
Using the cell prediction of the classifier $f^\text{classif}$ and the relative position from $f^\text{relative}$, we can predict the location of the image with sub-cell precision.

\begin{table}[t]
    \caption{{\bf Parameter Fine-tuning Strategies.} We compare the performance of different parameter fine-tuning strategies, in terms of performance, number of parameters, and training time.}
    \label{tab:method:training}
    \centering
    %\resizebox{\linewidth}{!}{
    \resizebox{\linewidth}{!}
    {
    \begin{tabular}{@{}lcccccccc}
        \toprule
        & \multirow{2}{*}{\makecell[b]{Param.\\(M)}} & \multirow{2}{*}{\makecell[b]{Train.\\ time}} & \multirow{2}{*}{\makecell[b]{Geo $\uparrow$\\score} } & \multirow{2}{*}{\makecell[b]{Dis $\downarrow$\\tance} } & \multicolumn{4}{c}{Classification accuracy $\uparrow$}\\%\cline{2-3}
        
        \cmidrule(lr){6-9} &  &   &  & & country & region & area & city \\
        \midrule
        \textcolor{BASELINECOLOR}{Frozen} & \hphantom{0}0.6 & \hphantom{0}22 &  2052 & 2992 & 35.7 & \hphantom{0}7.0 & 0.5 & 0.3\\\greyrule
        %LoRA-16 & 1.6 & 42 & 2488 & 2181 & 42.8 & 8.6 & 0.5 & 0.2\\
        \rowcolor{gray!10} LoRA-32 & \hphantom{0}2.4 & \hphantom{0}44 & 2101 & 2760 & 36.7 & \hphantom{0}6.4 & 0.4 & 0.0\\
        Last block & \hphantom{0}7.7 & \hphantom{0}26 & 2587 & 2372 & 46.7 & 12.9 & 1.0 & 0.5\\
        \rowcolor{gray!10}\textcolor{FINETUNECOLOR}{Fine-tuning} & 88.0 & 132 & \bf 2893 & \bf 2085 & \bf 54.9 & \bf 19.1 & \bf 1.6 & \bf 0.8\\
        %scratch & 88 & -\\
        \bottomrule
    \end{tabular}}
   %}
\end{table}

We train $f^\text{classif}$ with the cross-entropy, and $f^\text{relative}$ with the $L_2$ loss between the predicted and true relative coordinates {on the division that contains the true location}.
\paragraph{Analysis.} We report the performance of different prediction heads in \tabref{tab:method:prediction}, and make the following observations:
\begin{itemize}[itemsep=0.15em, wide, labelwidth=!, labelindent=2pt, topsep=0.15em, parsep=0em]
    \item[-] {\textit{Regression.}} Predicting sines and cosines does not improve the regression model's performance. We hypothesize that this is due to the non-linearity of the trigonometric formula.
    \item[-] {\textit{Classification.}} Classification methods generally perform well in Geoscore and starkly improve their respective classification rates, \eg $+23.2\%$ region accuracy for the region classifier compared to the regression model. However, their influence on the average error distance is smaller. Coarse partitions, like countries, are limited by the low precision of $f^\text{lookup}$. Inversely, overly refined partitions such as cities lead to a more challenging classification setting where most labels have only a few training examples. QuadTree-constructed labels achieve performance close to the administrative division-based classifier across all levels, \eg $54.8\%$ \vs $56.3\%$ for countries and $27.7\%$ \vs $30.2\%$ for regions. This compounds into an overall better performance, which shows that adapting the granularity of the label distribution according to the image density appears to be a successful heuristic. 
    \item[-] \textit{Hierarchical \& Hybrid.} Supervising on all levels simultaneously significantly improves the prediction. Hybrid methods bridge the gap between classification and regression, yielding high precision without relying on very fine-grained partitions. These results validate the underlying spatial hierarchical nature of geographical data~\cite{tobler1970computer}. We select both hybrid and hierarchical designs for the combined model. 
\end{itemize}
\begin{table}[t]
    \caption{{\bf Contrastive Learning.} We report the impact of adding a contrastive objective to our model, defined by various notions of positive matches between images.}
    \label{tab:contrastive}
    \centering
    \resizebox{0.95\linewidth}{!}{
    \begin{tabular}{@{}llccccccc}
        \toprule
        & \multirow{2}{*}{Pairs} & \multirow{2}{*}{\makecell[b]{Geoscore $\uparrow$} } & \multirow{2}{*}{\makecell[b]{Distance $\downarrow$} }  & \multicolumn{4}{c}{Classification accuracy $\uparrow$}\\
        \cmidrule(lr){5-8}
        & &  &  & country & region & area & city \\
        \midrule
        \multicolumn{2}{l}{\textcolor{FINETUNECOLOR}{no contrastive}} & 2893 & 2085 & 54.9 & 19.1 & 1.6 & 0.8\\\greyrule
        \multirow{5}{*}{\rotatebox{90}{geographic}} 
        &  \greycell country & \greycell 2903 & \greycell \bf 2005 & \greycell \bf \underline{66.8} & \greycell 13.7 & \greycell 0.7 & \greycell 0.3\\
        & region & \bf 3028 & 2131 & 60.0 & \bf \underline{33.3} & 2.9 & 1.0\\
        & \greycell area & \greycell2376 & \greycell2886 & \greycell 43.7 & \greycell 18.9 & \greycell \bf \underline{3.7} & \greycell 1.2\\
        &city & 2912 & 2209 & 56.3 & 24.5 & 3.2 & \underline{1.2}\\
        &  \greycell  cell & \greycell 2891 & \greycell 2310 & \greycell 55.9 & \greycell 25.4 & \greycell 3.5 & \greycell \bf  1.3\\ \greyrule
        \multicolumn{2}{l}{text-based} & 2812 & 2171 & 66.0 & 13.0 & 0.7 & 0.2\\ 
        \bottomrule
    \end{tabular}}
\end{table} 
%Adaptive partitions, hierarchical supervision, and hybrid models significantly improve upon border-based partitions and direct regression. 
%This experiment demonstrates the complexity of the task of global geolocation---a field which has received little attention in the literature.
%=============================================
\subsection{Parameter Fine-tuning}
%=============================================
\label{sec:training}
We evaluate different fine-tuning strategies to quantify the impact of learning dedicated features for geolocation. In all configurations, we learn $f^\text{loc}$ from random weight, and $f^\text{img}$ is fine-tuned as follows:
\begin{itemize}[itemsep=0.15em, wide, labelwidth=!, labelindent=2pt, topsep=0.15em, parsep=0em]
    \item[-] \textit{Frozen.} $f^\text{img}$ is initialized with pretrained weights and remains frozen.
    \item[-] \textit{LoRA-32.} We fine-tune $f^\text{img}$ using Low Rank Adaption~\cite{hu2021lora} and a rank of $32$ (more values in supplementary).
    \item[-] \textit{Last block.} We unfreeze the last transformer block of $f^\text{img}$, responsible for producing the image embedding. %which combines the token embeddings into a single representation.
    \item[-] \textit{Fine-tuning.} We fine-tune all parameters of $f^\text{img}$.
\end{itemize}

\begin{table}[t]
    \caption{{\bf Combined Model.} We report the improvements brought by each top-performing design choice and their combination and compare them with baselines and competing approaches.}
    
    \label{tab:final}
    \centering
    \scriptsize
    \resizebox{\linewidth}{!}{
    \begin{tabular}{lcccccc}
        \toprule
         & \multirow{2}{*}{\makecell[b]{Geo $\uparrow$\\score} } & \multirow{2}{*}{\makecell[b]{Dis $\downarrow$\\tance} }  & \multicolumn{4}{c}{Classification accuracy $\uparrow$} \\
        \cmidrule(lr){4-7}
        & &  & country & region & area & city \\
        \midrule
        {\tikzmark{base}~\textcolor{BASELINECOLOR}{Base model}} & \hphantom{+}2052 & \hphantom{-}\hspace{+1.61696pt}2992 & \hphantom{+}35.7 & \hphantom{+0}7.0 & \hphantom{0+}0.5 & \hphantom{+0}0.3\\\greyrule
       \rowcolor{gray!10}  {\tikzmark{backbone}\;\;\;ViT-L-14 DC }& +\hphantom{0}667 & -\hspace{+1.61696pt}1028 & +14.9 & +\hphantom{0}5.8 & \hphantom{0}+0.5 & +\hphantom{0}0.1 \\
       {\tikzmark{head}\;\;\;QuadTree} &  +\hphantom{0}720 & -\hspace{+1.61696pt}\hphantom{0}160 & +19.1 & +20.7 & \hphantom{0}+5.4 & +\hphantom{0}2.5\\
       \rowcolor{gray!10} {\tikzmark{hybrid}\;\;\;\;\;Hybrid} & +\hphantom{0}264 & -\hspace{+1.61696pt}\hphantom{0}314 & +\hphantom{0}6.0 & +\hphantom{0}8.6 & \hphantom{0}+4.5 & +\hphantom{0}2.9\\    {\tikzmark{hierarchical}\;\;\;\;\;Hierarchical} & +\hphantom{0}118 & -\hspace{+1.61696pt}\hphantom{0}178 & +\hphantom{0}2.6 & +\hphantom{0}0.2 & \hphantom{0}+0.5 & +\hphantom{0}0.1 \\
        \rowcolor{gray!10} {\tikzmark{train}\;\;\;\textcolor{FINETUNECOLOR}{Fine-tuning}} & +\hphantom{0}841 & -\hspace{+1.61696pt}\hphantom{0}907 & +19.2 & +12.1 & \hphantom{0}+1.1 & +\hphantom{0}0.5\\
    {\tikzmark{contrast}\;\;\;\;\;Region contrast.} & +\hphantom{0}135 & +\hphantom{00}46 & 
 -\hspace{+1.61696pt}\hphantom{0}5.1 & +14.2 & \hphantom{0}+2.1 & +\hphantom{0}0.2
       \\\greyrule
        \multirow{2}{*}{\textcolor{COMBINEDCOLOR}{Combined model}} & \greycell  +1309 & \greycell -\hspace{+1.61696pt}1178 & \greycell +32.3 & \greycell +32.4 & \greycell \hphantom{0}+9.8 & \greycell +\hphantom{0}5.6\\
        & \bf \hphantom{+}3361 & \bf \hphantom{-}\hspace{+1.61696pt}1814 & \bf \hphantom{+}68.0 & \bf \hphantom{+}39.4 & \hphantom{+}10.3 & \hphantom{+0}5.9\\
        \greyrule
        %Random & 404 & 7998 & 5.0 & 0.2 & 0.0 & 0.0 \\
        \rowcolor{gray!10} Random & %\hspace{+6pt}
        \hphantom{+0}328  & %\hspace{+2pt}
        \hphantom{+}8724  & %\hspace{+6pt}
        \hphantom{+}20.0 & %\hspace{+6.9pt}
        \hphantom{+0}2.0 & %\hspace{+6.3pt}
        \hphantom{+0}0.0 & %\hspace{+5.4pt}
        \hphantom{+0}0.0 \\
        {Human Evaluation} & %\hspace{+4pt}
        \hphantom{+}1009 & %\hspace{+5pt}
        \hphantom{+}6407 & %\hspace{+5.6pt}
        \hphantom{+}48.9 & %\hspace{+5.5pt}
        \hphantom{+}12.2 & %\hspace{+6.1pt}
        \hphantom{+0}3.0 & %\hspace{+5.7pt}
        \hphantom{0+}0.0\\
         \rowcolor{gray!10} {GeoEstimator} \cite{muller2018geolocation} & 
         \hphantom{+}3331 & \hphantom{-}\hspace{+1.61696pt}2308 & \hphantom{+}66.8 & \hphantom{+}\bf39.4 & \hphantom{+}\bf18.4 & \hphantom{+0}4.2 \\
        {StreetCLIP 0-shot \cite{haas2023learning}} & \hphantom{+}2273 & \hphantom{-}\hspace{+1.61696pt}2854 & \hphantom{+}38.4 & \hphantom{+}20.8 & \hphantom{0+}9.9 & \bf \hphantom{+}14.8 \\
        \bottomrule
    \end{tabular}
    \begin{tikzpicture}[overlay,remember picture]
    \draw[black,thick] ([yshift=1mm]pic cs:base) -- ([yshift=1mm]pic cs:backbone) -- ([xshift=2.0mm,yshift=1mm]pic cs:backbone) ;
    \draw[black,thick] ([yshift=1mm]pic cs:backbone) -- ([yshift=1mm]pic cs:head) -- ([xshift=2.0mm,yshift=1mm]pic cs:head) ;
    \draw[black,thick] ([xshift=1.25mm,yshift=1mm]pic cs:head) -- ([xshift=1.25mm,yshift=1mm]pic cs:hybrid) -- ([xshift=3.5mm,yshift=1mm]pic cs:hybrid) ;
    \draw[black,thick] ([xshift=1.25mm,yshift=1mm]pic cs:hybrid) -- ([xshift=1.25mm,yshift=1mm]pic cs:hierarchical) -- ([xshift=3.5mm,yshift=1mm]pic cs:hierarchical) ;
    \draw[black,thick] ([yshift=1mm]pic cs:head) -- ([yshift=1mm]pic cs:train) -- ([xshift=2.0mm,yshift=1mm]pic cs:train) ;
    \draw[black,thick] ([xshift=1.25mm,yshift=1mm]pic cs:train) -- ([xshift=1.25mm,yshift=1mm]pic cs:contrast) -- ([xshift=3.5mm,yshift=1mm]pic cs:contrast) ;
\end{tikzpicture}
    }
\end{table} 
\paragraph{Analysis.} In \tabref{tab:method:training}, we report the impact of different fine-tuning strategies. Training only the last transformer block instead of using LoRA leads to a ten times larger Geoscore improvement in only half the training time. This suggests that pretrained
models can extract relevant patch embeddings, while image encoding must be significantly adapted for geolocation. Fine-tuning the entire network leads to an even larger improvement but a five-fold increase in training time. However, the resulting performance is comparable to the frozen ViT-bigG-14 shown in Table~\ref{tab:backbone} and trains 9 times faster. We select the fine-tuning configuration as the top-performing approach and denote it in \textcolor{FINETUNECOLOR}{orange}.

% \paragraph{Auxiliary Tasks.} We use an MLP $f^\text{aux}$ to predict the image metadata in addition to its coordinates in the fine-tuning setting. While this model is capable of predicting complex geographic values from only images, its impact on geolocation remains modest: $+17$ geoscore. We argue that the size of OpenStreeView-5M allows the model to learn rich latent variables without auxiliary supervision. We provide further details on this experiment in the Appendix.

%However, training the network from a random initial state does not lead to comparable performance, suggesting that the supervisory signal of geolocation is too noisy and complex to train a network from scratch.

%=============================================
\subsection{Contrastive Objectives}
%=============================================
\label{sec:contrast}

Contrastive learning builds positive and negative sample pairs from the training set and pushes representations of positive pairs close to each other while contrasting negative ones \cite{chen2020simple,chopra2005learning}. Positive pairs can be formed within the same modality, such as different views of an object, or across modalities, such as images and captions. In the geolocation context, we propose two approaches to construct such pairs:
\begin{itemize}[itemsep=0.15em, wide, labelwidth=!, labelindent=2pt, topsep=0.15em, parsep=0em]
    \item[-] \textit{Geographic.} We match images if they are within the same administrative division: countries, regions, areas, cities, or QuadTree cells. We modify the dataloader to ensure each image is part of at least one positive pair. %It is worth noting that an image can be involved in multiple positive pairs within the same batch. 
    Contrary to Haas \etal \cite{haas2023learning}, we use the multi-positive MIL-NCE loss \cite{miech2020end} as our contrastive objective to account for images in several positive pairs, \eg in the same country. 
    \item[-] \textit{Text-Based.} Similar to Haas \etal \cite{haas2023learning}, we pair each image with a textual description of its location formed as the following string: ``An image of the city of \$CITY, in the area of \$AREA, in the region of \$REGION, in \$COUNTRY.''.
\end{itemize}
%  We adopt the MIL-NCE loss \cite{miech2020end} as our contrastive objective, which extends the InfoNCE loss \cite{oord2018representation} to cases where each sample can have multiple positive matches.

\if 1 0
\begin{align}
\sum_{i \in \mathcal{B}}
    \log\!\!\left(
    \frac{
        \displaystyle{\sum_{p \in \mathcal{P}_i}}
        \!
            e^{
            f^{\text{img}}(i)^\intercal 
            f^{\text{img}}(p)
            /T
            }
    }
    {
        \displaystyle{\!\sum_{p \in \mathcal{P}_i}}
        \!
            e^{
            f^\text{img}(i)^\intercal 
            f^\text{img}(p)/T
            }
            \!\!+\!\!\!\!\!\!
        \displaystyle{\sum_{n \in \mathcal{B} \setminus \mathcal{P}_i}}\!\!\!\!
            e^{
            f^\text{img}(i)^\intercal 
            f^\text{img}(n)
            /T
            }
    }
    \!\!\right),
\end{align}
with $\mathcal{P}_i \subset \mathcal{B}$ the set of image positively paired with $i$ and $T$ a temperature parameter set as $0.1$. If an image has only one positive match, this equation is reduced to the InfoNCE loss \cite{oord2018representation}.
\fi

\paragraph{Analysis.} In \tabref{tab:contrastive}, we measure the impact on the \textcolor{FINETUNECOLOR}{fine-tuned model} of different approaches for constructing contrastive pairs. 
We observe a consistent improvement in terms of performance when building positive pairs with regions, which may be the division most likely to present unique and homogeneous visual and cultural identities. In contrast, areas appear to hurt the performance when used contrastively. 
Overall, contrastive learning yields a much higher country and region classification rate compared to the classification-based approaches of Table \ref{tab:method:prediction}, suggesting that encouraging geographically consistent representations is advantageous for geolocation.
We also observe that using text as a proxy when geographically consistent pairs exist is not beneficial.

\begin{table}[t]
    \caption{{\bf Nearest Neighbors.} We report the performance of nearest neighbor retrieval using different encoders.}
    \label{tab:final_nn}
    \centering
    \resizebox{\linewidth}{!}{
    \begin{tabular}{lcccccc}
        \toprule
        & \multirow{2}{*}{\makecell[b]{Geo $\uparrow$\\score} } & \multirow{2}{*}{\makecell[b]{Dis $\downarrow$\\tance} }  & \multicolumn{4}{c}{Classification accuracy $\uparrow$} \\
        \cmidrule(lr){4-7}
        & &  & country & region & area & city \\
        \midrule
        CLIP-VIT-B32-LAION & 2511 & 3455 & 49.3 & 29.6 & \hphantom{0}1.9 & 13.1 \\
        %CLIP-VIT-B32-OpenAI & 2983 & 2582 & 57.9 & 37.3 & 24.7 & 18.0 \\
        \rowcolor{gray!10} DINOv2 & 2994 & 2542 & 61.1 & 37.1 & 22.9 & 16.4 \\
        CLIP-VIT-L14-DATACOMP & 3201 & 2047 & 64.5 & 38.4 & 23.3 & 16.6 \\
        \rowcolor{gray!10} CLIP-VIT-L14-OpenAI & 3545 & 1458 & 72.8 & 44.4 & 27.5 & 19.3 \\
        StreetCLIP & \bf 3597 & \bf 1386 & \bf 73.4 & \bf 45.8 & \bf 28.4 & \bf 19.9  \\ \greyrule
        %MBase model \\
%         Accuracy:  (country), 53.75 (region), 37.12 (sub-region), 26.70 (city)
% Haversine: 1028.88 (haversine),  (geoguessr)
        \rowcolor{gray!10} \textcolor{COMBINEDCOLOR}{Combined  model} &2734& 2608 & 54.9 & 24.5 & 13.6 & 9.4\\
        \bottomrule
    \end{tabular}
    }
\end{table} 

\if 1 0

\begin{figure}
    \centering
    \includegraphics[width=\linewidth, height=.15\textheight]{example-image}
    \caption{{Contrastive Training.} Similarity heatmap for Ireland.}
    \label{fig:enter-label}
\end{figure}
\fi
\if 1 0

%=============================================
\subsection{Auxiliary Tasks}
%=============================================
\label{sec:auxiliary}
We use an MLP $f^\text{aux}$ to predict the image's metadata in addition to its coordinates. All categorical variables are supervised with unweighted cross entropy terms, while the distance to the sea is supervised with the L1 norm.

\paragraph{Analysis.} Adding auxiliary tasks encourages the model to focus on relevant geographical cues. In practice, we observe a modest impact. However, our model is able to perform accurate predictions for complex geographic variables.

\begin{table}[t]
    \caption{{\bf Auxiliary Variables.} We report the impact on geolocation performance of adding various auxiliary prediction classes. We also report the performance on the test set for each auxiliary variable as the overall accuracy or the average error.}
    \label{tab:auxiliary}
    \centering
    %\resizebox{\linewidth}{!}{
    \scriptsize
    \begin{tabularx}{\linewidth}{@{}lXXXcXXXX@{}}
        \toprule
        &n of&perf & geo- & avg & \%@ & \%@ & \%@ & \%@ \\
        &classes &\multicolumn{1}{c}{\%} & score & dist $\downarrow$ & country & region & area & city \\
        \midrule
        \textcolor{FINETUNECOLOR}{no auxiliary} & - & - & 2893 & 2085 & 54.9 & 19.1 & 1.6 & 0.8\\\greyrule
        land cover& 11 & & \\
        climate& 31 & 58.3 & 2898 & 2022 & 53.7 & 18.8 & 1.7 & 0.8\\
        soil type &15 & 47.7 & 2826 & 2111 & 52.4 & 17.6 & 1.5 & 0.7\\
        driving side &1 & 94.6 & 2896 & 2025 & 54.5 & 18.7 & 1.6 & 0.7 \\
        dist to sea & - & 543km & 2870 & 2053 & 52.5 & 18.7 & 1.5 & 0.7\\\greyrule
        all & - & - & 2910 & 1987 & 54.0 & 19.8 & 1.6 & 0.8\\\bottomrule
    \end{tabularx}
    %}
\end{table} 
\fi
%=============================================
\subsection{Combined Model}
%=============================================
\label{sec:combined}

Summarizing our previous exploration and analysis, we combine the most impactful design choices for each experiment into a strong geolocation model, denoted in \textcolor{COMBINEDCOLOR}{green}: ViT-L-14 backbone pretrained on DATA\_COMP, QuadTree partition with hybrid prediction and hierarchical supervision, fully fine-tuned with a region-contrastive loss.
As shown in \tabref{tab:final}, this model starkly improves on the \textcolor{BASELINECOLOR}{base model}, with an increase of $+1309$ in Geoscore, an average distance reduced by $45$\%, and significantly better accuracy at all levels of administrative divisions.

\paragraph{Analysis.} 
In \tabref{tab:final}, we compare the performance of our combined model to a random baseline (select the location of a random image in the training set) and a human performance obtained by asking $80$ annotators to guess the locations of the same $50$ images randomly sampled from the test set \cite{mehta2016exploratory}.%, which can be accessed \href{https://huggingface.co/spaces/osv5m/plonk}{here}.
Despite the difficulty of the task, the average annotator's performance is significantly better than chance. Our baseline model, and more substantially our combined model, far surpasses the accuracy of annotators.
We also evaluate two state-of-the-art geolocation models: StreetCLIP \cite{haas2023learning} evaluated in zero-shot using the text string given in \secref{sec:contrast}, and the GeoEstimator model~\cite{muller2018geolocation} fine-tuned on our training set.
As both models are designed for geolocation, they yield good performance. Owing to its bespoke geocells, GeoEstimator reaches the highest accuracy for area classification, illustrating the benefit of architectures with built-in geographical priors. See the appendix for further experiments, notably on the impact of auxiliary variables.

 \paragraph{Nearest Neighbor.} We perform retrieval by matching each image from the test set with an image from the train set based on the cosine distance between the features of each image encoder.
 We perform approximate matching with the FAISS algorithm \cite{faiss} through the AutoFAISS package \cite{autofaiss}, without re-ranking \cite{qin2011hello,jegou2007contextual}. 
 % Performing retrieval with our baseline model for the entire test set takes $3$ hours and requires $19.1$~GB of simultaneous storage. In comparison, the inference time of our combined model is $80$~min and its size is $1.2$~GB.
\begin{figure}[!ht]
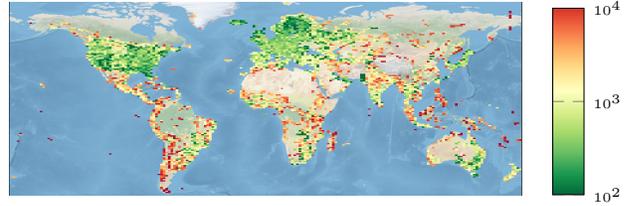

    \centering
    \include{./figures/errormap}
    \vspace{-5mm}
    \caption{{\bf Spatial  Distribution of Errors.} We plot the average prediction error of the combined model in km across the globe.
    }
    \label{fig:final:distribution}
\end{figure}
As reported in Table~\ref{tab:final_nn}, retrieval methods trained through contrastive learning exhibit high performance. However, the supervision of our combined model based on geographic coordinates and cells does not enhance its retrieval performance. In fact, its retrieval score is lower than that of its pretrained image encoder. These findings are consistent with observations that fine-tuning self-supervised models decreases retrieval performance \cite{wortsman2022robust}.
\if 1 0
 \begin{figure}[t]
    \centering
    \begin{tikzpicture}
\begin{axis}[
    xlabel={\footnotesize Number of training images ($\times 10^6$)},
    ylabel={Geoscore},
    xmin=0, xmax=5000,
    ymin=3150, ymax=3300,
    xtick={100, 200, 500, 1000, 2000, 5000},
    xticklabels={0.1, , 0.5, 1, , 5},
    ytick={3150, 3300},
    legend pos=north west,
    ymajorgrids=true,
    grid style=dashed,
    height=4cm,
    width=\linewidth,
    ylabel style={at={(axis description cs:0.0,.5)},rotate=0}, 
    xlabel style={at={(axis description cs:0.55,-0)}}, % Adjust y-axis label position
    legend style={at={(0.43,0.4)}}
]

\addplot[
    color=blue,
    mark=square,
    very thick,
    smooth,
    ]
    coordinates {
    (100,3179)(200,3201)(500,3188)(1000,3209)(2000,3213)(5000,3286)
    };
    \addlegendentry{\footnotesize \textcolor{FINETUNECOLOR}{No auxiliary supervision}}

\addplot[
    color=red,
    mark=o,
    smooth,
    very thick,
    ]
    coordinates {
    (100,3209)(200,3175)(500,3210)(1000,3197)(2000,3289)(5000,3273)
    };
    \addlegendentry{\footnotesize Auxiliary supervision}

\end{axis}
\end{tikzpicture}
    % \vspace{-3mm}
    \caption{{\bf Auxiliary Supervision.} We report the performance with and without auxiliary supervision for training sets of different sizes. }
    \label{fig:aux}
\end{figure}
\fi

\paragraph{Error Distribution.}
We report in \figref{fig:final:distribution} a heatmap of the average error distance. Areas sparsely populated with training images, such as South America, have a significantly higher error rate. We report a Pearson correlation coefficient of $-0.25$ between image density and error, suggesting that image density is not the only factor in the mistakes of our proposed model. See \figref{fig:final} for a visualization of the error distribution. Over half of the combined model's predictions are within 250km of the true image locations.

\if 1 0
\fi

\if 1 0
\begin{table}[t]
    \caption{{\bf Annotator Performance.} We report the average performance of $80$ annotators on a subset of $50$ images.}
    \label{tab:humans}
    \centering
    \resizebox{\linewidth}{!}{
    \scriptsize
    \begin{tabular}{lccccc}
        \toprule
        & \multirow{2}{*}{\makecell[b]{Geo $\uparrow$\\score} } & \multirow{2}{*}{\makecell[b]{Dis $\downarrow$\\tance} }  & \multicolumn{3}{c}{Classification accuracy $\uparrow$} \\
        \cmidrule(lr){4-6}
        &   &  & continent & country & region \\\midrule
        
        Annot. performance & 1009 & \hphantom{0}6407 & 48.9 & 12.2 & \hphantom{0}3.0\\
         \rowcolor{gray!10} Annot. ensemble oracle & \bf 3919 & \bf  \hphantom{00}443 & \bf  98.0 & \bf 70.0 & 28.0\\ \greyrule
        Random location & \hphantom{0}120  & 10273   & 16.0 & \hphantom{0}0.0 & \hphantom{0}0.0\\
        \rowcolor{gray!10}  Random image & \hphantom{0}328   & \hphantom{0}8724 & 20.0 & \hphantom{0}2.0 & \hphantom{0}0.0 \\\greyrule
        \textcolor{BASELINECOLOR}{Base model} & 2235 & \hphantom{0}3247 & 74.0 & 36.0 & \hphantom{0}8.0\\
        \rowcolor{gray!10} 
 \textcolor{COMBINEDCOLOR}{Combined model} & 3333 & \hphantom{0}1948 & 86.0 & \bf 70.0 & \bf 34.0\\\bottomrule
    \end{tabular}
    }
\end{table} 
\fi
% \begin{table}[t]
%     \caption{{\bf Annotator Performance.} We report the average performance of $80$ annotators on a subset of $50$ images.}
%     \label{tab:humans}
%     \centering
%     %\resizebox{\linewidth}{!}{
%     \scriptsize
%     \begin{tabularx}{\linewidth}{@{}lccccc@{}}
%         \toprule
%         & \multicolumn{1}{l}{geo-} & \multicolumn{1}{l}{avg} & \multicolumn{1}{l}{\%@} & \multicolumn{1}{l}{\%@} & \multicolumn{1}{l}{\%@} \\
%         & \multicolumn{1}{l}{score} & \multicolumn{1}{l}{dist $\downarrow$} & \multicolumn{1}{l}{continent} & \multicolumn{1}{l}{country} & \multicolumn{1}{l}{region} \\\midrule
        
%  Annotator performance & 1009 & 6407 & 48.9 & 12.2 & 3.0\\
%          \rowcolor{gray!10} Annotator ensemble oracle & \bf 3919 & \bf  443 & \bf  98.0 & \bf 70.0 & 28.0\\ \greyrule
%         Random location & 120  & 10273   & 16.0 & 0.0 & 0.0\\
%         \rowcolor{gray!10}  Random image & 328   & 8724 & 20.0 & 2.0 & 0.0 \\\greyrule
%         \textcolor{BASELINECOLOR}{Base model} & 2235 & 3247 & 74.0 & 36.0 & 8.0\\
%         \rowcolor{gray!10} 
%  \textcolor{COMBINEDCOLOR}{Combined model} & 3333 & 1948 & 86.0 & \bf 70.0 & \bf 34.0\\\bottomrule
%     \end{tabularx}
%     %}
% \end{table} 
%\setlength{\textfloatsep}{\oldtextfloatsep}

%\if 1 0

%\fi
\if 1 0

\begin{figure}
    \centering
    \begin{tikzpicture}
\begin{polaraxis}[
    yticklabel style={font=\tiny},
    ytick={ 20,  40,  60, 80},
    yticklabels={20, 40, 60, 80},
    xtick={0, 72, 144, 216, 288},
    xticklabels={, , , , },
    %extra x tiks = {0,24,48}
    grid=both,
    ymin=0,
    ymax=100,
    ymajorgrids=true,
    line width=1pt,
    width=0.45\textwidth,
    height=0.45\textwidth,
    clip=false,
    legend style={,font=\scriptsize, at={(0,0)}},
    legend cell align={left},
]

\node [anchor=center] at (36,110) {\rotatebox{306}{Img2GPS}};
\node [anchor=center] at (108,110) {\rotatebox{18}{Img2GPS3k}};
\node [anchor=center] at (180,110) {\rotatebox{90}{YFCC4k}};
\node [anchor=center] at (256,110) {\rotatebox{346}{YFCC26k}};
\node [anchor=center] at (324,110) {\rotatebox{54}{GWS15k}};

\draw[thick, black] (0,0) -- (72,100);
\draw[black!30] (24,0) -- (24,100);
\draw[black!30] (48,0) -- (48,100);
\draw[thick, black] (0,0) -- (72,100);
\draw[black!30] (0,0) -- (96,100);
\draw[black!30] (0,0) -- (120,100);
\draw[thick, black] (0,0) -- (144,100);
\draw[black!30] (0,0) -- (169,100);
\draw[black!30] (0,0) -- (192,100);
\draw[thick, black] (0,0) -- (216,100);
\draw[black!30] (0,0) -- (240,100);
\draw[black!30] (0,0) -- (264,100);
\draw[thick, black] (0,0) -- (288,100);
\draw[black!30] (0,0) -- (312,100);
\draw[black!30] (0,0) -- (336,100);

\node [anchor= center] at (291,89) {\rotatebox{291}{\scriptsize 25km}};
\node [anchor= center] at (315,88) {\rotatebox{315}{\scriptsize 200km}};
\node [anchor= center] at (339,88) {\rotatebox{339}{\scriptsize 750km}};
\node [anchor= center] at (357,85) {\rotatebox{357}{\scriptsize 2000km}};

\addplot+[mark=none, blue, thick] coordinates {
    (0, 48.1) %IM2GPS 25km
    (24,64.6) %IM2GPS 200km
    (48,75.6) %IM2GPS 750km
    (72, 86.7) %IM2GPS 2000km
    (72, 31.1) %IM2GPS3k 25km
    (96,46.7) %IM2GPS3k 200km 
    (120,58.9) %IM2GPS3k 750km
    (144, 80.1) %IM2GPS3k 2000km
    (144,18.6) %YCC4k 25km
    (168,27) %YCC4k 200km
    (192,41.1) %YCC4k 750km
    (216, 60.4) %YCC4k 2000km
    (216, 17.8) %YCC26k 25km
    (240,28) %YCC26k 200km
    (264,41.3) %YCC26k 750km
    (288, 60.6) %YCC26k 2000km
    (288,1.1) %GSW15k 25km
    (312,8.0) %GSW15k 200km
    (336,25.5) %GSW15k 750km
    (360,48.3) %GSW15k 2000km
};

\addplot+[mark=none, red, thick] coordinates {
    (0, 50.2) %IM2GPS 25km
    (24,69.0) %IM2GPS 200km
    (48,80.0) %IM2GPS 750km
    (72, 89.1) %IM2GPS 2000km
    (72, 33.5) %IM2GPS3k 25km
    (96,45.9) %IM2GPS3k 200km 
    (120,61.0) %IM2GPS3k 750km
    (144, 76.1) %IM2GPS3k 2000km
    (144,24.4) %YCC4k 25km
    (168,33.9) %YCC4k 200km
    (192,50) %YCC4k 750km
    (216, 68.7) %YCC4k 2000km
    (216, 23.9) %YCC26k 25km
    (240,34.1) %YCC26k 200km
    (264,49.6) %YCC26k 750km
    (288, 69) %YCC26k 2000km
    (288,1.5) %GSW15k 25km
    (312,8.8) %GSW15k 200km
    (336,26.9) %GSW15k 750km
    (360,50.5) %GSW15k 2000km
};

\addplot+[mark=none, green, thick] coordinates {
    (0, 0.03) %IM2GPS 25km
    (24, 2.7) %IM2GPS 200km
    (48, 20.4) %IM2GPS 750km
    (72, 52.8) %IM2GPS 2000km
    (72, 0.03) %IM2GPS3k 25km
    (96, 2.5) %IM2GPS3k 200km 
    (120, 19.65) %IM2GPS3k 750km
    (144, 51.98) %IM2GPS3k 2000km
    (144, 0.04) %YCC4k 25km
    (168, 2.06) %YCC4k 200km
    (192, 15.9) %YCC4k 750km
    (216, 44.13) %YCC4k 2000km
    % (216, 23.9) %YCC26k 25km
    % (240,34.1) %YCC26k 200km
    % (264,49.6) %YCC26k 750km
    % (288, 69) %YCC26k 2000km
    % (288,1.5) %GSW15k 25km
    % (312,8.8) %GSW15k 200km
    % (336,26.9) %GSW15k 750km
    % (360,50.5) %GSW15k 2000km
};

\legend{Translocator \cite{translocator}, Clark \cite{Google-World-Streets-15K-geolocation}, Ours}
\end{polaraxis}
\end{tikzpicture}
    \caption{{\bf Evaluation on other Datasets.} We evaluate our base and best models on 5 public geolocation benchmarks. We report the proportion of predictions within 25km, 200km, 750km, and 2000km of the true image position.}
    \label{fig:spider}
\end{figure}
\fi

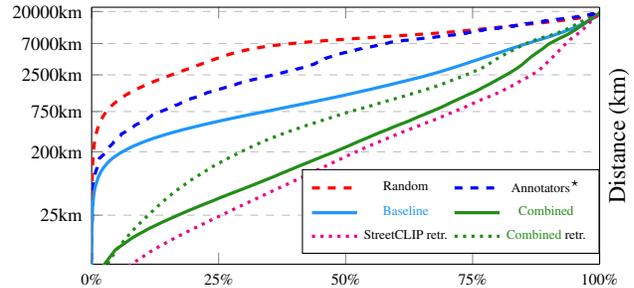
\begin{figure}[!h]
    \pgfplotstableread[col sep=comma]{./data/histogram_base.csv}\datatablebaseline
\pgfplotstablecreatecol[
    create col/expr={
        \pgfmathaccuma + \thisrow{density}
    }
]{cumulative}{\datatablebaseline}

\pgfplotstableread[col sep=comma]{./data/histogram_random.csv}\datatablerandom
\pgfplotstablecreatecol[
    create col/expr={
        \pgfmathaccuma + \thisrow{density}
    }
]{cumulative}{\datatablerandom}
   
\pgfplotstableread[col sep=comma]{./data/histogram_humans.csv}\datatablehumans
\pgfplotstablecreatecol[
    create col/expr={
        \pgfmathaccuma + \thisrow{density}
    }
]{cumulative}{\datatablehumans}   

\pgfplotstableread[col sep=comma]{./data/histogram_best.csv}\datatablecombined
\pgfplotstablecreatecol[
    create col/expr={
        \pgfmathaccuma + \thisrow{density}
    }
]{cumulative}{\datatablecombined}   

\pgfplotstableread[col sep=comma]{./data/histogram_streetclip_retrieval.csv}\datatablestreetclip
\pgfplotstablecreatecol[
    create col/expr={
        \pgfmathaccuma + \thisrow{density}
    }
]{cumulative}{\datatablestreetclip}   

\pgfplotstableread[col sep=comma]{./data/histogram_best_retrieval.csv}\datatablebestretrieval
\pgfplotstablecreatecol[
    create col/expr={
        \pgfmathaccuma + \thisrow{density}
    }
]{cumulative}{\datatablebestretrieval}   

    \centering
    \begin{tikzpicture}
    \begin{axis}[
    ymode=log,
    ylabel={\small Distance (km)},
    ylabel style={at={(axis description cs:1.00,0.5)}, anchor=north},
    legend style={at={(1,0.37)}},
    legend columns=2, 
    ytick={25, 200, 750, 2500, 7000, 20000}, % Specified y-ticks
    yticklabels={25km, 200km, 750km,  2500km, 7000km, 20000km}, % Custom y-tick labels
    xtick={0, 0.25, 0.5, 0.75, 1}, % Specified x-ticks
    xticklabels={0\%, 25\%, 50\%, 75\%, 100\%}, % Custom x-tick labels
    xmin=0, xmax=1, % x-axis limits
    ymin=5, ymax = 23000, % y-axis minimum
    enlargelimits=0.00,
    font=\fontsize{6pt}{6pt}\selectfont,
    ymajorgrids=true, % to add major grid lines on the y-axis
    grid style=dashed,
    yticklabel style={
        font=\scriptsize,
        xshift=0cm
    },
    axis y line*=left, % Keeps y-ticks on the left
    yticklabel pos=left, % Puts y-tick labels on the left
    height=5cm,
    width=\linewidth,
     after end axis/.code={
        \draw (rel axis cs:1,0) -- (rel axis cs:1,1);
    }
]

\addplot [no markers, red, very thick, dashed] table[x=cumulative, y=bin] {\datatablerandom};

\addplot [no markers, blue, very thick, dashed] table[x=cumulative, y=bin] {\datatablehumans};

\addplot [no markers, BASELINECOLOR, very thick] table[x=cumulative, y=bin] {\datatablebaseline};

\addplot [no markers, COMBINEDCOLOR, very thick] table[x=cumulative, y=bin] {\datatablecombined};

\addplot [no markers, magenta, very thick, dotted] table[x=cumulative, y=bin] {\datatablestreetclip};

\addplot [no markers, COMBINEDCOLOR, very thick, dotted] table[x=cumulative, y=bin] {\datatablebestretrieval};

\legend{
\tiny Random,
\tiny Annotators$^\star$,
\tiny \textcolor{BASELINECOLOR}{Baseline},
\tiny \textcolor{COMBINEDCOLOR}{Combined},
\tiny StreetCLIP retr.,
\tiny \textcolor{COMBINEDCOLOR}{Combined} retr.
}

\end{axis}
\end{tikzpicture}
    \vspace{-7mm}
    \caption{{\bf Error Distribution.} Proportion of predictions within a set distance in the test set.  $^\star$ evaluated on $50$ images only.}
    \label{fig:final}
\end{figure}
%===================================================
\section{Conclusion}
%===================================================
We introduced a new open-access street view dataset of unprecedented size and quality, enabling the consistent training and evaluation of global geolocation models for the first time. Through an extensive experimental framework, we demonstrate that our dataset is a competitive benchmark for developing and evaluating general and bespoke state-of-the-art computer vision approaches for geolocation. Through its scale and quality, we expect OSV-5M to also be useful for self-supervised learning and generative modeling, valuable tasks beyond the scope of visual geolocation.

\paragraph*{Acknowledgements.} 
OSV-5M was made possible through the generous support of the Mapillary team, which helped us navigate their vast street view image database. Our work was supported by the ANR project READY3D ANR-19-CE23-0007, and the HPC resources of IDRIS under the allocation AD011014719 made by GENCI. We thank Valérie Gouet for her valuable feedback.
\pagebreak\pagebreak

\FloatBarrier
%\pagebreak
%\pagebreak
{\small
%\balance
\bibliographystyle{templates/ICCV/ieee_fullname}
\bibliography{mybib}

}

\ARXIV{
    \FloatBarrier
    \pagebreak
    %\balance
    %\section*{\centering \LARGE Appendix}
    \setcounter{section}{0}
    \setcounter{figure}{0}
    \setcounter{table}{0}
    \renewcommand{\thesection}{\Alph{section}}
    \renewcommand{\thefigure}{\Alph{figure}}
    \renewcommand{\thetable}{\Alph{table}}
        \FloatBarrier
        \pagebreak
%\appendix
% Appendix
\twocolumn[\centering
  \section*{\Huge Appendix} % Use \section*{} for unnumbered section, which spans across both columns
   % Center the title; optional if you want the title itself centered
]

    This supplementary material starts by providing further details on the construction and analysis of our dataset OpenStreetView-5M in \secref{sec:sup:construction}, showcasing indicative samples in \figref{fig:sup:illu}. Then, we provide additional experiments in \secref{sec:sup:xp} and qualitative results in \figref{fig:sup:answer}. Finally, \secref{sec:sup:implem} further implementation details can be found and \secref{sec:datasheet} outlines our Datasheet~\cite{gebru2021datasheets} for OpenStreetView-5M.
%===============================================================
\section{OpenStreetView-5M Dataset}
%===============================================================
\label{sec:sup:construction}  
OpenStreetView-5M is designed to achieve an open, large-scale, balanced, and global geographical coverage. Through the  Mapillary API and the support of the Mapillary team, we gained access to the locations of all $1.8$B images \cite{mapillary}. To provide a more manageable and better distributed dataset,  we design a specific construction approach, presented in this section. The code to reproduce the treatment can be found at \GITHUB.

\subsection{Construction Approach}

\paragraph{Sampling.} We start by ensuring that regions with high image density are not disproportionately represented. We define a $100\times100$m grid across the entire world and randomly choose one image per cell. Then, both the training and test sets are sampled with a weight proportional to the local image density raised to the power of $-0.75$. Such a strategy balances density-based sampling (which tends to be biased towards urban centers) and area-based sampling (which might favor larger countries). We eliminate images from the test set that are either located within a 1km radius of any train image or share a sequence ID.

\paragraph{Handcrafted Filters.}
We apply a series of handcrafted filters to remove low-quality images
\begin{itemize}
[itemsep=0.25em, wide, labelwidth=!, labelindent=2pt, topsep=0.25em, parsep=0em]
\item[-] \textit{Blurriness.} Blurry images indicate low quality and potentially low localizability. We remove images whose average logarithmic magnitude spectrum is below $120$dB.
\item[-] \textit{Radiometry.} %Unlike Google Street View, 
Certain images hosted on Mapillary are too dark to be meaningfully analyzed, while other have a distinct encoding errors giving them a purple tint. To remove those, we first filter out images whose average brightness (average value over pixels and RGB channels) is below $50$. To handle purple images, we remove images for which over $50\%$ of pixels meet the following criteria: $R>60 \;\& \;G>60 \;\&\; B<50$.
\item[-] \textit{Exposition.} The exposure of Dash-cam images can be badly exposed, for example, when they face the sun. To filter them, we remove images for which $70$\% of pixels have a brightness over $250$ (overexposed) or under $5$ (underexposed).
\end{itemize}

\renewcommand\thesubfigure{\arabic{subfigure}}
\begin{figure*}
    \centering
    \begin{tabular}{c@{\;}c@{\;}c@{\;}c}
    \begin{subfigure}{0.24\linewidth}
    \includegraphics[width=1\linewidth, height=.10\textheight]{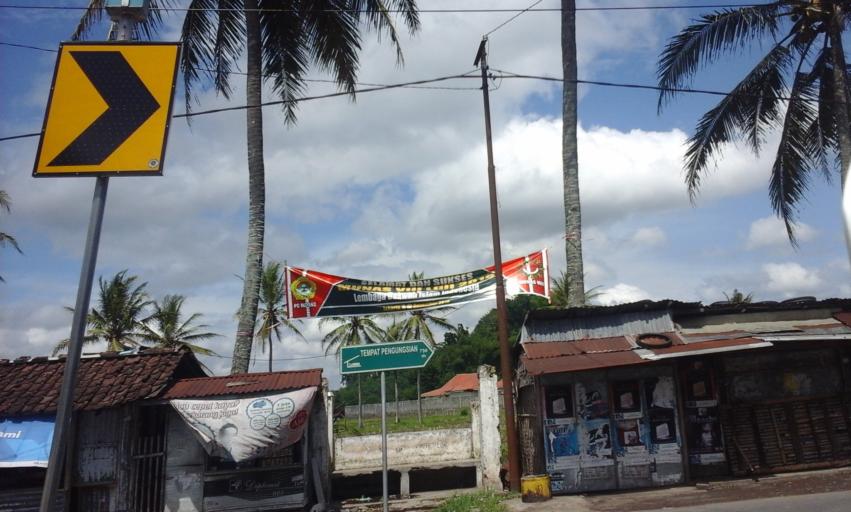}
    \caption{arizalkawamuna}
    \end{subfigure}
    &
    \begin{subfigure}{0.24\linewidth}     \includegraphics[width=1\linewidth, height=.10\textheight]{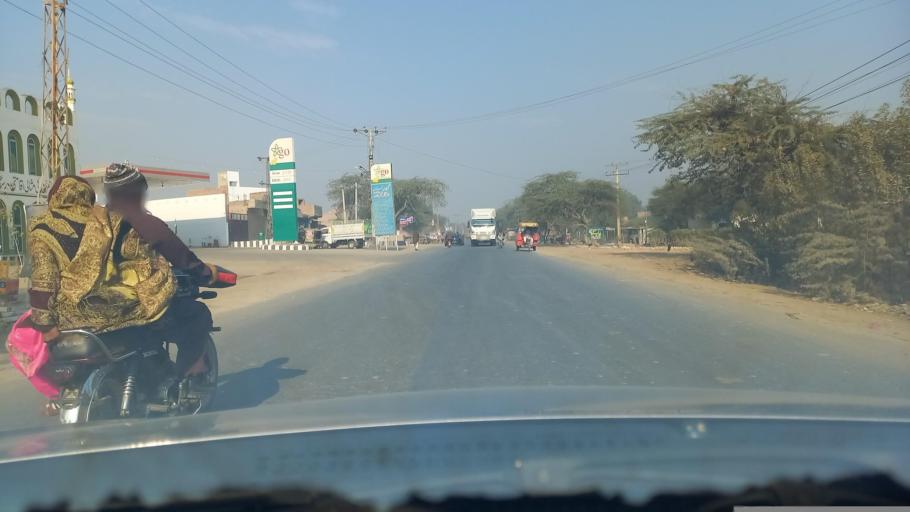}
    \caption{plannerqadeer for City Pulse}
    \end{subfigure}
    &
    \begin{subfigure}{0.24\linewidth}     \includegraphics[width=1\linewidth, height=.10\textheight]{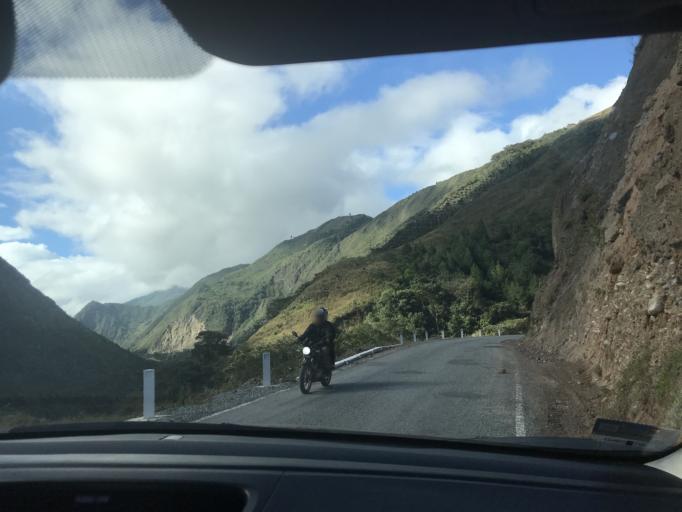}
    \caption{sedicla}
    \end{subfigure}
    &
    \begin{subfigure}{0.24\linewidth}     \includegraphics[width=1\linewidth, height=.10\textheight]{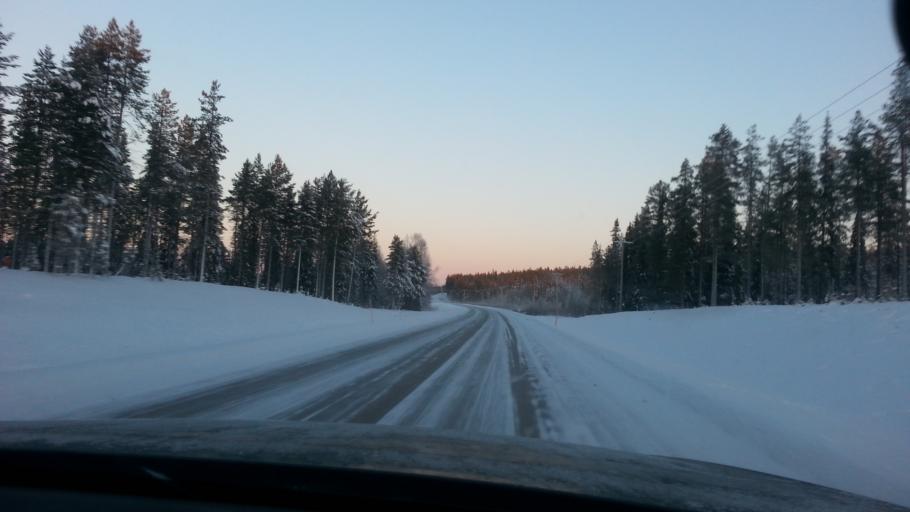}
    \caption{ kmajcher for Here}
    \end{subfigure}
    \\
    \begin{subfigure}{0.24\linewidth}     \includegraphics[width=1\linewidth, height=.10\textheight]{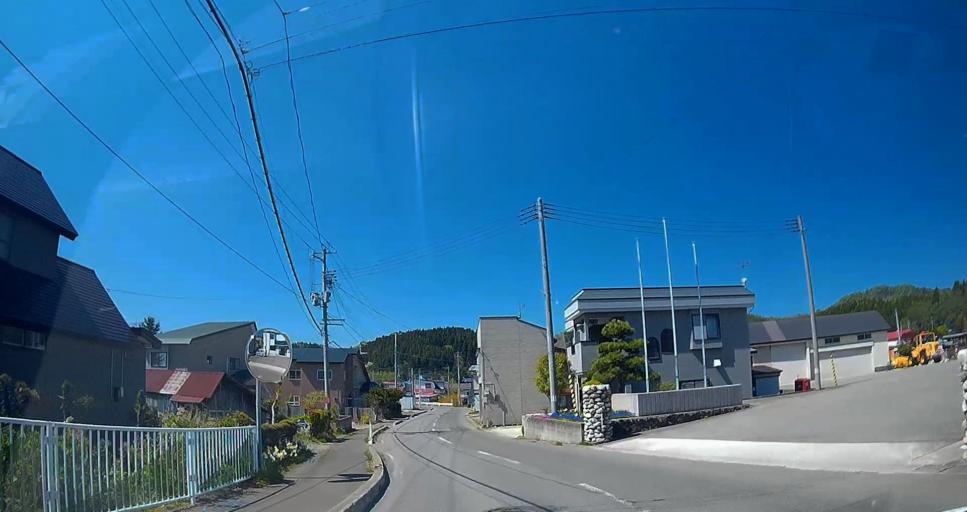}
    \caption{caesium}
    \end{subfigure}
    &
    \begin{subfigure}{0.24\linewidth}     \includegraphics[width=1\linewidth, height=.10\textheight]{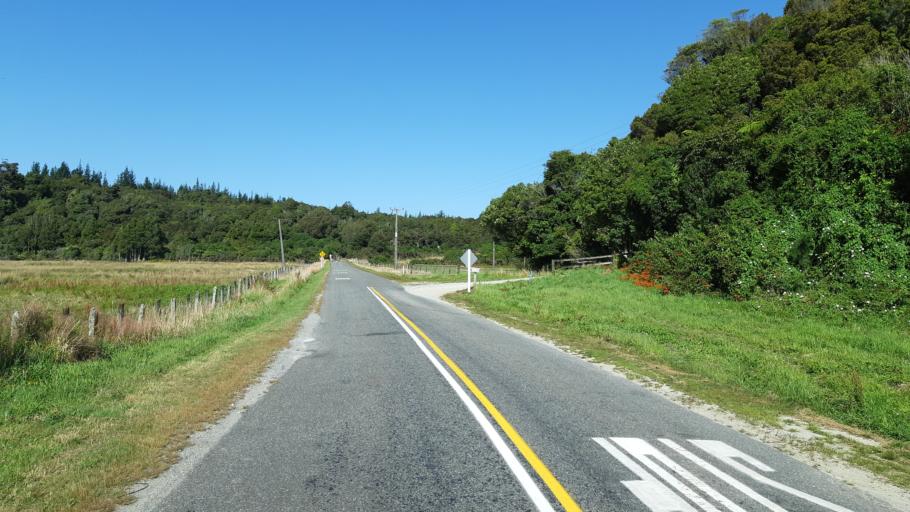}
    \caption{3stripes}
    \end{subfigure}
    &
    \begin{subfigure}{0.24\linewidth}     \includegraphics[width=1\linewidth, height=.10\textheight]{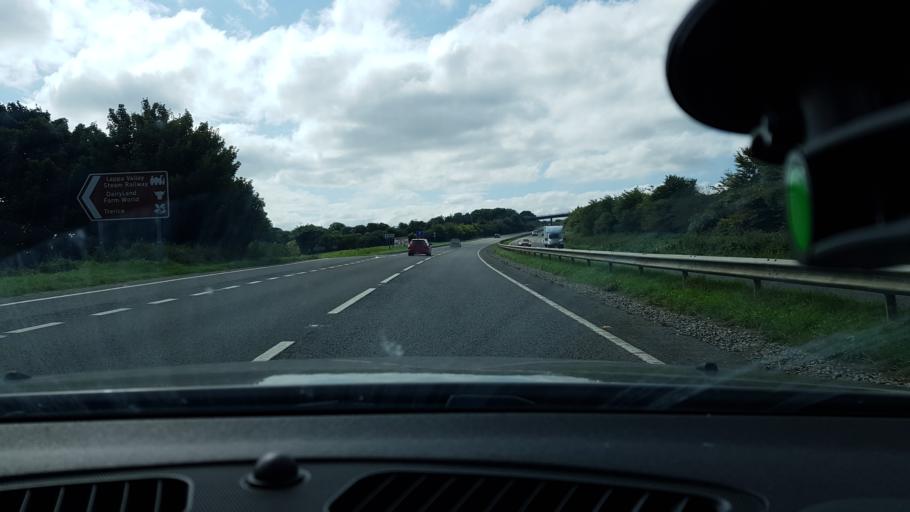}
    \caption{themadcabbie}
    \end{subfigure}
    &
    \begin{subfigure}{0.24\linewidth}     \includegraphics[width=1\linewidth, height=.10\textheight]{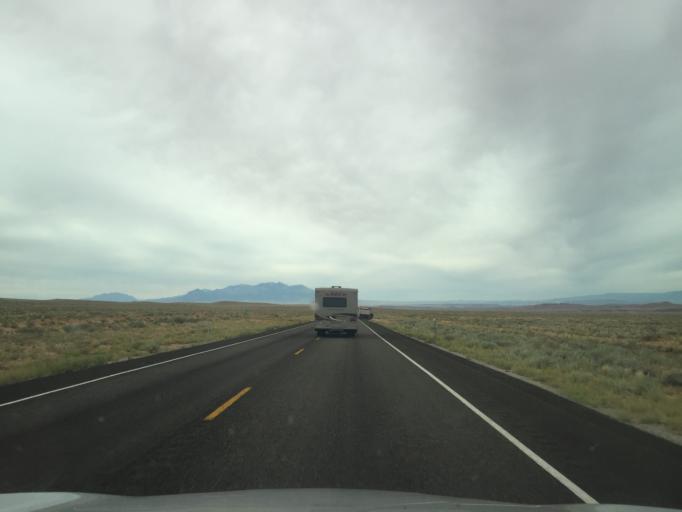}
    \caption{canadarunner}
    \end{subfigure}
    \\
    \begin{subfigure}{0.24\linewidth}     \includegraphics[width=1\linewidth, height=.10\textheight]{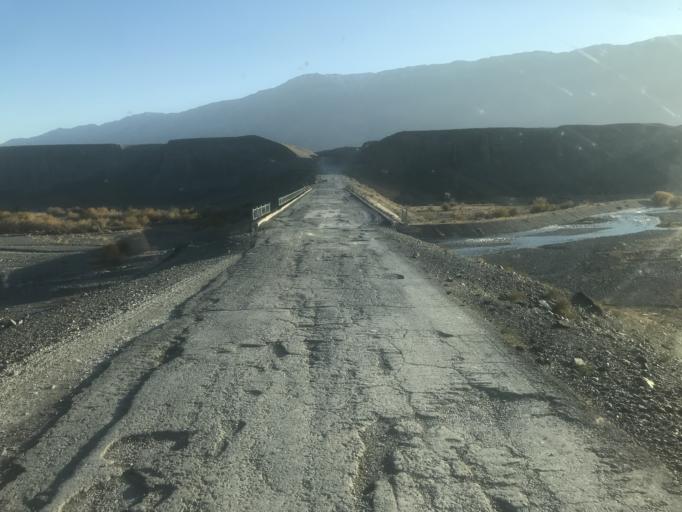}
    \caption{vik1607}
    \end{subfigure}
    &
    \begin{subfigure}{0.24\linewidth}     \includegraphics[width=1\linewidth, height=.10\textheight]{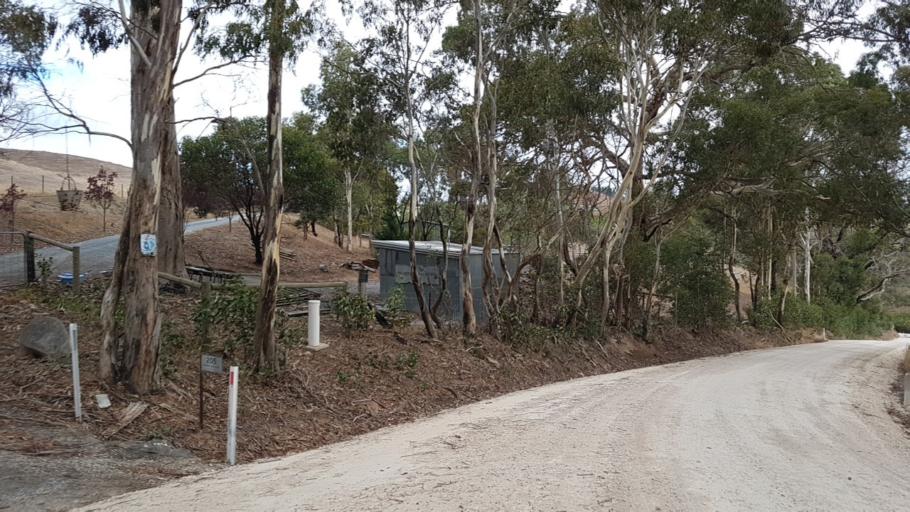}
    \caption{weinshaum}
    \end{subfigure}
    &
    \begin{subfigure}{0.24\linewidth}     \includegraphics[width=1\linewidth, height=.10\textheight]{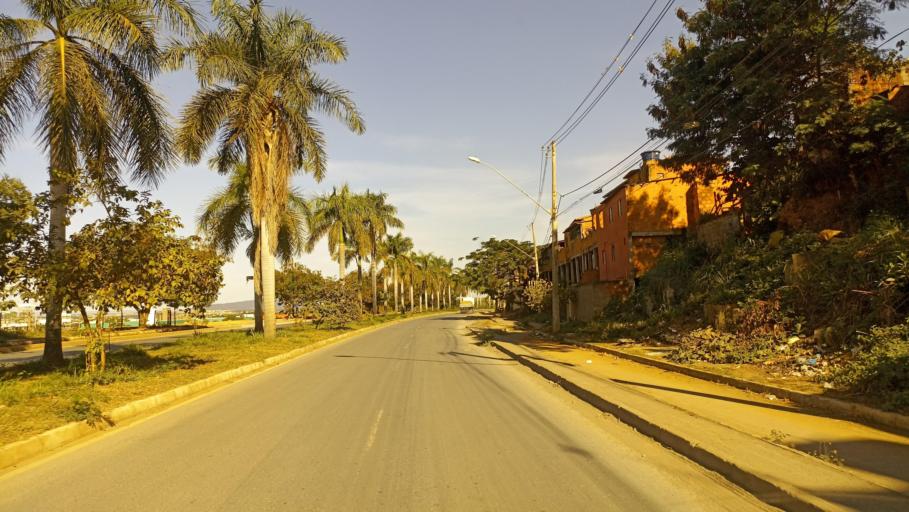}
    \caption{tulliomf}
    \end{subfigure}
    &
    \begin{subfigure}{0.24\linewidth}     \includegraphics[width=1\linewidth, height=.10\textheight]{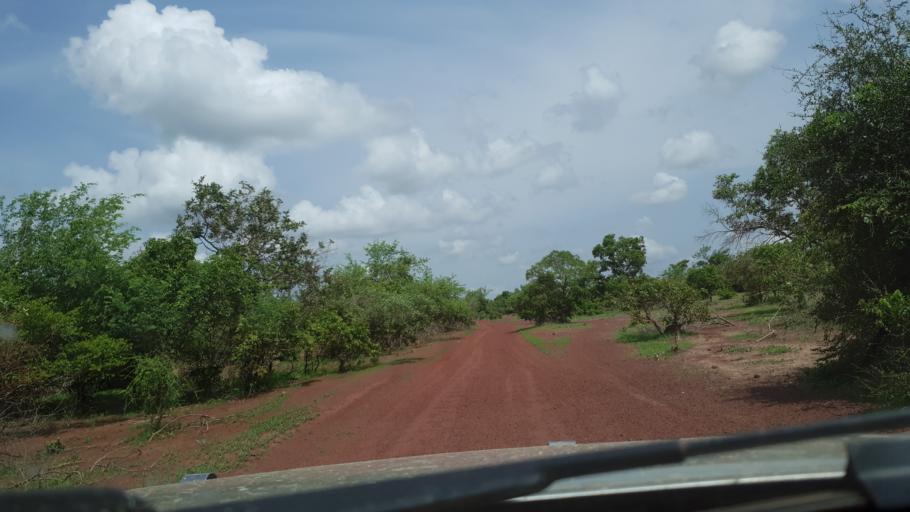}
    \caption{kosanka}
    \end{subfigure}
    \\
    \begin{subfigure}{0.24\linewidth}     \includegraphics[width=1\linewidth, height=.10\textheight]{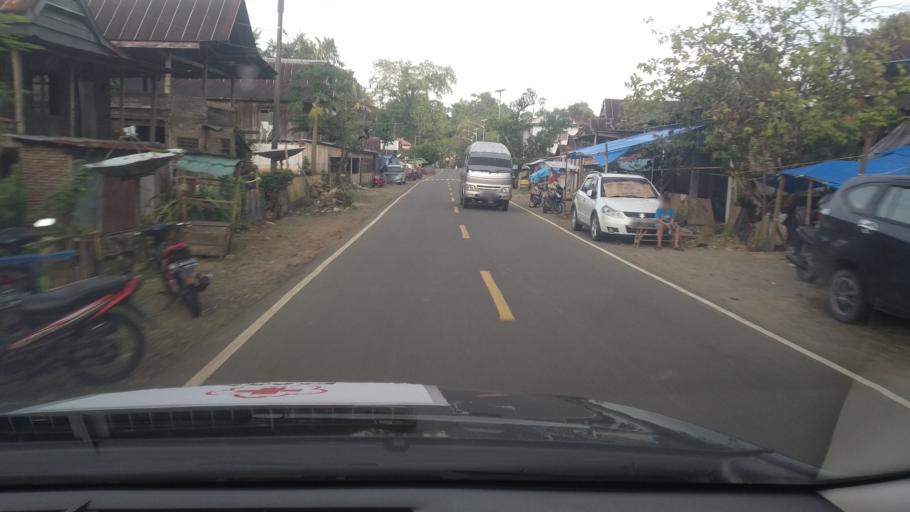}
    \caption{benjidad}
    \end{subfigure}
    &
    \begin{subfigure}{0.24\linewidth}     \includegraphics[width=1\linewidth, height=.10\textheight]{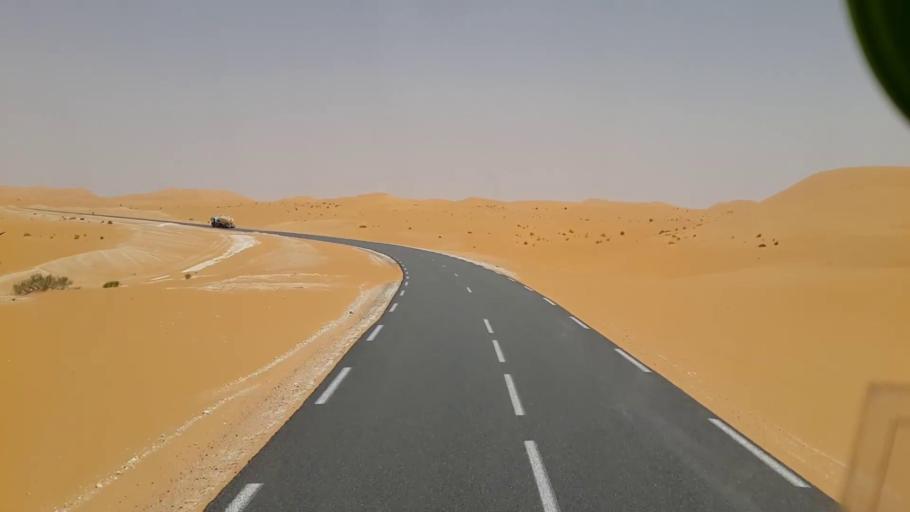}
    \caption{cut}
    \end{subfigure}
    &
    \begin{subfigure}{0.24\linewidth}     \includegraphics[width=1\linewidth, height=.10\textheight]{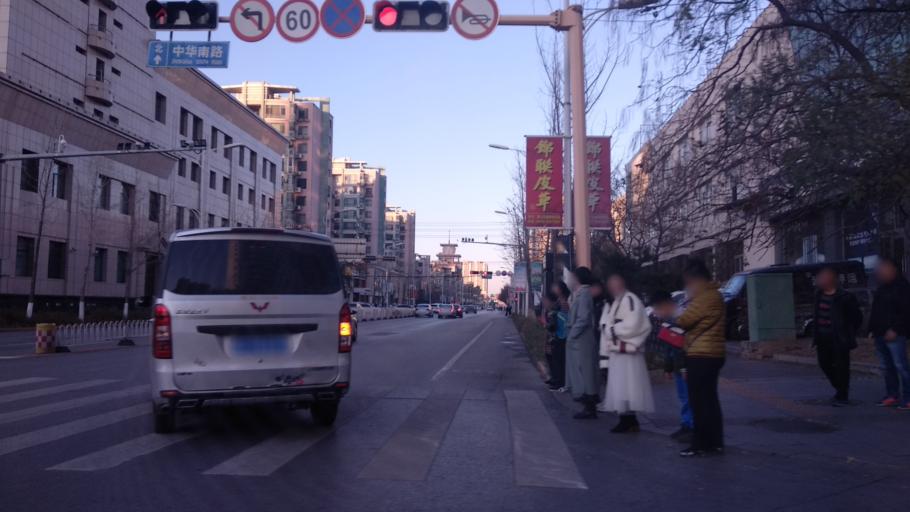}
    \caption{mapillario}
    \end{subfigure}
    &
    \begin{subfigure}{0.24\linewidth}     \includegraphics[width=1\linewidth, height=.10\textheight]{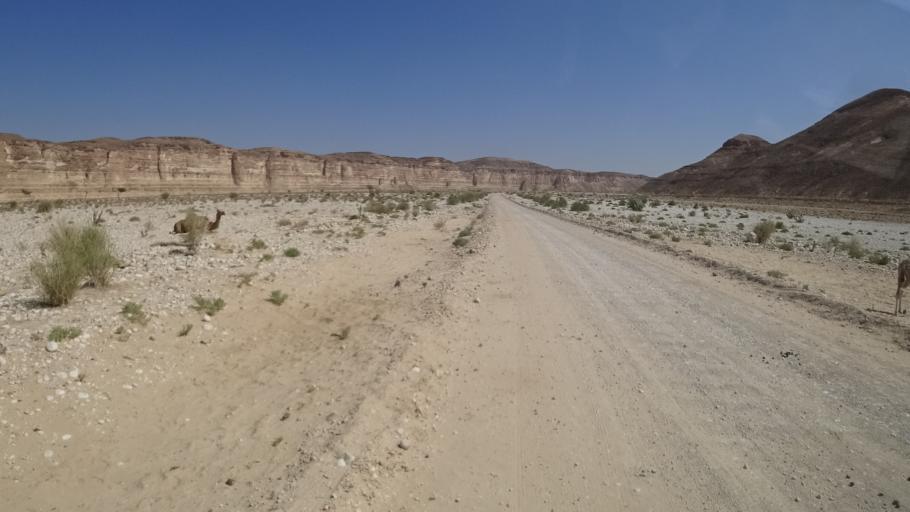}
    \caption{vbombaerts}
    \end{subfigure}
    \end{tabular}
    \caption{{\bf Images from OSV-5M.} The true locations can be found on the next page. The Mapillary users are credited in the subcaptions.}
    \label{fig:sup:illu}
\end{figure*}

\paragraph{Rotation-Based Filtering.} We perform a learning-based filtering based on a pretrained and frozen RotNet network \cite{gidaris2018unsupervised}. This model learns self-supervised image representations by training for the pretext task of predicting a random rotation applied to an input image. Although it it used as a pretext task in the original paper, it becomes useful for filtering out  images downloaded from Mapillary's website that are incorrectly rotated. We use the pretrained network to infer the rotation of various images and then use the following filtering strategy depending on RotNet's prediction:

\begin{itemize}
[itemsep=0.25em, wide, labelwidth=!, labelindent=2pt, topsep=0.25em, parsep=0em]
\item[-] \textit{0$^\circ$ }\textbf{(96\% of images)} For normal street view images the cues that signify an absence of rotation are multiple: the sky is up, and cars and pedestrians are upward. We keep these images unchanged.
\item[-] \textit{180$^\circ$ } \textbf{(4\%)} Over $90\%$ images predicted to be rotated by $180^\circ$ are, in fact, actually upside down. We rotate all these images by a half-turn. For the images in the test, we perform an additional visual inspection to remove the small proportion of non-localizable images not removed by the previous filters.
\item[-] \textit{90$^\circ$ or 270$^\circ$ }\textbf{(0.2\%)} Images predicted as rotated by a quarter-turn are in the vast majority taken indoors or in tunnels. We remove all such images from both the train and test set.
\end{itemize}

\begin{figure*}
    \centering
    \begin{tikzpicture}
    \begin{axis}[
        enlargelimits=false,
        axis on top,
        scale only axis,
        width=\textwidth,
        grid=major,
        xtick={-180,-90,0,90,180},
        ytick={-90,-45,0,45,90},
        height=0.5\textwidth,
        xmin=-180, xmax=180,
        ymin=-90, ymax=90,
        ]
        % Insert the map image here
         \addplot graphics[
            xmin=-173.5, xmax=186.6,
            ymin=-83.5, ymax=96.6
        ] {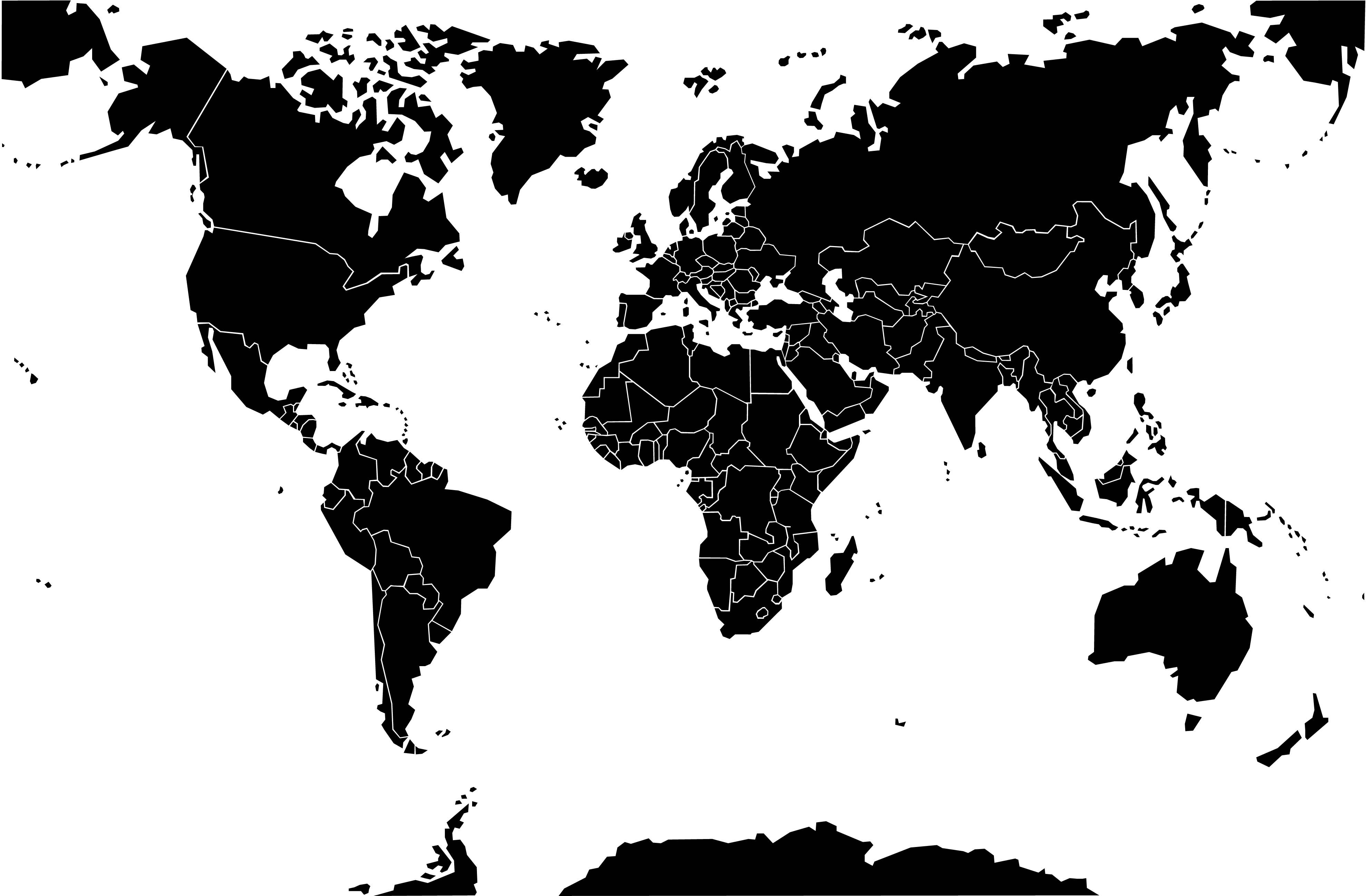};

    \draw[cyan, thick, smooth,->] plot coordinates {
   (axis cs:72.34, 30.05)
   (axis cs:70.34, 31.07)
   (axis cs:69.33, 31.57)
   (axis cs:68.3, 32.06)
   (axis cs:67.26, 32.55)
   (axis cs:66.22, 33.02)
   (axis cs:65.16, 33.49)
   (axis cs:64.09, 33.94)
   (axis cs:63.0, 34.39)
   (axis cs:61.91, 34.83)
   (axis cs: 59.69, 35.67)
    };

    \draw[cyan, thick, smooth,->] plot coordinates {
   (axis cs: -75.55, -10.38)
   (axis cs:-65.23, -1.53)
   (axis cs:-60.13, 2.93)
   (axis cs:-54.99, 7.36)
   (axis cs:-49.75, 11.74)
   (axis cs:-44.34, 16.01)
   (axis cs:-38.7, 20.15)
   (axis cs:-32.77, 24.09)
   (axis cs:-26.47, 27.78)
   (axis cs:-19.75, 31.16)
   (axis cs:-4.89, 36.73)
    };

    \draw[cyan, thick, smooth,->] plot coordinates {
   (axis cs: -110.62, 38.54)
   (axis cs:-110.29, 39.63)
   (axis cs:-110.13, 40.17)
   (axis cs:-109.96, 40.72)
   (axis cs:-109.78, 41.26)
   (axis cs:-109.61, 41.8)
   (axis cs:-109.43, 42.34)
   (axis cs:-109.25, 42.89)
   (axis cs:-109.06, 43.43)
   (axis cs:-108.88, 43.97)
   (axis cs:-108.49, 45.05)
    };

    \draw[cyan, thick, smooth] plot coordinates {
   (axis cs: 81.96, 45.69)
   (axis cs:90.61, 60.82)
   (axis cs:98.74, 67.99)
   (axis cs:113.6, 74.41)
   (axis cs:143.05, 78.73)
   (axis cs:180, 78.73)
    };
    \draw[cyan, thick, smooth,->] plot coordinates {
   (axis cs:-180, 78.34)
   (axis cs:-176.15, 78.34)
   (axis cs:-149.32, 73.56)
   (axis cs:-135.83, 66.98)
   (axis cs:-128.29, 59.75)
   (axis cs:-123.48, 52.24)
   (axis cs:-117.48, 36.82)
    };

    \draw[cyan, thick, smooth,->] plot coordinates {
   (axis cs: -43.88, -16.68)
   (axis cs:-47.08, -16.99)
   (axis cs:-48.68, -17.13)
   (axis cs:-50.28, -17.25)
   (axis cs:-51.89, -17.36)
   (axis cs:-53.5, -17.46)
   (axis cs:-55.11, -17.55)
   (axis cs:-56.72, -17.62)
   (axis cs:-58.33, -17.68)
   (axis cs:-59.94, -17.73)
   (axis cs:-63.17, -17.78)
    };

    \draw[cyan, thick, smooth,->] plot coordinates {
   (axis cs: 6.98, 31.43)
   (axis cs:-7.15, 41.89)
   (axis cs:-15.9, 46.32)
   (axis cs:-26.03, 49.98)
   (axis cs:-37.58, 52.61)
   (axis cs:-50.22, 54.01)
   (axis cs:-63.29, 54.01)
   (axis cs:-75.93, 52.61)
   (axis cs:-87.47, 49.98)
   (axis cs:-97.6, 46.32)
   (axis cs:-113.90, 36.87)
    };
    
 \draw[cyan, thick, smooth,->] plot coordinates {
   (axis cs:122.99, 41.11)
   (axis cs:121.57, 40.26)
   (axis cs:120.87, 39.83)
   (axis cs:120.18, 39.4)
   (axis cs:119.49, 38.96)
   (axis cs:118.82, 38.51)
   (axis cs:118.16, 38.07)
   (axis cs:117.5, 37.62)
   (axis cs:116.85, 37.16)
   (axis cs:116.21, 36.71)
   (axis cs: 114.95, 35.78)
    };

    \node[text=red, anchor=center] at (axis cs:113.79, -8.18) {\bf 1};
    \node[text=red, anchor=center] at (axis cs:72.34, 30.05) {\bf 2};
    \node[text=red, anchor=center] at (axis cs:-75.6, -10.38) {\bf 3};
    \node[text=red, anchor=center] at (axis cs:25.43, 66.83) {\bf 4};
    \node[text=red, anchor=center] at (axis cs:140.82, 41.14) {\bf 5};
    \node[text=red, anchor=center] at (axis cs:171.05, -42.66) {\bf 6};
    \node[text=red, anchor=center] at (axis cs:-4.96, 50.37) {\bf 7};
    \node[text=red, anchor=center] at (axis cs:-110.62, 38.54) {\bf 8};
    \node[text=red, anchor=center] at (axis cs:138.87, -34.84) {\bf 10};
    \node[text=red, anchor=center] at (axis cs:81.66,45.69) {\bf 9};
    \node[text=red, anchor=center] at (axis cs:-43.88, -16.68) {\bf 11};
    \node[text=red, anchor=center] at (axis cs:-7.05, 13.33) {\bf 12};
    \node[text=red, anchor=center] at (axis cs:118.85, -2.74) {\bf 13};
    \node[text=red, anchor=center] at (axis cs:6.98, 31.43) {\bf 14};
    \node[text=red, anchor=center] at (axis cs:122.99, 41.11) {\bf 15};
    \node[text=red, anchor=center] at (axis cs:52.93, 17.29) {\bf 16};
    %\node[text=red, anchor=center] at (axis cs:78.13, 29.93) {17};
    %\node[text=red, anchor=center] at (axis cs:-14.47, 15.22) {18};

    %\draw[blue] (axis cs:113.79, -8.18) -- (axis cs:113.90, -8.41); % 137033365014320 1
    %\draw[cyan] (axis cs:72.34, 30.05) -- (axis cs:59.69, 35.67); % 1622455258212885 2
    %\draw[cyan] (axis cs:-75.55, -10.38) -- (axis cs:-4.89, 36.73); % 210429954039398 3
    %\draw[cyan] (axis cs:25.43, 66.83) -- (axis cs:25.42, 66.79); % 320105979491109 4
    %\draw[cyan] (axis cs:140.82, 41.14) -- (axis cs:141.36, 41.25); % 464804747946887 5
    %\draw[cyan] (axis cs:171.04, -42.66) -- (axis cs:171.17, -42.15); % 474519253765020 6
    %\draw[cyan] (axis cs:-4.96, 50.37) -- (axis cs:-4.80, 50.39); % 517681129360654 7
    %\draw[cyan] (axis cs:-110.62, 38.54) -- (axis cs:-108.49, 45.05); % 827026111505364 8
    %\draw[cyan] (axis cs:81.96, 45.69) -- (axis cs:-117.48, 36.82); % 217887346522004 9
    %\draw[cyan] (axis cs:138.87, -34.84) -- (axis cs:138.99, -35.01); % 975112963027888 10
    %\draw[cyan] (axis cs:-43.88, -16.68) -- (axis cs:-63.17, -17.78); % 1113744692540897 11
    %\draw[cyan] (axis cs:-7.05, 13.33) -- (axis cs:-6.44, 12.54); % 1139006019900435 12
    %\draw[cyan] (axis cs:118.85, -2.74) -- (axis cs:119.91, -2.97); % 1207955469639614 13
    %\draw[cyan] (axis cs:6.98, 31.43) -- (axis cs:-113.90, 36.87); % 169782348369756 14
    %\draw[cyan] (axis cs:122.99, 41.11) -- (axis cs:114.95, 35.78); % 2872700852992867 15
    %\draw[cyan] (axis cs:52.93, 17.29) -- (axis cs:53.10, 17.50); % 3882682481821500 16

    \end{axis}
\end{tikzpicture}

    \caption{{\bf True Locations.} Location of the images of \figref{fig:sup:illu}. With \textcolor{cyan}{blue} we visualize errors of the \textcolor{COMBINEDCOLOR}{combined model} that are superior to $500$ km. Most of the images (9 out of 16) are predicted within $500$km of where they were taken. We observe that two difficult images (9 and 14) are erroneously mapped to the US, which could be explained by the geographical bias of the training set.}
    \label{fig:sup:answer}
\end{figure*}

\subsection{Discussion}
\paragraph{Why Not Just Subsample YFCC100M?}
The wide adoption of YFCC100M, with its nearly 50 million geotagged images,, might question the need for creating yet another geotagged image dataset. However, several compelling reasons justify creating OpenStreetView-5M instead of subsampling YFCC100M:
\begin{itemize}[itemsep=0.25em, wide, labelwidth=!, labelindent=2pt, topsep=0.25em, parsep=0em]
\item[-] \textit{Data Distribution.} The images shared on Flickr do no aim to capture our world in an  objective way, but instead focus on aesthetic and cultural value. For example, recognizable landmarks like the Eiffel Tower or the Louvre, are a cultural symbol of the city of Paris, yet they lack any information that is useful in identifying other cities as French or even other streets as Parisian. Additionally, many images are renders or infographics. In contrast, OSV-5M only features dashcam pictures, that offer a consistent front-view perspective, that is more objective as it doesn't focus on something specific, and thus may be more beneficial for learning visual geographical representations.
\item[-] \textit{Localizability.} From a manual inspection of $1000$ images we find that fewer than $10$\% ($\pm1.3$\%, $95$\% confidence) of YFCC100M's images are perceptually localizable. In stark contrast, OSV-5M boosts this perceptual localizability to a rate of $96.1$\% ($\pm0.57$, $95$\% confidence), making it a more suitable candidate for a standard evaluation benchmark for global geolocation.
\item[-] \textit{Geographical Bias.} Images in the YFCC100M dataset exhibit a high cultural bias towards the Western world, with over 35\% of images from the US and nearly 70\% from North America and Europe \cite{yfcc-analysis}. OSV-5M offers a more equitable global representation, as detailed in Figure~2 of the main paper.
\item[-] \textit{Selection Challenges.} Subsampling YFCC100M based on metadata alone is ambiguous: 30\% of images lack titles, 68\% lack descriptions, 30\% lack tags, and 50\% lack geotagging. The tags  ``travel'' and ``nature'' cover fewer than 2 million images. Using instead automated selection methods may  inadvertently propagate existing biases, such as filtering street views of non-Western countries.
\item[-] \textit{Persistence.} As happens with a lot of large research dataset, YFCC comes only as a collection of image URLs that need to be downloaded directly from Flickr. Such a dataset construction approach, even if the only feasible choice for very large datasets, is very volatile and can prevent future reproducibility. For example, 60\% of the 2014 YFCC-split \cite{flickr-mouselly} was deleted by 2020 \cite{izbicki2020exploiting}. While YFCC100M used to be hosted on Yahoo’s Webscope, this option is no longer available \cite{multimediacommons}. Instead users need to create an AWS account, that requires a credit card to acquire API credentials for downloading the data through a designated S3 bucket \cite{yfccwebsite}. Even if no charge is applied, this setting may be prohibitive for academics or residents of certain countries. Also, due to the sensitive nature of the Flickr data, users need to make a formal request to download the dataset, something that isn't needed for our dataset. Instead, OSV-5M ensures persistence, open and easy access for long-term and broad usage.
\end{itemize}

To summarize, YFCC100M is a vast and unstructured set of images, a subset of which may be well suited for localization and place recognition. However, the ambiguous localizability, geographical content, metadata, persistence, and access to its images highlight the need for a dedicated dataset like OSV-5M, specifically designed for the task of global visual geolocation.

\paragraph{Visible GeoTags.} Due to the diversity in user input data, we found that a small percentage of images ($< 5\%$) have a visible overlayed text on the bottom part that tags their location. This should be taken into consideration when constructing a benchmark for a future dataset. However, due to the standard ViTs resampling of images to $224\times 224$, these coordinates become indecipherable, as demonstrated in \figref{fig:decipher}. We implement for our data loader the option to add a Gaussian blur with a width of $2$ to the bottom 14 rows. When training and/or testing a baseline model with this blur, we observe only small and inconclusive differences in score: training without blurring but testing with it yielded slightly better results than both training and testing without the blur, yet training and testing with the blur produced inferior outcomes.
This indicates that (i) the network is not able to read the coordinates, and (ii) the bottom rows do not contain critical geographic information.
However, we recommend using the blur for methods that use higher-resolution models to obscure any potential location-specific details in the text.

\begin{figure*}
    \centering
    \begin{tabular}{cc}
    
    \begin{subfigure}{.8\columnwidth}
    \includegraphics[width=\linewidth]{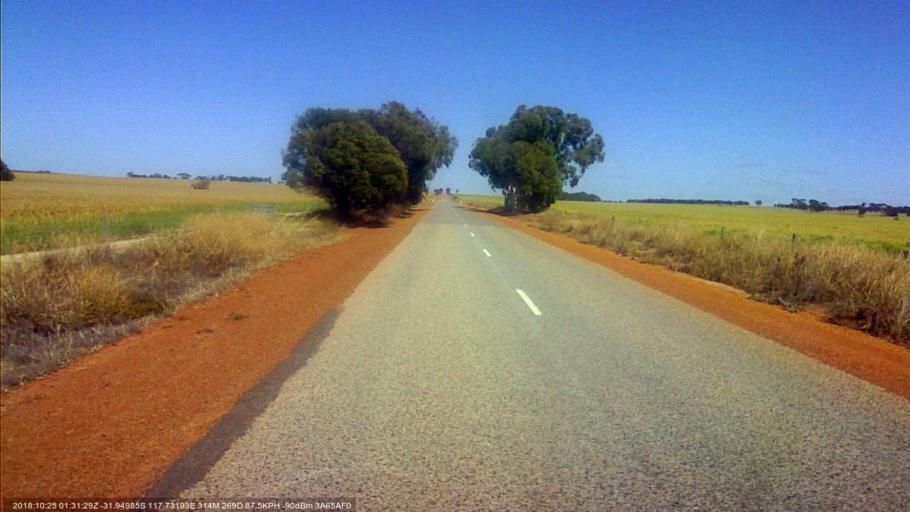}
   
    \caption{Full resolution image}
    \label{fig:leak:a}
    \end{subfigure}
         &  
    \begin{subfigure}{1.2\columnwidth}
    
    \begin{tikzpicture}
    \node [anchor=south west] at (0,0) {\includegraphics[width=0.32\linewidth]{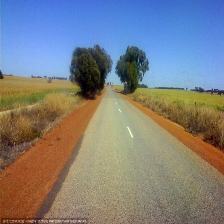}};
    \draw[red, thick] (0.2,0.15) rectangle (1,0.25);
    \node  [anchor=south west]  (zoom) at (3.5,0) {
    \includegraphics[width=.5\columnwidth]{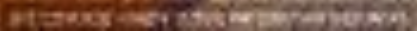}};
    \draw[red, thick, ->] (1,0.2) -- (zoom); 
    
    \node  [anchor=south west]  (zoom2) at (3.5,1.5) {
    \includegraphics[width=.5\columnwidth]{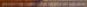}};

    \node [above=-2mm of zoom] {no blurring};
    \node [above=-2mm of zoom2] {blurring};
    \end{tikzpicture}

    \caption{Image rescaled in dataloader.}
    \label{fig:leak:b}
    \end{subfigure}
    \end{tabular}
    % \vspace{5mm}
    \caption{{\bf Visible Geotagging.} A small minority of images ($<$ 5\%) have visually overlayed geotags at their bottom left corner \subref{fig:leak:a}. For those images as resized by our data loader to $224 \times 224$ and as optionally blurred, we empirically measure to not provide any important information that the network can use to improve its performance \label{fig:leak:b}.}
    
    \label{fig:decipher}
\end{figure*}
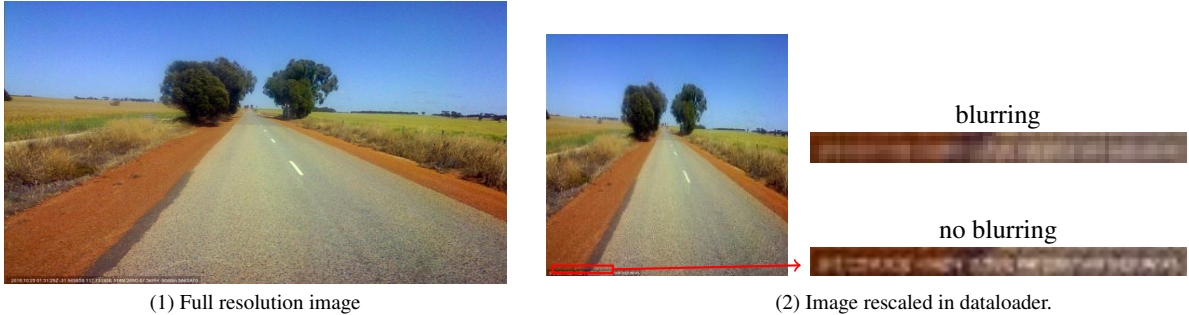

\paragraph{Limitations.}
We list three main limitations of our OSV-5M dataset:
\begin{itemize}[itemsep=0.25em, wide, labelwidth=!, labelindent=2pt, topsep=0.25em, parsep=0em]
    \item[(i)] \textit{Geographical Bias in Training.} Due to our reliance crowd-sourced from Mapillary users, the distribution of locations is biased towards Western countries. We designed our test set to explicitly balance this distribution, but the training set remains affected by the number of selected images. 
    \item[(ii)] \textit{User separation.} We successfully separated images from the same sequence between training and test sets. However, we could not separate images uploaded by the same user on different days, as the required metadata was not available at the time of the dataset construction.
    \item[(iii)] \textit{Resolution.} The dataset provides images with a vertical resolution of $512$ pixels. This restricts the ability to zoom in and read distant texts, for example in street signs, potentially obscuring valuable visual cues. However, through our metadata users can access higher-resolution versions of all our images on the mapillary website.
\end{itemize}

\paragraph{Training SOTA methods on OSV-5M.} Many state-of-the-art geolocation methods \cite{haas2023learning,haas2023pigeon,Google-World-Streets-15K-geolocation,PlaNet} either rely on private datasets or lack publicly available code, that prevents their evaluation. In our main paper we evaluated the performance of the pretrained StreetCLIP model both for zero-shot retrieval (Tab~6 and Fig~6) and as a pretrained image encoder (Tab~2), yet the implementation required to  fine-tune the model is not publicly available. Similarly, the complete training code of Translocator \cite{translocator} is also not available. We managed to train the publicly available ISN model \cite{muller2018geolocationshort} on OSV-5M, achieving good performance which we attribute to its bespoke geocell module.
The aforementioned difficulty in training and evaluating SOTA models show the clear need for open-source datasets and implementations of visual geolocation approaches, that our paper directly addresses.

\paragraph{Geoscore.} In our paper geoscore is introduced as a better evaluation method as it strikes a balance between rewarding precision and not being oversensitive to outlier predictions.
Let us consider, for example, a model which produces nine accurate predictions but fails on the tenth image, choosing New Zealand instead of Ireland, a $20\,000$km mistake. Contrast this with another model which consistently mispredicts by $2\,000$km. Solely examining the mean error might misleadingly favor the latter model, when the first one has a higher geographic proficiency.
In terms of geoscore, the model with one major error would achieve an average score close to $4500$, while the one that is consistently off would score $1300$. In that way, geoscore provides a more intuitive way to compare the performance of models on our dataset. See \figref{fig:geoscore} for an illustration of Geoscore.

\begin{figure}[t]
    \centering
    \includegraphics[width=\linewidth, height=.15\textheight]{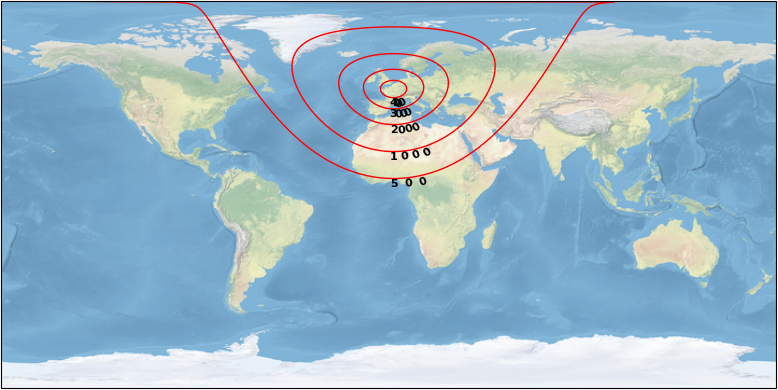}
    \caption{{\bf Geoscore.} From a point centered in Paris, red contours highlight level sets of the score along the earth's spherical geometry.}
    \label{fig:geoscore}
\end{figure}

%===========================
\section{Additional Experiments}
%===========================
\label{sec:sup:xp}
This section presents further results and analysis of our proposed framework.

\paragraph{Auxiliary Supervision.}  We start by evaluating the performance gained by learning to predict various auxiliary information. Based on their coordinates, we associate to each image of our dataset the following meta-data, according to its latitude and longtitude coordinates:
\begin{itemize}[itemsep=0.25em, wide, labelwidth=!, labelindent=2pt, topsep=0.25em, parsep=0em]
    \item[-] \textit{Land Cover.} Relying on the Global Land Cover Share Database \cite{latham2014global}, we classify each image of our dataset into one of $11$ land cover types, such as artificial, forest, or crops.
    \item[-] \textit{Climate.} We use recent Köppen-Geiger climate classification maps \cite{beck2018present} to associate each image with a climate type among $31$, such as tropical rainforest, arid steppe, or temperate with dry winter.
    \item[-] \textit{Soil Type.} Thanks to the Digital World Soil Map \cite{sanchez2009digital}, we characterize the local soil with a $15$ class nomenclature, such as acrisols, fluvisols, or ferralsols.
     \item[-] \textit{Driving Side.} We also add a binary indicator for whether a country uses left or right-hand traffic.
     \item[-] \textit{Distance to the Sea.} For all locations we compute the distance to their nearest sea.
\end{itemize}
The maps we used to extract land cover, climate, and soil types come in a resolution of $1$~km (or $30$ arc-seconds).

\begin{table}[t]
    \caption{{\bf Auxiliary Variables.} We report the impact on geolocation performance of learning to predict various auxiliary variables. We also report the performance on the test set for each variable as the overall accuracy or the average error.}
    \label{tab:auxiliary}
    \centering
    \resizebox{\linewidth}{!}
    {
    \begin{tabular}{@{}lcccccccc}
        \toprule
        & \multirow{2}{*}{\makecell[b]{Num of \\classes.}} & \multirow{2}{*}{\makecell[b]{Perf.\\ test}} & \multirow{2}{*}{\makecell[b]{Geo $\uparrow$\\score} } & \multirow{2}{*}{\makecell[b]{Dis $\downarrow$\\tance} } & \multicolumn{4}{c}{Classification accuracy $\uparrow$}\\%\cline{2-3}
        
        \cmidrule(lr){6-9} &  &   &  & & country & region & area & city \\
        \midrule
        \textcolor{FINETUNECOLOR}{no auxiliary} & - & - & 2893 & 2085 & \bf 54.9 & 19.1 & 1.6 & \bf 0.8\\\greyrule
        \rowcolor{gray!10} land cover& 11 & 54.8 & 2821 & 2102 & 52.2 & 16.9 & 1.4 & 0.7\\
        climate& 31 & \textcolor{white}{\;\;\;}58.3 & 2898 & 2022 & 53.7 & 18.8 & \bf 1.7 & \bf 0.8\\
        \rowcolor{gray!10} soil type &15 & \textcolor{white}{\;\;\;}47.7 & 2826 & 2111 & 52.4 & 17.6 & 1.5 & 0.7\\
        driving side &\textcolor{white}{0}1 & \textcolor{white}{\;\;\;}94.6 & 2896 & 2025 & 54.5 & 18.7 & 1.6 & 0.7 \\
        \rowcolor{gray!10} dist to sea & - & 543km & 2870 & 2053 & 52.5 & 18.7 & 1.5 & 0.7\\\greyrule
        all & - & - & \bf 2910 & \bf 1987 & 54.0 & \bf 19.8 & 1.6 & \bf 0.8\\\bottomrule
    \end{tabular}}
    %}
\end{table} 

We use an MLP $f^\text{aux}$ to predict the image's metadata in addition to its coordinates. All categorical variables are supervised with the unweighted sum of cross-entropy terms, while the distance to the sea is supervised with the L1 loss.
Adding auxiliary tasks encourages the model to focus on relevant geographical cues. As seen in \tabref{tab:auxiliary}, we only observe a modest impact, indicating that the large train set of OSV-5M allows our model to already learn good latent variables for geolocation. It should be noted that our model can perform accurate predictions for complex geographic variables in the test set, which may have some useful applications in itself.
 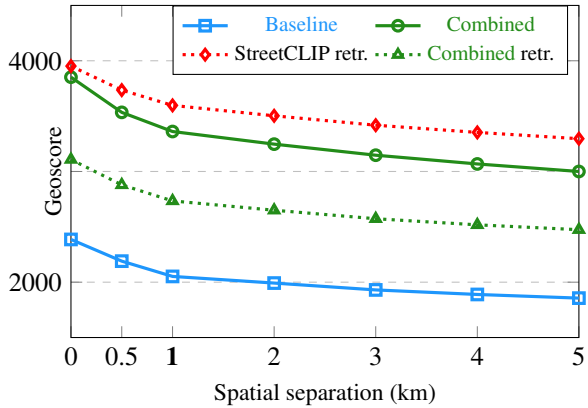
\begin{figure}[t]
    \centering
    \begin{tikzpicture}
\begin{axis}[
    xlabel={\small Spatial separation (km)},
    ylabel={\small Geoscore},
    xmin=0, xmax=5000,
    ymin=1500, ymax=4500,
    xtick={0, 500,1000,2000,3000,4000, 5000},
    xticklabels={0, 0.5, \bf 1,2,3,4, 5},
    ytick={2000,3000,4000},
    yticklabels={2000,,4000},
    legend pos=north west,
    ymajorgrids=true,
    grid style=dashed,
    height=6cm,
    width=\linewidth,
    ylabel style={at={(axis description cs:0.0,.5)},rotate=0}, 
    %xlabel style={at={(axis description cs:0.55,-0)}}, % Adjust y-axis label position
    legend style={at={(0.2,1)}},
    legend columns=2,
]

\addplot[
    color=BASELINECOLOR,
    mark=square,
    very thick,
    ]
    coordinates {
    (0,2386)
    (500,2190)
    (1000,2052)
    (2000,1992)
    (3000,1930)
    (4000,1889)
    (5000,1856)
    };
    \addlegendentry{\footnotesize \textcolor{BASELINECOLOR}{Baseline}}

\addplot[
    color=COMBINEDCOLOR,
    mark=o,
    very thick,
    ]
    coordinates {
    (0,3852)
    (500,3534)
    (1000,3361)
    (2000,3247)
    (3000,3147)
    (4000,3068)
    (5000,3000)
    };
    \addlegendentry{\footnotesize \textcolor{COMBINEDCOLOR}{Combined}}

    \addplot[
    color=red,
    mark=diamond,
    very thick,
    dotted,
    mark options={solid},
    ]
    coordinates {
    (0,3952)
    (500,3735)
    (1000,3597)
    (2000,3503)
    (3000,3418)
    (4000,3353)
    (5000,3296)
    };
    \addlegendentry{\footnotesize StreetCLIP retr.}

    \addplot[
    color=COMBINEDCOLOR,
    mark=triangle,
    very thick,
    mark options={solid},
    dotted
    ]
    coordinates {
    (0,3108)
    (500,2879)
    (1000,2734)
    (2000,2650)
    (3000,2573)
    (4000,2518)
    (5000,2474)
    };
    \addlegendentry{\footnotesize \textcolor{COMBINEDCOLOR}{Combined} retr.}

\end{axis}
\end{tikzpicture}
    % \vspace{-3mm}
    \caption{{\bf Spatial Separation.} We report the performance of different approaches for test sets defined by various separation radii for the train set. }
    \label{fig:sep}
\end{figure}

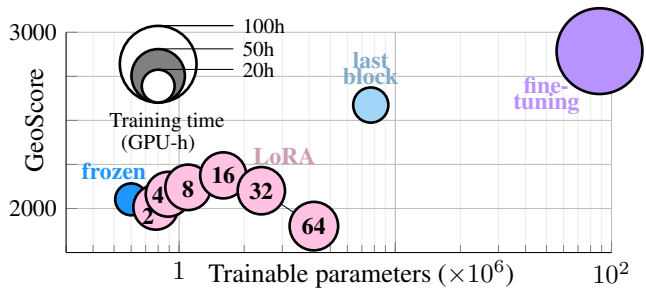
\begin{figure}[t]
\definecolor{FROZENCOLOR}{RGB}{30, 150, 252}
\definecolor{LASTCOLOR}{RGB}{162, 214, 249}
\definecolor{FINECOLOR}{RGB}{184, 146, 255}
\definecolor{LORACOLOR}{RGB}{255, 194, 226}

\centering
%\resizebox{\columnwidth}{!}{
\hspace{-.40cm}
\begin{tikzpicture}
\begin{semilogxaxis}[%
    width=.87\linewidth,
    height=.13\textheight,
    scale only axis,
    ytick={2000,2250,2500,2750,3000},
    yticklabels={\small{2000},,,,\small{3000}},
    xtick={0.1, 1,10,100},
    xticklabels={$10^{-1}$,$1$,,$10^2$},
    xmin=0.3,
    xmax=100,
    ymin=1750,
    ymax=3000,
    axis x line*=bottom,
    axis y line*=left,
    grid={major},
    xminorgrids=true,
    minor grid style={gray!15,line width=.25pt},
    legend pos=south east,
    ylabel={GeoScore},
    xlabel={Trainable parameters ($\times 10^6$)},
    ylabel style={xshift=3mm,yshift=-7mm},
    xlabel style={xshift=0.3cm,yshift=+0.55cm},
    clip marker paths=false,
    clip mode=individual,
    ]
]

\addplot[
    scatter=true,
    line width=0pt,
    mark=*,
    nodes near coords*={\size},
    scatter/@pre marker code/.code={
    \def\markopts{fill=FROZENCOLOR, line width=1pt}
    \expandafter\scope\expandafter[\markopts]},
    scatter/@pre marker code/.append style={/tikz/mark size=1.5*sqrt(\size)-1pt},
    scatter/@post marker code/.code={\endscope},
    visualization depends on={value \thisrow{f} \as \size} 
] table [x={s},y={m}] {
    f        m          s
    22     2052         0.6
};
\node[FROZENCOLOR, above] at (axis cs:0.5,2110.0){\small{\bf {frozen}}};

\addplot[
    scatter=true,
    line width=0pt,
    mark=*,
    nodes near coords*={\size},
    scatter/@pre marker code/.code={
    \def\markopts{fill=LASTCOLOR, line width=1pt}
    \expandafter\scope\expandafter[\markopts]},
    scatter/@pre marker code/.append style={/tikz/mark size=1.5*sqrt(\size)-1pt},
    scatter/@post marker code/.code={\endscope},
    visualization depends on={value \thisrow{f} \as \size} 
] table [x={s},y={m}] {
    f        m          s
    26     2587         7.7
};
\node[LASTCOLOR!80!black, above] at (axis cs:7.7,2740.0){\small{\bf {last}}};
\node[LASTCOLOR!80!black, above] at (axis cs:7.7,2650.0){\small{\bf {block}}};

\addplot[
    scatter=true,
    line width=0pt,
    mark=*,
    nodes near coords*={\size},
    scatter/@pre marker code/.code={
    \def\markopts{fill=FINECOLOR, line width=1pt}
    \expandafter\scope\expandafter[\markopts]},
    scatter/@pre marker code/.append style={/tikz/mark size=1.5*sqrt(\size)-1pt},
    scatter/@post marker code/.code={\endscope},
    visualization depends on={value \thisrow{f} \as \size} 
] table [x={s},y={m}] {
    f        m          s
    132     2893         88
};
\node[FINECOLOR, below] at (axis cs:50,2800.0){\small{\bf {fine-}}};
\node[FINECOLOR, below] at (axis cs:50,2710.0){\small{\bf {tuning}}};

\addplot[
    scatter=true,
    line width=0pt,
    mark=*,
    nodes near coords*={\size},
    scatter/@pre marker code/.code={
    \def\markopts{fill=LORACOLOR, line width=1pt}
    \expandafter\scope\expandafter[\markopts]},
    scatter/@pre marker code/.append style={/tikz/mark size=1.5*sqrt(\size)-1pt},
    scatter/@post marker code/.code={\endscope},
    visualization depends on={value \thisrow{f} \as \size} 
] table [x={s},y={m}] {
    f        m          s
    40 	   2007         0.78
    40     2081         0.89
    41     2120         1.1
    42     2188         1.6
    44     2101         2.4
    46     1900        4.2
    %48     1760	    7.7
    %49	   1707         14.8
};
\node[LORACOLOR!80!black, right] at (axis cs:2,2300.0){\small{\bf {LoRA}}};
\node[black] at (axis cs:0.72,1960.0){\small{\bf {2}}};
\node[black] at (axis cs:0.80,2075.0){\small{\bf {4}}};
\node[black] at (axis cs:1.1,2110.0){\small{\bf {8}}};
\node[black] at (axis cs:1.6,2188.0){\small{\bf {16}}};
\node[black] at (axis cs:2.4,2101.0){\small{\bf {32}}};
\node[black] at (axis cs:4.2,1900.0){\small{\bf {64}}};
%\node[black] at (axis cs:7.7,1760.0){\small{\bf {128}}};
%\node[black] at (axis cs:14.8,1707.0){\small{\bf {256}}};

\node[draw=black,right, circle, fill=white, minimum size = 15*sqrt(100)-5pt, scale=0.2, anchor=south,line width=1pt] at (axis cs:0.8,2600) (n1){};
\node[draw=black,right, circle, fill=black!50!white, minimum size = 15*sqrt(50)-5pt, scale=0.2, anchor=south,line width=1pt] at (axis cs:.8,2600) (n2) {};
\node[draw=black,right, circle, fill=white, minimum size = 15*sqrt(20)-5pt, scale=0.2, anchor=south,line width=1pt] at (axis cs:.8,2600) (n4) {};

\draw [-] (n1.north) -- ++ (1cm,0cm);
\draw [-] (n2.north) -- ++ (1cm,0cm);
\draw [-] (n1.north) -- ++ (1cm,0cm);
\draw [-] (n4.north) -- ++ (1cm,0cm);

\node[draw=none,anchor=north, right=of n1.north] (x) {\footnotesize 100h};
\node[draw=none,anchor=west, right=of n2.north] (x) {\footnotesize  50h};
\node[draw=none,anchor=west, right=of n4.north] (x) {\footnotesize 20h};
\node[draw=none,anchor=south, below=of n4, yshift=1cm] (x) { \footnotesize \;\;\;Training time};
\node[draw=none,anchor=south, below=of n4, yshift=0.7cm] (x) { \footnotesize (GPU-h)};

\end{semilogxaxis}
\end{tikzpicture}
\caption{
{\bf Effect of LoRA Bottleneck Width.} 
We report the performance of finetuning with LoRA of different bottleneck widths, in comparison to finetuning the last block, or the whole network. For each experiment, the marks' radius are proprtional to the training time.}% We notice that last block finetuning yields much better performance than LoRA on our task for even less time.}
\captionsetup{belowskip=0pt}
\label{fig:lora}
\end{figure}
\paragraph{Spatial Separation.} We study the impact of the radius of spatial separation between the train and the test set. We do this by creating test sets along different radii of separation from the training set: $0$m ($488$k images), $500$m ($294$k), $1$km ($210$k), $2$km ($166$k), $3$km ($136$k), $4$km ($117$k) and $5$km ($107$k). As observed in \figref{fig:sep}, all methods, including retrieval-based approaches, are equally affected by this phenomenon, indicating that, as expected, the problem of global geolocation becomes harder as the separation radius increases. This allows us to define different versions of our test set tiered by difficulty.
% suggests that it is not specific to our method and confirms our intuition that the task gets truly harder. 
In particular, if we remove the separation between train and test makes the task becomes significantly easier:
 $3952$ geoscore for StreetCLIP in retrieval mode and $3852$ for our best model, corresponding to an average distance error of $1191$km.% in its regression mode which is more surprising. 

\paragraph{LoRA.}
Fig~\ref{fig:lora} shows the results with different widths of the LoRA bottleneck, ranging from 2 to 64. We share similar observations with the LoRA paper \cite[7.2]{hu2021lora}: higher ranks do not increase or even slightly decrease performance. Unfreezing the last transformer block remains more efficient in terms of training time, and fine-tuning the entire model leads to even better performance.

\paragraph{Erroneous Predictions.}
In Fig~\ref{fig:error-examples} we illustrate some sources of geolocation errors not related to the density of training images. These include landscapes that are:
(i) similar between very distant countries (Fig~\ref{fig:error-examples}~(a,b)), or
(ii) any key information is far away from the camera (Fig~\ref{fig:error-examples}~(b,c)), or are
(iii) monotonous and nearly featureless (Fig~\ref{fig:error-examples}~(c)).

\begin{figure}\centering
\begin{tabular}{@{}c@{\;}c@{\;}c@{}}
     \begin{subfigure}{0.15\textwidth}
     \includegraphics[width=1\textwidth, height=.08\textheight]{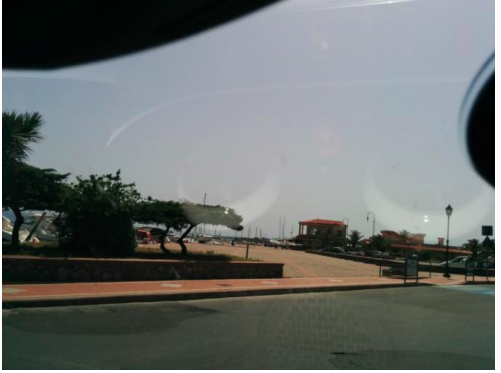}
     \caption{\textbf{GT:} Sardinia \\ \textbf{\textcolor{COMBINEDCOLOR}{Comb.}}: Senegal \\ \textbf{\textcolor{BASELINECOLOR}{Base}}: Mali \\ \textbf{\textcolor{magenta}{SCLIP}}: Italy }
     \label{fig:error:a}
     \end{subfigure}
     &
     \begin{subfigure}{0.15\textwidth}
     \includegraphics[width=1\textwidth, height=.08\textheight]{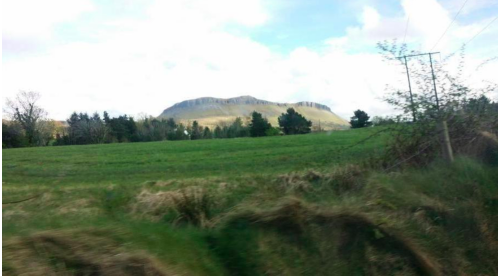}
     \caption{\textbf{GT}: Irland \\ \textbf{\textcolor{COMBINEDCOLOR}{Comb.}}: Lesotho \\ \textbf{\textcolor{BASELINECOLOR}{Base}}: Australia \\ \textbf{\textcolor{magenta}{SCLIP}}: USA}
     \label{fig:error:b}
     \end{subfigure}
     &
     \begin{subfigure}{0.15\textwidth}
     \includegraphics[width=1\textwidth, height=.08\textheight]{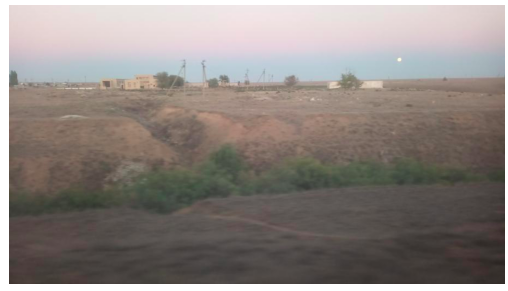}
      \caption{\textbf{GT}: Russia \\ \textbf{\textcolor{COMBINEDCOLOR}{Comb.}}: Erythrea \\ \textbf{\textcolor{BASELINECOLOR}{Base}}: Saudi Arabia \\ \textbf{\textcolor{magenta}{SCLIP}}: Turkmenistan }
      \label{fig:error:c}
     \end{subfigure}
\end{tabular}
\caption{
{\bf Erroneous Predictions.} Images that are consistently predicted wrongly despite being sampled from areas with relatively high density of training images.
}
\label{fig:error-examples}
\end{figure}

\begin{figure}
    \centering
    \begin{tabular}{ccc}    
        \begin{subfigure}[t]{.3\columnwidth}
            \includegraphics[width=\linewidth]{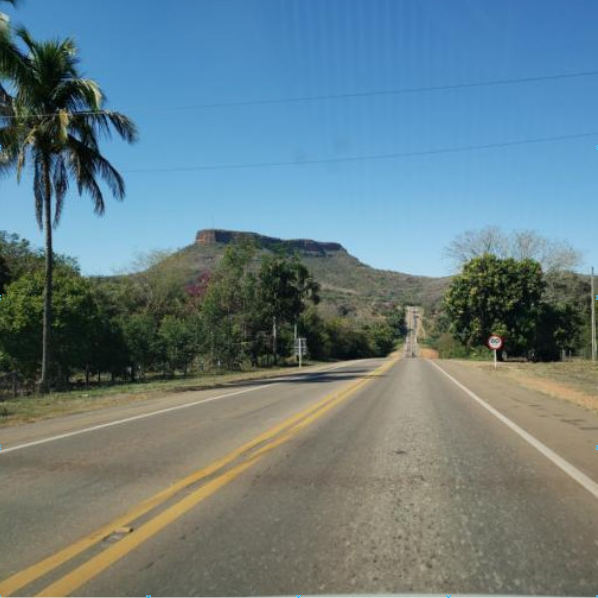}
           
            %\caption{Original Image}
            \label{fig:att:a}
        \end{subfigure}
        &
        \begin{subfigure}[t]{.3\columnwidth}
            \includegraphics[width=\linewidth]{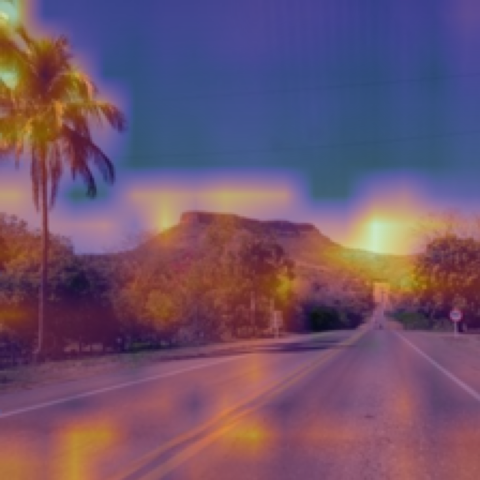}
           
            %\caption{Mean attentions}
            \label{fig:att:b}
        \end{subfigure}
        & 
        \begin{subfigure}[t]{.3\columnwidth}
            \includegraphics[width=\linewidth]{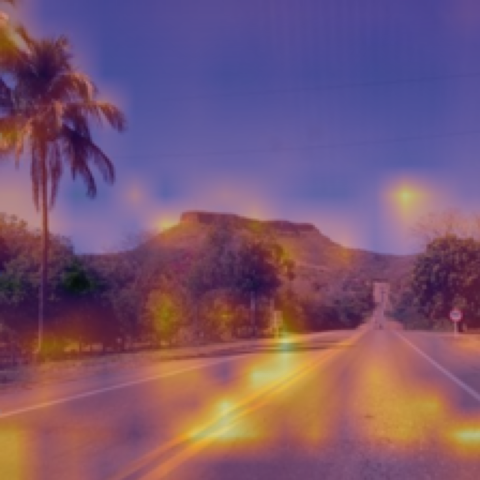}
           
            %\caption{Selected head}
            \label{fig:att:c}
        \end{subfigure} 
        \\
        \begin{subfigure}[t]{.3\columnwidth}
            \includegraphics[width=\linewidth]{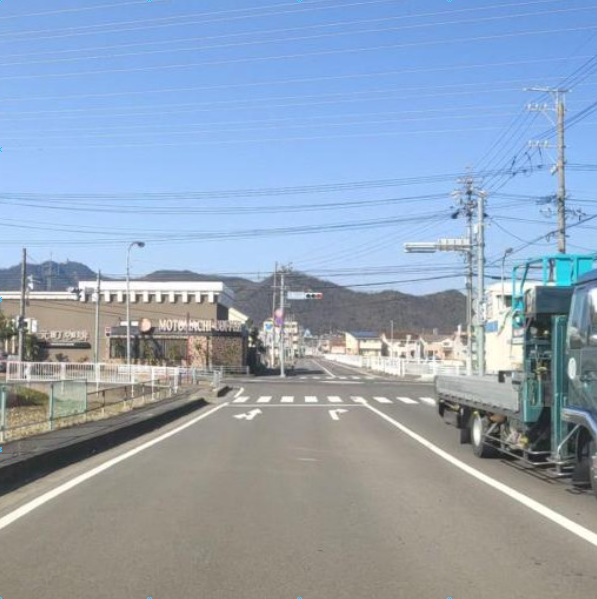}
           
            \caption{Input Image}
            \label{fig:att:d}
        \end{subfigure}
        & 
        \begin{subfigure}[t]{.3\columnwidth}
            \includegraphics[width=\linewidth]{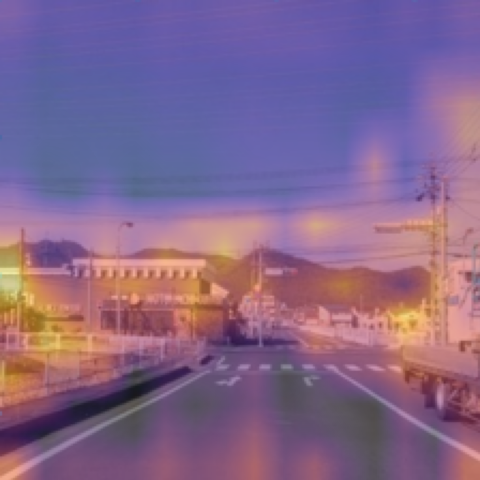}
           
            \caption{Mean attention}
            \label{fig:att:e}
        \end{subfigure}
        & 
        \begin{subfigure}[t]{.3\columnwidth}
            \includegraphics[width=\linewidth]{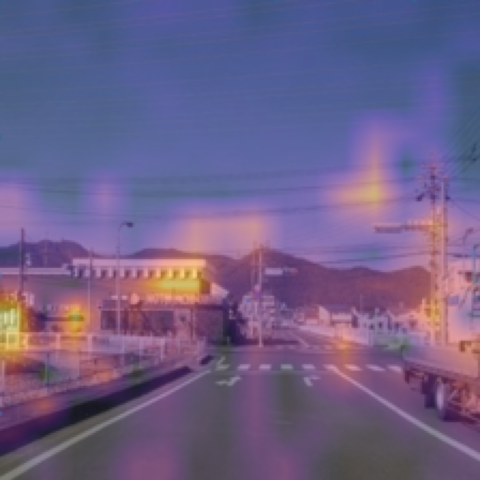}
           
            \caption{Selected head}
            \label{fig:att:f}
        \end{subfigure}
    \end{tabular}
   
    \vspace{-2mm}
    \caption{{\bf Attention Maps.} We visualize the self-attention maps of the \texttt{[CLS]} token of the last layer of the image encoder of the combined model. We show the mean across all heads in \subref{fig:att:e}, and manually selected an interesting layer in \subref{fig:att:f}.}
     \label{fig:att}
\end{figure}

\paragraph{Humans and Baselines.} We compare in \tabref{tab:humans} 
our models against two random baselines: selecting randomly a location on the map or the location of a random image from the training set. We also construct an Annotator Ensemble Oracle by selecting the most accurate prediction for each image from all annotators. Our baseline model, and more substantially our combined model, far surpasses the accuracy of individual annotators, but is still outmatched by the Annotator Ensemble Oracle.

\paragraph{Attention Maps.} We represent in \figref{fig:att} the self-attention maps of
the \texttt{[CLS]} token of the last layer of the combined model of images from the teaser. We observe that the network focuses on regions of interest containing useful geographical cues, such as the double yellow road line---a specific trait to certain countries---or vegetation and buildings.

\begin{table}[t]
    \caption{{\bf Annotator Performance.} We report the average performance of $80$ annotators on a subset of $50$ images.}
    \label{tab:humans}
    \centering
    \resizebox{\linewidth}{!}{
    \scriptsize
    \begin{tabular}{lccccc}
        \toprule
        & \multirow{2}{*}{\makecell[b]{Geo $\uparrow$\\score} } & \multirow{2}{*}{\makecell[b]{Dis $\downarrow$\\tance} }  & \multicolumn{3}{c}{Classification accuracy $\uparrow$} \\
        \cmidrule(lr){4-6}
        &   &  & continent & country & region \\\midrule
        
        Annot. performance & 1009 & \hphantom{0}6407 & 48.9 & 12.2 & \hphantom{0}3.0\\
         \rowcolor{gray!10} Annot. ensemble oracle & \bf 3919 & \bf  \hphantom{00}443 & \bf  98.0 & \bf 70.0 & 28.0\\ \greyrule
        Random location & \hphantom{0}120  & 10273   & 16.0 & \hphantom{0}0.0 & \hphantom{0}0.0\\
        \rowcolor{gray!10}  Random image & \hphantom{0}328   & \hphantom{0}8724 & 20.0 & \hphantom{0}2.0 & \hphantom{0}0.0 \\\greyrule
        \textcolor{BASELINECOLOR}{Base model} & 2235 & \hphantom{0}3247 & 74.0 & 36.0 & \hphantom{0}8.0\\
        \rowcolor{gray!10} 
 \textcolor{COMBINEDCOLOR}{Combined model} & 3333 & \hphantom{0}1948 & 86.0 & \bf 70.0 & \bf 34.0\\\bottomrule
    \end{tabular}
    }
\end{table} 
%===============================================================
\section{Implementation Details}
%===============================================================
\label{sec:sup:implem}

In this section, we detail our architecture, loss, metrics, and the retrieval algorithm.

\paragraph{Architecture.} All considered networks have a base image encoder $\cI \mapsto \bR^d$, with a $d$ which depends on each architecture ($d=768$ for the model ViT-B-32, and $d=1024$ for all the other encoders). We then add one or several heads to map the image representation to geographical information:
\begin{itemize}[itemsep=0.25em, wide, labelwidth=!, labelindent=2pt, topsep=0.25em, parsep=0em]
\item[-] \textit{Regression $f^\text{loc}$.} This network directly predicts the longitude and latitude of an image with a MLP of size $d \mapsto d \mapsto 64 \mapsto 2$ with group norms of 4 groups \cite{wu2018group} and without normalizing the last layer.
\item[-] \textit{Regression $f^\text{loc}$ sin/cos.} For this variation, we predict the cosine and sine of both coordinates with an MLP: $d \mapsto d \mapsto 64 \mapsto 4$ with a normalization that ensures that the squared sum between coordinate $0,1$ and $2,3$ is 1. We then use the $\mathbf{atan2}$ function to recover the corresponding coordinates.
\item[-] \textit{Classification $f^\text{classif}$.} To predict in which of the $K$ geographic divisions an image was taken, we use an MLP: $d \mapsto d \mapsto 512 \mapsto K$.
\item[-] \textit{Hybrid $f^\text{relative}$.} 
In the hybrid model, we predict both the division and the position of the image within this cell. The relative position is predicted \emph{for all cells} with an MLP $\phi^\text{relative}: d \mapsto d  \mapsto 512 \mapsto \bR^2K$ with a specialized normalization for the last layer, explained below. During inference, we select the relative prediction of the cell with the highest prediction score for $f^\text{classif}$. During training, we only supervise the relative prediction that corresponds to the true cell.

For this network, a key implementation detail is the normalization of the last layer of $\phi^\text{relative}$. We require that for each cell a prediction of $(0,0)$ should correspond to the centroid $h^\star,w^\star \in \cC^2$ of the training set images in the cell, and that a range of prediction of $[-1,1]^2$ covers the entire bounding box of size $h,w$. As illustrated in \figref{fig:hybrid}, we denote by $x^\star,y^\star \in [0,1]^2$ the relative position of the centroid in the cell and by $x,y$ the prediction of the MLP $\phi^\text{aux}$. The output of $f^\text{relative}$ is defined as follows:
\begin{align}
w^\star+w&
    \begin{cases}
         -{x}{x^\star}&\text{if}\; x\leq0\\
         x({1-x^\star}) &\text{else}
    \end{cases},
    \\
    h^\star+h&
     \begin{cases}
         -{y}{y^\star}&\text{if}\; y\leq0\\
         y({1-y^\star}) &\text{else}
    \end{cases}
~.
\end{align}
This normalization allows the network $\phi^\text{relative}$ to easily predict the centroid of the cell, which facilitates learning the distribution of images of that cell. This is particularly crucial for cells with an off-centered centroid, as it provides increased precision in high density areas. In practice, removing this normalization decreases the performance of the hybrid model by $59$ points of geoscore, or $22$\% from the benefit brought by using a hybrid model over pure classification.

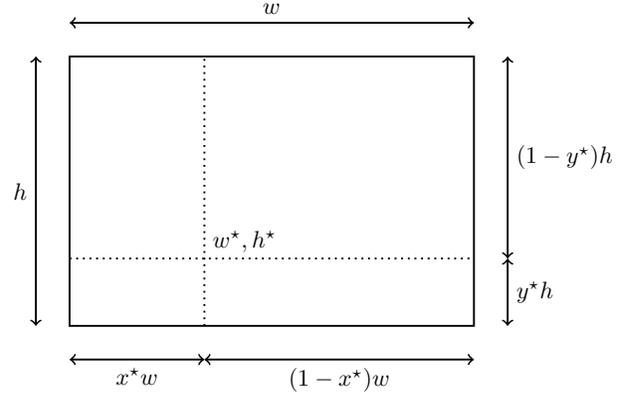
\begin{figure}[t]
    \centering
    \resizebox{\linewidth}{!}{
    \begin{tikzpicture}
        \draw [thick] (0,0) rectangle (6,4);
        \draw [thick, <->] (0,4.5) --node[above]{$w$}  (6,4.5);
        \draw [thick, <->] (-0.5,0) --node[left]{$h$}  (-0.5,4);
        \draw[thick, dotted] (2,0) -- (2,4);
        \draw[thick, dotted] (0,1) -- (6,1);
        \draw [thick, <->] (0,-0.5) --node[below]{$x^\star w$}  (2,-0.5);
        \draw [thick, <->] (2,-0.5) --node[below]{$(1-x^\star)w$}  (6,-0.5);
         \draw [thick, <->] (6.5,0) --node[right]{$y^\star h$}  (6.5,1);
        \draw [thick, <->] (6.5,1) --node[right]{$(1-y^\star)h$}  (6.5,4);
        \node[anchor=west] at (2,1.25) {$w^\star,h^\star$};
    \end{tikzpicture}
    }
    \caption{{\bf Hybrid Model.} The normalization of the hybrid model requires special considerations to ensure that the output $(x,y)$ of $\phi^\text{aux}$ is such $(0,0)$ maps to the cell's centroid $w^\star,h^\star$, and that $[-1,1]^2$ maps the entire cell.}
    \label{fig:hybrid}
\end{figure}

\item[-] \textit{Auxiliary $f^\text{aux}$.} Finally, the auxiliary network is an MLP $d \mapsto d \mapsto 64 \mapsto A $, where $A$  corresponds to the number of auxiliary task predictions: $11$ for land cover, $31$ for climate, $15$ for soil type, $1$ for the driving side, and $1$ for the distance to the nearest sea. For all classification tasks (\ie everything except the distance to the sea), we softmax the output logits.
\end{itemize}

\paragraph{Contrastive Learning.}  We use the MIL-NCE loss \cite{miech2020end} as our contrastive objective, which extends the InfoNCE loss \cite{oord2018representation} to cases where each sample can have multiple positive matches.
\begin{align}
\sum_{i \in \mathcal{B}}
    \log\!\!\left(
    \frac{
        \displaystyle{\sum_{p \in \mathcal{P}_i}}
        \!
            e^{
            f^{\text{img}}(i)^\intercal 
            f^{\text{img}}(p)
            /T
            }
    }
    {
        \displaystyle{\!\sum_{p \in \mathcal{P}_i}}
        \!
            e^{
            f^\text{img}(i)^\intercal 
            f^\text{img}(p)/T
            }
            \!\!+\!\!\!\!\!\!
        \displaystyle{\sum_{n \in \mathcal{B} \setminus \mathcal{P}_i}}\!\!\!\!
            e^{
            f^\text{img}(i)^\intercal 
            f^\text{img}(n)
            /T
            }
    }
    \!\!\right),
\end{align}
with $\mathcal{P}_i \subset \mathcal{B}$ the set of image positively paired with $i$ and $T$ a temperature parameter set as $0.1$. If an image has only one positive match, this equation becomes the InfoNCE loss \cite{oord2018representation}.

\paragraph{Nearest Neighbors Retrieval}
To perform nearest neighbor retrieval, we create a HNSW32 indexe using the FAISS library \cite{faiss} through the autofaiss package (\url{https://github.com/criteo/autofaiss}). This approach achieves fewer than 200 self-consistency errors per million with over 90\% compression rate.

During retrieval, our training set is divided into five parts, each requiring $15$ minutes for index computation and collectively consuming $15.6$GB of storage for StreetCLIP embeddings, our most resource-intensive model. This setup enables us to predict locations for 12,000 to 32,000 test images per second, depending on the model size.

Although retrieval methods demonstrate high performance and have been made efficient with approximate methods, it is important to note that they are not a learning technique, as they rely on already geographically relevant representations that are already learned.

%Instead of quadratic complexity nearest neighbors, we perform approximate nearest neighbors with HNSW32 indexes ($<200/1M$ self-consistency errors per 1M with more than 90\% compression rate) using FAISS\footnote{We interface FAISS through the autofaiss package: \url{https://github.com/criteo/autofaiss}} \cite{faiss}. We split our data in 5 parts, where building indexes takes a bit less than 15 minutes and where the cumulative storage is in the order of 15.6GB for StreetCLIP (out of 19.1GB needed to store features - the model weights are close to 2GB for comparison), from which we can easily do 16it/sec without parallelization (batched inference of 512 samples with clip can vary between around 23it/sec to 64it/sec depending on the model) but can be further sped up using parallelization. Albeit the space limitations, our non-parametric nearest neighbors model seems to be the most performance efficient method, as expected from ML literature in comparison to learning linear projection on top of our representations or kernels. Note however, that nearest neighbors are a way of specializing the inference of an already pretrained feature extractors to our dataset and are not responsible for learning good representations. In fact we show that models with more geographically specific representations work significantly better for this inference method. The results are in fact impressive with: The baseline performance starts from $XXX$.
%  $XXXX$ geoscore compared to $XXXX$ for our hybrid geolocation module. \TODO{The image encodings learned with our approach significantly improve on StreetCLIP's and DINOv2's.}

\if 10
\begin{figure}
    \centering
    \includegraphics[width=\linewidth, height=.10\textheight]{example-image}
    \caption{{Contrastive Training.} Similarity heatmap for Ireland.}
    \label{fig:enter-label}
\end{figure}
\fi

%===================================================
\section{Datasheet for Dataset}
%===================================================
\label{sec:datasheet}

\subsection{Motivation}

\begin{enumerate}[label=Q\arabic*]

\item \textbf{For what purpose was the dataset created?} Was there a specific task in mind? Was there a particular gap that needed to be filled? Please provide a description.

\begin{itemize}
\item 
OpenStreetView-5M (OSV-5M) is the first global scale, open-access, large dataset of street view images. Its goal is to enable the training and evaluation of modern computer vision approaches for global visual geolocation, which would depend until now on proprietary or expensive APIs such as Google Street View. More broadly, OSV-5M can be used to evaluate and improve representation learning.
\end{itemize}

\item \textbf{Who created the dataset (e.g., which team, research group) and on behalf of which entity (e.g., company, institution, organization)?}

\begin{itemize}
\item 
The dataset was created as part of a the ``IMAGINE Summer Hackathon'', an internal event of the LIGM/ENPC/UGE laboratory. All images of OSV-5M come from the Mapillary website, which is a platform where users upload georeferenced images.
\end{itemize}

\item \textbf{Who funded the creation of the dataset?} If there is an associated grant, please provide the name of the grantor and the grant name and number.

\begin{itemize}
\item 
This work was partially supported by the ANR project READY3D ANR-19-CE23-0007 and used the HPC resources of IDRIS under the allocation 2022-AD011012096R1 made by GENCI.
\end{itemize}

\item \textbf{Any other comments?} 

\begin{itemize}
\item All the images of OSV-5M are already openly accessible through Mapillary's heavily moderated database. We only selected a small fraction distributed across the globe, and added metadata from public sources. 
\end{itemize}

\subsection{Composition}

\item \textbf{What do the instances that comprise the dataset represent (e.g., documents, photos, people, countries)?} 

\begin{itemize}
\item OSV-5M is composed of street view images depicting various street scenes, captured by dash-cams of different vehicles from across the world.
\end{itemize}

\item \textbf{How many instances are there in total (of each type, if appropriate)?}

\begin{itemize}
\item  The training set contains 4,894,685 images, and the test set 210,122.
\end{itemize}

\item \textbf{Does the dataset contain all possible instances or is it a sample (not necessarily random) of instances from a larger set?} 

\begin{itemize}
\item OSV-5M is a small subset of $5.1$M images from the 1.8 billion images hosted on the Mapillary website.
\end{itemize}

\item \textbf{What data does each instance consist of?} 

\begin{itemize}
\item Each instance consists of a georeferenced street view image with a height of 512 pixels.
\end{itemize}

\item \textbf{Is there a label or target associated with each instance?} 

\begin{itemize}
\item \answerYes\; Each image is associated with the following targets: longitude and latitude, administrative division (country, region, sub-region, closest city), and labels corresponding to the local land cover, soil, and climate type at a resolution of 30 arc seconds (1km). We also add the distance to the nearest sea and the driving side of the country. 
\end{itemize}

\item \textbf{Is any information missing from individual instances?} 

\begin{itemize}
\item \answerYes\; Sub-regions are not defined for all countries, about $30$\% of the instances do not have a value for this field. 
\end{itemize}

\item \textbf{Are relationships between individual instances made explicit (e.g., users' movie ratings, social network links)?} 

\begin{itemize}
\item \answerNo\; The data is organized as a collection of images with no particular order or relations. However, the metadata allows a user to organize them based on different geographical criteria.
\end{itemize}

\item \textbf{Are there recommended data splits (e.g., training, development/validation, testing)?} 

\begin{itemize}
\item \answerYes\; We provide an official training and test set. Our implementation also proposes a validation split.
\end{itemize}

\item \textbf{Are there any errors, sources of noise, or redundancies in the dataset?} %\textit{If so, please provide a description.}

\begin{itemize}
\item \answerYes\; We have heavily filtered the dataset using semi-automatic methods to discard low-quality images and wrong localization, as presented in \secref{sec:sup:construction}. We have estimated through the manual inspection of $4500$ images that 96.1\% (±0.57\% with a 95\% confidence level) of the images in OpenStreetView-5M are perceptually localizable, \ie provide a clear enough overview of their surroundings.
\end{itemize}

\item \textbf{Is the dataset self-contained, or does it link to or otherwise rely on external resources (e.g., websites, tweets, other datasets)?} 
\begin{itemize}
\item \answerNo\; OSV-5M is self-contained and will be stored and distributed on \url{huggingface.co}.
\end{itemize}

\item \textbf{Does the dataset contain data that might be considered confidential (e.g., data that is protected by legal privilege or by doctor–patient confidentiality, data that includes the content of individuals’ non-public communications)?} 

\begin{itemize}
\item \answerNo\; OSV-5M relies on crowdsourced data, whose license is respected by providing usernames for each image, which is include in our metadata.
\end{itemize}

\item \textbf{Does the dataset contain data that, if viewed directly, might be offensive, insulting, threatening, or might otherwise cause anxiety?} \textit{If so, please describe why.}

\begin{itemize}
\item \textbf{Highly unlikely}: OSV-5M contains $5$ million images of streets that come from Mapillary, which imposes a strong crowd-sourced moderation policy.
\end{itemize}

\item \textbf{Does the dataset relate to people?} 
\begin{itemize}
\item  \answerYes\; Many of the images of OSV-5M contain vehicles and some contain pedestrians, yet Mappilary performs highly accurate privacy blurring.\footnote{\label{af}See \url{https://blog.mapillary.com/update/2018/04/19/accurate-privacy-blurring-at-scale.html}}
\end{itemize}

\item \textbf{Does the dataset identify any subpopulations (e.g., by age, gender)?}

\begin{itemize}
\item \answerNo\; The metadata contains no information about the people present in the photography beyond, who are also privacy blurred.$^{\textcolor{red}{\textbf{1}}}$
\end{itemize}

\item \textbf{Is it possible to identify individuals (i.e., one or more natural persons), either directly or indirectly (i.e., in combination with other data) from the dataset?}

\begin{itemize}
\item \answerNo\; The license plates and faces of pedestrians have been privacy blurred by Mapillary using an automatic algorithm with over $99$\% recall for faces and $99.9$\% recall for license plates.$^{\textcolor{red}{\textbf{1}}}$ Furthermore, users can signal images that violate privacy.

We also manually inspected 4500 images and observed no confidentiality leak. With a confidence of $95$\% we can assume that fewer than $0.067$\% of the dataset contains leaks.
\end{itemize}

\item \textbf{Does the dataset contain data that might be considered sensitive in any way (e.g., data that reveals racial or ethnic origins, sexual orientations, religious beliefs, political opinions or union memberships, or locations; financial or health data; biometric or genetic data; forms of government identification, such as social security numbers; criminal history)?} 

\begin{itemize}
\item \answerNo\;
\end{itemize}

\item \textbf{Any other comments?}

\begin{itemize}
\item \answerNo\; 
\end{itemize}

\subsection{Collection Process}

\item \textbf{How was the data associated with each instance acquired?} 

\begin{itemize}
The images of Mapillary are taken and uploaded by users of the Mapillary platform. 
We downloaded the images directly from Mapillary's API. 
Additional metadata was collected from the following open-access sources:
(i) land cover: Global Land Cover Share Database \cite{latham2014global}
(ii) climate: Köppen-Geiger climate classification maps \cite{beck2018present}, 
(iii) soil type: Digital World Soil Map \cite{sanchez2009digital}
(iv) administrative division: reverse geocoder \cite{geocoder}.
\end{itemize}

\item \textbf{What mechanisms or procedures were used to collect the data (e.g., hardware apparatus or sensor, manual human curation, software program, software API)?} 

\begin{itemize}
\item We used Mapillary's web API and a Python script running on a standard workstation.
\end{itemize}

\item \textbf{If the dataset is a sample from a larger set, what was the sampling strategy (e.g., deterministic, probabilistic with specific sampling probabilities)?}

\begin{itemize}
\item We first defined a $100\times100$m grid across the entire world and sampled one image per cell among the $1.8$B images of Mapillary. We then sample the train and test sets with a weight proportional to the local image density raised to the power of $-0.75$. We then filter the images based on both learned and handcrafted filters, as described in \secref{sec:sup:construction}.
\end{itemize}

\item \textbf{Who was involved in the data collection process (e.g., students, crowdworkers, contractors) and how were they compensated (e.g., how much were crowdworkers paid)?}

\begin{itemize}
\item The images are crowdsourced by Mapillary users who agree on Mapillary's \href{https://www.mapillary.com/terms}{terms of use}. To the best of our knowledge, users are not compensated.
\end{itemize}

\item \textbf{Over what timeframe was the data collected? Does this timeframe match the creation timeframe of the data associated with the instances (e.g., recent crawl of old news articles)?} 

\begin{itemize}
\item The images used in OSV-5M were uploaded between January 2011 and August 2023.
\end{itemize}

\item \textbf{Were any ethical review processes conducted (e.g., by an institutional review board)?} 

\begin{itemize}
\item \answerNo\;
\end{itemize}

\item \textbf{Did you collect the data from the individuals in question directly, or obtain it via third parties or other sources (e.g., websites)?}

\begin{itemize}
\item \answerNA\; The images were downloaded through Mapillary's API.
\end{itemize}

\item \textbf{Were the individuals in question notified about the data collection?} 

\begin{itemize}
\item \answerNo\; We followed the terms of use of Mapillary.
\end{itemize}

\item \textbf{Did the individuals in question consent to the collection and use of their data?}

\begin{itemize}
\item \answerYes\; Following the 
Mapillary terms of use, a user agrees for their data to be be used respecting the CC BY-SA 2.0 DEED license.
\end{itemize}

\item \textbf{If consent was obtained, were the consenting individuals provided with a mechanism to revoke their consent in the future or for certain uses?} 

\begin{itemize}
\item \answerNA\;
\end{itemize}

\item \textbf{Has an analysis of the potential impact of the dataset and its use on data subjects (e.g., a data protection impact analysis) been conducted?} 

\begin{itemize}
\item \answerNo\; However, users of OSV-5M can signal potential issues with the images to the corresponding authors. Flagged images will be removed and Mapillary will be further contacted.

\end{itemize}

\item \textbf{Any other comments?}

\begin{itemize}
\item All the images of OSV-5M are already openly accessible through Mapillary's heavily moderated database. We only added additional metadata from public sources.
\end{itemize}

\subsection{Preprocessing, Cleaning, and/or Labeling}

\item \textbf{Was any preprocessing/cleaning/labeling of the data done (e.g., discretization or bucketing, tokenization, part-of-speech tagging, SIFT feature extraction, removal of instances, processing of missing values)?} 

\begin{itemize}
\item \answerYes\; We removed the images based on learned and handcrafted filters, as described in \secref{sec:sup:construction}. In particular, we removed images that were classified as blurry, too dark or purple, or badly exposed. We also used a pretrained model \cite{gidaris2018unsupervised} to detect and remove images with potential spurious orientation.
\end{itemize}

\item \textbf{Was the “raw” data saved in addition to the preprocessed/cleaned/labeled data (e.g., to support unanticipated future uses)?} \textit{If so, please provide a link or other access point to the “raw” data.}

\begin{itemize}
\item \answerYes\; The removed images are saved on a local server but are not public. Note that all these images, including the filtered ones, are still available on Mapillary's website.
\end{itemize}

\item \textbf{Is the software used to preprocess/clean/label the instances available?} 

\begin{itemize}
\item \answerYes\; The script used for cleaning the dataset will be released alongside the dataset.
\end{itemize}

\item \textbf{Any other comments?}

\begin{itemize}
\item \answerNo\;
\end{itemize}

\subsection{Uses}

\item \textbf{Has the dataset been used for any tasks already?} 

\begin{itemize}

\item \answerYes\; To train and evaluate geolocation models, the subject of the paper.
\end{itemize}

\item \textbf{Is there a repository that links to any or all papers or systems that use the dataset?} 

\begin{itemize}
\item \answerNo\; But once we release the dataset we will maintain an updated list on the project page.
\end{itemize}

\item \textbf{What (other) tasks could the dataset be used for?}

\begin{itemize}
\item The images of OSV-5M can be used for both self-supervised learning and generative modeling, both as a pretraining or fine-tuning dataset. The meta-data beyond geolocation can be used as targets for separate tasks.
\end{itemize}

\item \textbf{Is there anything about the composition of the dataset or the way it was collected and preprocessed/cleaned/labeled that might impact future uses?} 

\begin{itemize}
\item The density-based sampling leads to a spatial distribution that may not fit other datasets and tasks.
\end{itemize}

\item \textbf{Are there tasks for which the dataset should not be used?} 

\begin{itemize}
\item \answerYes The same limitations that apply for Mapillary data (CC BY-SA 2.0 DEED), also apply to our dataset.
\item 
{\bf Privacy Concerns.} Despite being heavily moderated, the dataset may contain images of individuals or private residences. Usage must avoid applications that can infringe on personal privacy or exercise surveillance and open-source intelligence (OSINT).

\item {\bf Cultural and Ethical Sensitivity.} The dataset spans a wide range of cultures and countries, each with its own set of ethical norms and cultural sensitivities. We strongly advise against using OSV-5M in a way that might propagate stereotypes, misrepresent cultures, or otherwise harm the dignity and representation of the featured communities.

\item {\bf Manipulation and Misrepresentation.} The dataset should not be used to create misleading representations of locations or to manipulate images in a way that distorts or misrepresents the reality of the places and the depicted people.
\end{itemize}

\item \textbf{Any other comments?}

\begin{itemize}
\item \answerNo\;
\end{itemize}

\subsection{Distribution}

\item \textbf{Will the dataset be distributed to third parties outside of the entity (e.g., company, institution, organization) on behalf of which the dataset was created?} 

\begin{itemize}
\item \answerYes\; The dataset will be open-access and accessible to the research community.
\end{itemize}

\item \textbf{How will the dataset be distributed (e.g., tarball on website, API, GitHub)?} 

\begin{itemize}
\item The data will be hosted on \url{huggingface.co}.
\end{itemize}

\item \textbf{When will the dataset be distributed?}

\begin{itemize}
\item The dataset will be distributed upon the publication of the preprint on arXiv, which should be in Q2 of 2024.
\end{itemize}

\item \textbf{Will the dataset be distributed under a copyright or other intellectual property (IP) license, and/or under applicable terms of use (ToU)?} \textit{If so, please describe this license and/or ToU, and provide a link or other access point to, or otherwise reproduce, any relevant licensing terms or ToU, as well as any fees associated with these restrictions.}

\begin{itemize}
\item \answerYes\; The dataset inherits from Mappilary CC-BY-SA license: free of use with attribution to the authors of the images~\cite{licence}.
\end{itemize}

\item \textbf{Have any third parties imposed IP-based or other restrictions on the data associated with the instances?} 

\begin{itemize}
\item \answerNo\;
\end{itemize}

\item \textbf{Do any export controls or other regulatory restrictions apply to the dataset or to individual instances?} 

\begin{itemize}
\item \answerNo\;
\end{itemize}

\item \textbf{Any other comments?}

\begin{itemize}
\item \answerNo\;
\end{itemize}

\subsection{Maintenance}

\item \textbf{Who will be supporting/hosting/maintaining the dataset?}

\begin{itemize}
\item The authors will maintain the dataset. The dataset will be hosted on \url{huggingface.co}.
\end{itemize}

\item \textbf{How can the owner/curator/manager of the dataset be contacted (e.g., email address)?}

\begin{itemize}
\item A dedicated email will be created.
\end{itemize}

\item \textbf{Is there an erratum?} 

\begin{itemize}
\item \answerNo\; There is no erratum for our initial release. Errata will be documented as future releases on the dataset website.
\end{itemize}

\item \textbf{Will the dataset be updated (e.g., to correct labeling errors, add new instances, delete instances)?} 

\begin{itemize}
\item \answerYes\;
\end{itemize}

\item \textbf{If the dataset relates to people, are there applicable limits on the retention of the data associated with the instances (e.g., were individuals in question told that their data would be retained for a fixed period of time and then deleted)?} 

\begin{itemize}
\item \answerNA\;
\end{itemize}

\item \textbf{Will older versions of the dataset continue to be supported/hosted/maintained?} 

\begin{itemize}
\item \answerYes\; We are dedicated to providing ongoing support for the OSV-5M dataset.
\end{itemize}

\item \textbf{If others want to extend/augment/build on/contribute to the dataset, is there a mechanism for them to do so?} 

\begin{itemize}
\item \answerYes\; The data is free of use under Mappilary CC-BY-SA license. User making explicit use of our proposed split should cite our paper. %Users willing to contribute can also propose pull-requests on the dataset's repository.
\end{itemize}

\item \textbf{Any other comments?}

\begin{itemize}
\item \answerNo\;
\end{itemize}

\end{enumerate}

}{}

\end{document}